\documentclass[table, sigconf, pdftex]{acmart}

\usepackage{tikz}
\usepackage{amsmath}
\usepackage{lipsum}
\usepackage{url}
\usepackage{tipa}
\usepackage{graphicx}
\usepackage{subfigure}
\usepackage{epstopdf}
\usepackage{multirow}
\usepackage{booktabs}
\usepackage{bigstrut}
\usepackage{hyperref}
\usepackage{xspace}
\usepackage{titlesec}
\usepackage{colortbl}
\usepackage{makecell}
\usepackage{enumitem}
\usepackage{CJKutf8}
\usepackage[most]{tcolorbox}

\setcopyright{none} 
\acmConference[Anonymous Submission to ACM CCS 2021]{ACM Conference on Computer and Communications Security}{Due 6 May 2021}{London, TBD}
\acmYear{2021}

\newcommand{\fw}{fuzzy words\xspace}
\newcommand{\FW}{Fuzzy Words\xspace}

\newcommand{\sys}{\textit{FakeWake}\xspace}

\begin{document}
\begin{CJK*}{UTF8}{gbsn}
\title{\sys: Understanding and Mitigating Fake Wake-up Words of Voice Assistants} 

\begin{abstract}
In the area of Internet of Things (IoT) voice assistants have become an important interface to operate smart speakers, smartphones, and even automobiles. To save power and protect user privacy, voice assistants send commands to the cloud only if a small set of pre-registered wake-up words are detected. However, voice assistants are shown to be vulnerable to the \sys phenomena, whereby they are inadvertently triggered by innocent-sounding \fw. In this paper, we present a systematic investigation of the \sys phenomena from three aspects. To start with, we design the first fuzzy word generator to automatically and efficiently produce \fw instead of searching through a swarm of audio materials. We manage to generate 965 \fw covering 8 most popular English and Chinese smart speakers. To explain the causes underlying the \sys phenomena, we construct an interpretable tree-based decision model, which reveals phonetic features that contribute to false acceptance of \fw by wake-up word detectors. Finally, we propose remedies to mitigate the effect of \sys. The results show that the strengthened models are not only resilient to \fw but also achieve better overall performance on original training datasets.
\end{abstract}

\author{Yanjiao Chen}
\affiliation{%
  \institution{Zhejiang University}
}
\email{chenyanjiao@zju.edu.cn}

\author{Yijie Bai}
\affiliation{%
  \institution{Zhejiang University}
}
\email{baiyj@zju.edu.cn}

\author{Richard Mitev}
\affiliation{%
  \institution{Technical University of Darmstadt}
}
\email{richard.mitev@trust.tu-darmstadt.de}

\author{Kaibo Wang}
\affiliation{%
  \institution{Zhejiang University}
}
\email{kaibo@zju.edu.cn}

\author{Ahmad-Reza Sadeghi}
\affiliation{%
  \institution{Technical University of Darmstadt}
}
\email{ahmad.sadeghi@trust.tu-darmstadt.de}

\author{Wenyuan Xu}
\affiliation{%
  \institution{Zhejiang University}
}
\email{xuwenyuan@zju.edu.cn}



\maketitle

	\section{Introduction}
	
	
	Voice assistants are popular interfaces embedded in smart Internet of Things (IoT) devices (e.g., smart speakers), which enable us to use voice commands to execute various operations, e.g., send messages, make calls, and even control (e.g., open the door) their IoT ecosystem (e.g., smart home appliances). Despite the recession under the influence of COVID-19, the global smart speaker market is expected to grow by 21\% in 2021~\cite{forecast}. With the omnipresence of voice assistants in the near future, the potential threats to user privacy and security regarding misconduct of voice assistants have to be addressed.
	
	
	Almost all voice assistants adopt the wake-up mechanism. Before being triggered for receiving voice commands, voice assistants actively listen to the surrounding environment for wake-up words, which are usually short and catchy words chosen by manufacturers to brand their products\footnote{A few voice assistants, e.g., Xiaomi, allow users to customize their own wake-up words, which may even aggravate the \sys phenomena if users choose convenient but commonly-used words.}. Once the lightweight local detection model believes that it has detected a wake-up word, the voice assistant will record and send audio to the cloud for further analysis. 
	
\begin{figure}
    \centering
    \includegraphics[width=0.97\linewidth, bb=0 0 12.12in 5.92in ]{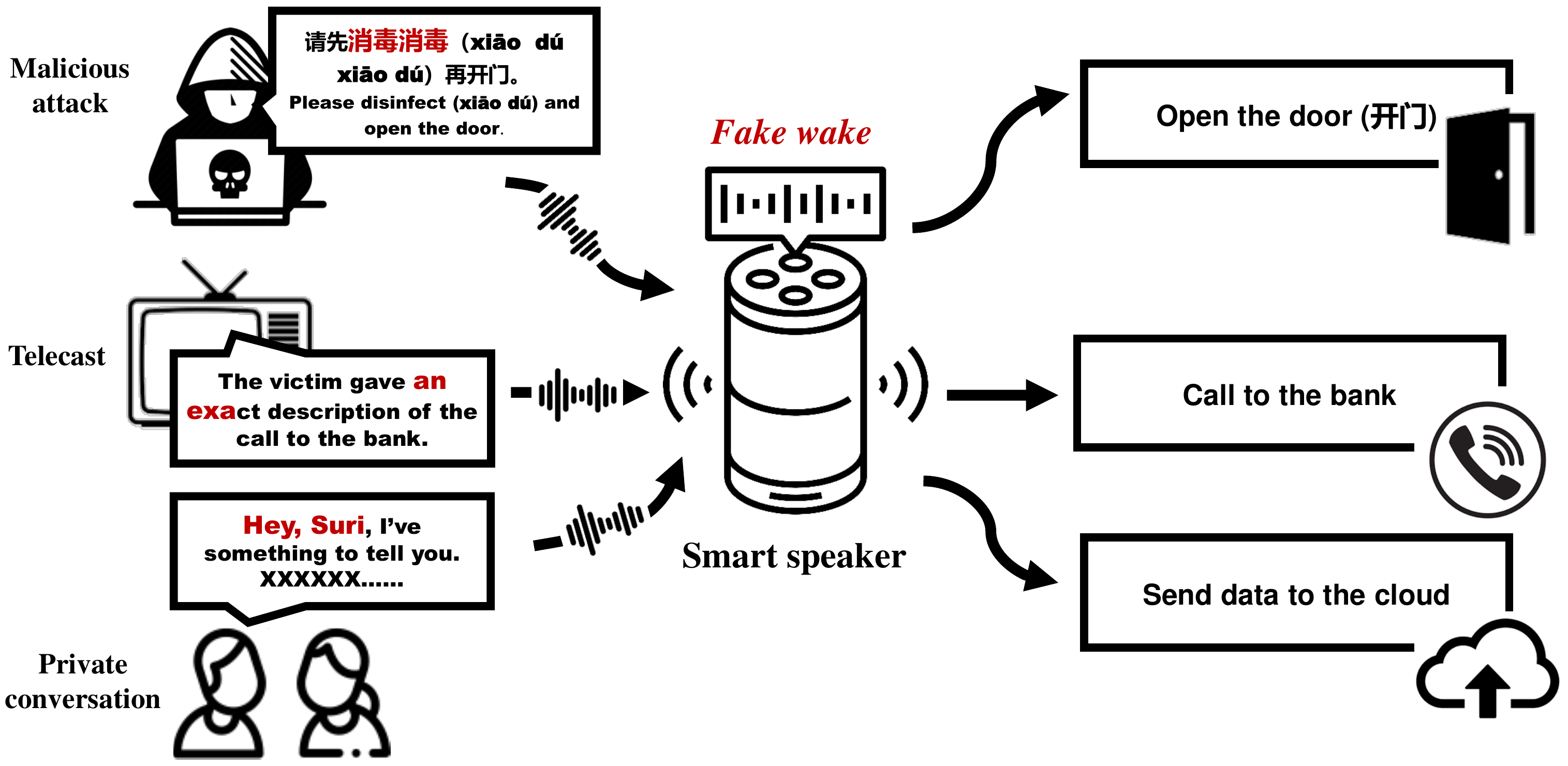}
    \caption{An illustration of the \sys phenomena. Various sources from the attacker-generated audios, TV shows and private conversations may incur the \sys phenomena, and result in privacy leakage and security threats. }
    \vspace{-0.2in}
    \label{scenario}
    \end{figure}

	Unfortunately, voice assistants suffer from the \sys phenomena, whereby they can be wrongly activated by words that are not wake-up words. We define the words that are not wake-up words but induce the \sys phenomena as \emph{\fw}, and the ones that do not activate voice assistants as non-\fw. The \sys phenomena is fairly prevalent: Recent surveys show that 50\% of users wake their voice assistants up by mistake once a week and 28.5\% of them even experience daily accidental wake-up~\cite{manifestsurvey, voicebotsurvey}. As shown in Figure~\ref{scenario}, the \sys phenomena can be incurred by sources such as human conversation, TV shows~\cite{When, schonherr2020unacceptable}, and TTS-spoken texts~\cite{mitev2020leakypick, schonherr2020unacceptable}. The \sys phenomena pose privacy and security risks, e.g., uploading audio with sensitive information to the cloud or accepting malicious commands without be noticing. The Amazon Echo has been reported to be activated mistakenly and sent the recorded private conversation of family members to various random contacts~\cite{news}. 
	Prior efforts have found several \fw~\cite{When, mitev2020leakypick, schonherr2020unacceptable} without understanding why and how to defend against them. Thus, in this paper, we aim to systematic study the root causes and mitigation of the \sys problems. 
	
	Particularly, we focus on studying the \sys phenomena, aiming to answer the following questions. 
	
\begin{itemize}[leftmargin=*]
\item \emph{How to efficiently generate a large collection of \fw for a given voice assistant?}
\item \emph{What are the causes that lead to false acceptance of \fw by wake-up detectors?}
\item \emph{How to strengthen wake-up detectors of voice assistants to be resilient against \fw?}
\end{itemize}

\textbf{Generating}. First of all, we target at generating large quantities of \fw for a given voice assistants in an efficient manner, which provides the samples for analyzing the causes of the \sys phenomena and to strengthen the wake-up word detector. A naive solution is to continuously play audio materials and record whether the smart speakers are activated or not, which takes days or even weeks to find a few dozens of mis-activation incidents~\cite{When, mitev2020leakypick, schonherr2020unacceptable}, and in many cases, the triggers are the real wake-up words themselves articulated in audio materials. For the sake of security, we are interested in \fw that are not only able to activate the voice assistant but also sound dissimilar to the real wake-up word to avoid being detected by users. The task is made challenging because commodity voice assistants are typically black-box and we have little information 
of the AI-based wake-up word detection model. To address this challenge, we carefully design a framework for fuzzy word generation, which mutates the best candidates for \fw to quickly create new \fw through multiple evolutionary generations, and balances the wake-up rate and the dissimilarity distance. 

Additionally, we investigate voice assistants for both English and Chinese, which have the most speakers worldwide~\cite{languagespeaker}. To customize the generation framework for English and Chinese, we need to encode the English and Chinese words into vectors and quantify the dissimilarity distance between two English or Chinese words. Nonetheless, the word composition and pronunciation rules of English and Chinese are different, making it difficult to apply the same encoding system and dissimilarity measurement to the two languages. After carefully investigating the word structure and pronunciation patterns of English and Chinese, we tailor the generation framework to cater to the linguistic features of the two languages respectively.

Using our generation framework, we manage to find a total of 965 \fw within 4 hours instead of 13 days~\cite{schonherr2020unacceptable}, covering 8 popular English and Chinese smart speakers, i.e., Amazon Echo, Echo Dot, Google Assistant, Apple Siri,  Baidu, Xiaomi, AliGenie, and Tencent. In particular, we have found 130 \fw for Echo Dot and 322 for AliGenie. The subjective tests with human volunteers verify that the generated \fw sound far from the real wake-up words, which means that these \fw may be used to wrongly activate voice assistants in a more surreptitious way.

\textbf{Understanding}. Given the generated \fw, we target at revealing why wake-up word detectors wrongly accept these \fw. Under the black-box settings, explaining the \sys phenomena is challenging since we have no knowledge of the internal structure and parameters of wake-up word detectors, thus unable to gauge the cause of \sys at the model level. A possible way of explanation is to measure the Levenshtein distance between \fw and the real wake-up words~\cite{schonherr2020unacceptable}. However, experiments show that our generated \fw have similar Levenshtein distance as non-\fw to the real wake-up words. To address this problem, we develop a more sophisticated explanation framework. To start with, we train an interpretable tree-based binary classifier to distinguish \fw from non-\fw, based on which we deduce a dissimilarity score that can well separate \fw and non-\fw. Then, we pinpoint the features that contribute the most to the false acceptance of \fw based on the SHAP value~\cite{shap}. It is demonstrated that the decisive factors usually concentrate on a small snippet of the word, e.g., \emph{ks} in Alexa and \emph{ai} in Xiaomi (the wake-up word is xi\v ao \`ai t\'ong xu\'e, i.e., 小爱同学). We show that a wake-up word detector that concentrates on fewer decisive factors will have more \fw. 

Knowing the decisive factors that lead to false acceptance of \fw is helpful in two aspects. On the one hand, we can quickly construct \fw by keeping the decisive factors and alter the other parts of the words. On the other hand, wake-up word detectors may be strengthened against \fw by paying special attention to the decisive factors.

\begin{figure}[tt]
        \centering
        \includegraphics[width=0.95\linewidth,bb=0 0 9.18in 7.44in]{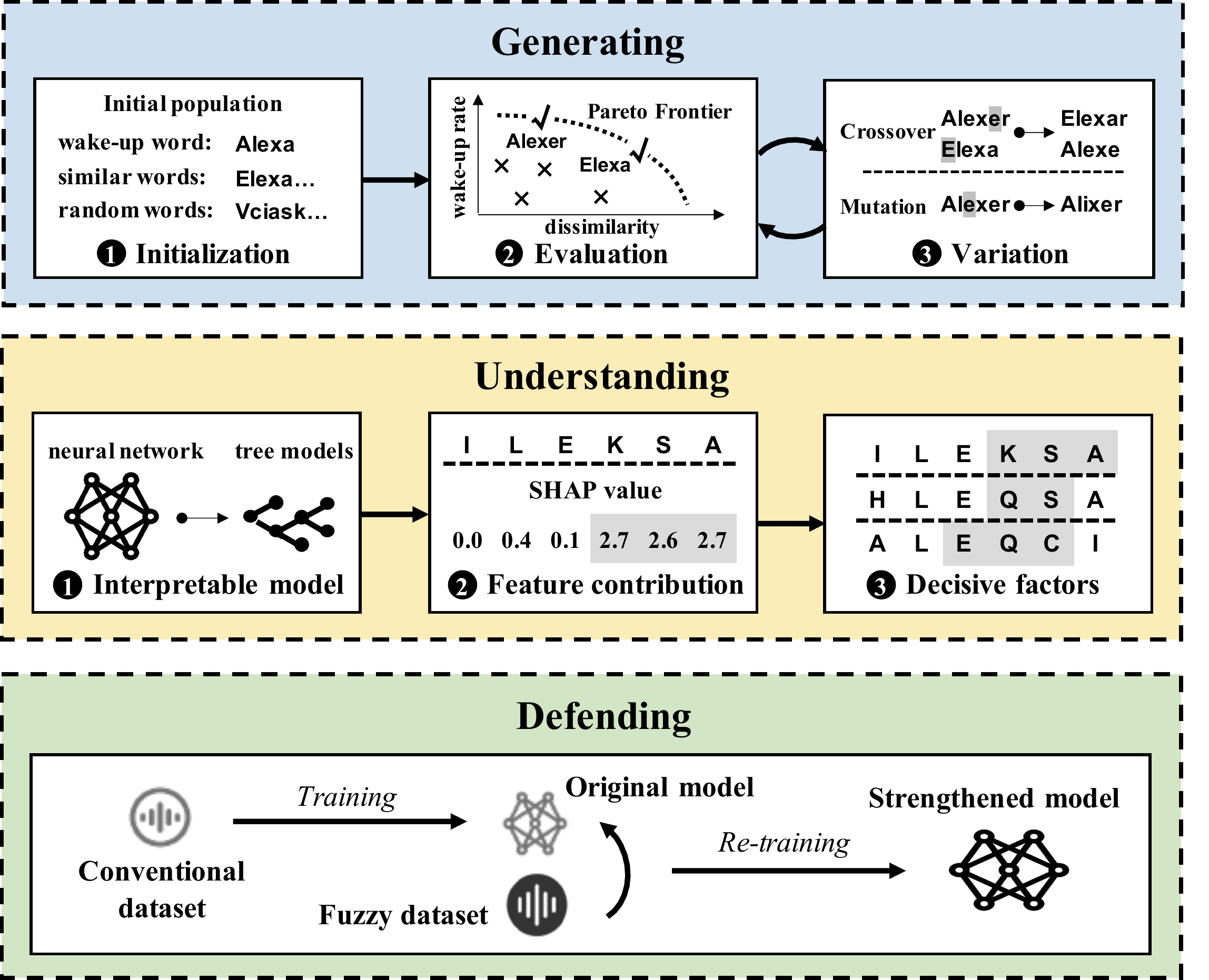}
        \caption{A systematic study on the \sys phenomena. We generate large quantities of \fw for both English and Chinese smart speakers, and then analyze these \fw to unveil the causes of their false acceptance by wake-up word detectors, based on which we propose remedies to mitigate the \sys phenomena. }
        \vspace{-0.2in}
        \label{overflow}
    \end{figure}

\textbf{Mitigating}. After understanding the causes of false acceptance of \fw, we can leverage the findings to help defend against the \sys phenomena, which is an unexplored territory, possibly due to a lack of access to commercial models. In regard to this, we propose two potential remedies. The first approach is to screen input audios for decisive factors, e.g., \emph{ks}. If there is no decisive factor, the audio is fed into the lightweight wake-up word detector for decision-making; otherwise, the audio will be scrutinized by more complicated speech recognition models. The second method is to strengthen wake-up word detectors by retraining with the generated \fw. As the wake-up word detectors on commercial voice assistants are unavailable, we resort to the open-source GRU recurrent network model \textsc{Precise}~\cite{mycroftprecise}. Surprisingly, our experiments on five wake-up word detectors of "Alexa", "Computer", "Athena", "Hi Xiaowen" and "Hi Mia" show that the strengthened models not only reject more than 97\% of the \fw, but also become better at distinguishing non-\fw. The possible reason is that the \fw are near the decision boundaries of wake-up word detectors, which helps the detectors to learn the decision boundaries in a more efficient and more precise way. The main flow of our paper is summarized in Figure~\ref{overflow}.

In summary, our main contributions are as follows.
	\begin{enumerate}
        \item We propose a systematic and automatic generation framework for producing \fw and customize the framework for both English and Chinese voice assistants.
        We conduct extensive evaluations on eight most popular English and Chinese smart speakers, and find a total of 965 \fw.
        \item We build an explanation framework for the \sys phenomena, which locates the decisive factors that lead to false acceptance of \fw. 

        \item We present countermeasures to strengthen wake-up word detectors against \fw, which improve the overall performance of wake-up word detectors. 
    \end{enumerate}

	 Currently, manufacturers choose wake-up words with more concerns on commercial interests than on security. With our effort on dissecting the \sys problem, we hope to raise the attention on potential risks of wake-up words and motivate future works on improving the security of wake-up words and the robustness of wake-up word detectors.


	
	\section{Background}
	
	
\subsection{Voice Assistant}
	
Voice assistants enable smart devices to take voice commands from users to control connected smart systems. Almost all popular smart speakers and most smartphones are equipped with embedded voice assistants, e.g., Amazon Echo, Apple Siri, and Google Home. To reduce power consumption and protect user privacy, nearly every voice assistant uses a wake-up mechanism, i.e., no uploading recorded audios to the cloud until the wake-up word is detected. The wake-up word(s) for a voice assistant are usually unique or limited to a pre-registered set of words. For instance, Amazon Echo uses "Alexa" as its wake-up word and Google Home can be woken up by either "OK Google" or "Hey Google". Some Chinese voice assistants use their brand names as the wake-up words, e.g., "ti\=an m\=ao jīng l\'ing" for AliGenie (named 天猫精灵) and "xi\v ao d\`u xi\v ao d\`u" for Baidu smart speakers (named 小度).

	
It is known that voice assistants can be mistakenly woken up by \fw other than the authentic wake-up words~\cite{manifestsurvey, voicebotsurvey}, which raises security and privacy concerns. If such \fw occur inadvertently in conversations or if malicious attackers play innocent-sounding audio files containing \fw, smart speakers may be activated by mistake and transmit the recorded voice afterwards to the cloud or to specific contacts (e.g., the attacker). To make matters worse, attackers may issue malicious commands to the activated voice assistants, e.g., open the door or turn off the alarm system, which poses great threat to user safety. Therefore, to protect voice assistants from being wrongly activated by \fw is of great importance. 
	
\subsection{Wake-up Word Detection}
%

Wake-up word detection is essential to voice assistants. During the standby mode, voice assistants listen to the environment and record snippets of audio samples (e.g., 3s) to check the presence of wake-up words. 
A voice-activity detection module confirms the presence of voice, and then extract features, e.g., the mel-frequency cepstrum coefficients (MFCC), to feed into detection models (e.g., GMM, HMM, DNN) to determine the existence of wake-up words. For most voice assistants, there is a lightweight local detector model deployed on the smart devices and a more complicated model deployed on the cloud. Only the audio samples that are believed to contain wake-up words by the local model will be sent to the remote model for further examination. 
Different from speech recognition models, wake-up word detectors are keyword-spotting models that only focus on differentiating a specific keyword from all other words rather than translating the texts for any audio content. The output of wake-up word detectors (accept or reject) is only known to the manufacturer but will not be fed back to the users.

\subsection{Threat Model}

Considering searching for \fw for commercial voice assistants, we make the following assumptions. 

\textbf{No access to the wake-up word detection model (black-box)}. The attacker has no knowledge of the detection model, including model structure, parameters and hyperparameters\footnote{Reverse-engineering the wake-up word detector, e.g., using model extraction methods, is possible. However, most recent model extraction methods achieve only about 70\% agreement rate between the substitute model and commercial APIs~\cite{yu2020cloudleak}. }. The output labels and confidence scores of the wake-up word detector is also unavailable. 
The attacker can only interact with the voice assistant and observe whether it is activated or not, e.g., on/off of the LED.

\textbf{No access to the training dataset}. The training datasets of wake-up word detectors are privately collected by manufacturers with regard to the unique wake-up word of their products. The attacker has no access to the training dataset, thus cannot obtain the exact wake-up word detection model or infer any deficiency of the training process. 

\textbf{Attacker's ability}. We assume that the attacker can acquire smart devices (e.g., smart speakers, smartphones) equipped with the targeted voice assistant. The attacker can query the smart devices for unlimited times, and indicate whether the voice assistant is activated or not. The attacker has speakers to play the generated fuzzy word candidates. Ultimately, the goal of the attacker is to insert the generated \fw into innocent-sounding music or video clips to activate the voice assistant without users noticing, then to issue hidden commands to conduct malicious operations, e.g., upload private conversations to the cloud or open the door.

\section{Generating \FW}\label{sec:search}


\subsection{Framework Overview}
    
In the strict black-box settings, no information about the wake-up word detector is available, thus gradient-based optimization methods cannot be used to generate \fw. Therefore, we resort to heuristic algorithms, which stochastically search for solutions without the gradient information. Among commonly-used heuristic algorithms, genetic algorithm is the most suitable one to solve the problem of fuzzy word generation. Simulated annealing suffers from slow convergence, and it is difficult to apply ant colony optimization (ACO) or particle swarm optimization (PSO) to generate \fw, since ACO deals with problems that can be converted into shortest path finding problems on a graph, while PSO requires the position information in order to move a group of particles in a search-space towards the optimal solutions. Genetic algorithms treat each candidate solution as an individual that contains several chromosomes, and these chromosomes can be mutated and crossed-over to evolve into new individuals. For example, we can regard "alexa" as an individual consisting of chromosomes "a", "l", "e", "x" and "a". If we mutate the chromosome "a" to "i", we get a new individual "alexi", and if we cross "alexa" with "olive" at "e" and "i", we obtain two new individuals "alive" and "olexa".


The key to utilizing genetic algorithm to generate \fw is how to create a diversified initial batch of words that can efficiently evolve into \fw and how to measure whether a word is "good" in terms of its ability to activate the voice assistants and its dissimilarity to the real wake-up word. To tackle these problems, we design the fuzzy word generation framework as follows.

\begin{enumerate}[leftmargin=*]
	\item \textbf{Initialization.} To achieve both diversity and fast convergence, we include three groups of individuals in the initial batch: the real wake-up word itself, words that are similar to the wake-up word (measurement of similarity will be given in \ref{3.3}), and randomly-generated words. Note that we have tried to adopt an entirely random initial population, but found that most random words are non-\fw, and will be killed in the first generation, leaving few to breed useful offspring.   
	\item \textbf{Evaluation.} If we only evaluate an individual in terms of its wake-up rate, the algorithm will end up producing individuals that are almost identical to the wake-up word to achieve high wake-up rate. To prevent this, we formulate the fuzzy word generation as a multi-objective optimization problem, which aims to find \fw that have both high wake-up rate and high dissimilarity distance from the real wake-up word. Instead of simply using weighted sum to combine the two objectives, we leverage the concept of Pareto frontier to select non-dominated individuals~\cite{nsga2}, which preserves as many words as possible to improve diversity of the next generation of descendants.   
	We rank individuals in a non-increasing order according to their wake-up rate and dissimilarity distance respectively, and non-dominated individuals are maintained for reproduction. An individual $x_1$ is dominated by $x_2$ if for all objective functions $f_i(x), \ i=1,...,N$, we have
		\begin{equation}
			f_i(x_1) \le f_i(x_2), \quad \forall i =1,...,N
		\end{equation}
	If there is no individual dominating $x_i$, $x_i$ is said to be non-dominated. In our problem, a word is non-dominated if there is no other word that has both higher wake-up rate and larger dissimilarity distance than the word. 
	
	\item \textbf{Variation}. A new population is created by varying the survived individuals to maintain important pronunciation units and adjust other pronunciation units to find more fuzzy word candidates in the problem space. Commonly-used variation methods include crossover, recombination and mutation.
		
	\end{enumerate}

To customize the generation framework to different languages, there are two aspects that require specific design. First, we need to encode words by determining the number of variables representing each word and the range of each variable, thus an individual can be easily evaluated and transformed into a new individual that represents a valid word. For instance, how to encode "Alexa" or "xi\v ao d\`u xi\v ao d\`u" so that we can perform mutation or cross-over operations to generate new individuals that represent valid English or Chinese words? Secondly, we need to define the dissimilarity distance between two individuals. It is difficult to obtain the dissimilarity distance directly from the encoding, since the encoded vectors consist of numbers that do not carry the pronunciation information of the English or Chinese words. For example, "a", "e" and "f" may be encoded as "1", "5", and "6" respectively in the alphabetic order, but the dissimilarity distance between "e" and "a" are obviously smaller than that between "e" and "f". 

English and Chinese voice assistants cover more than 85\% of the global market~\cite{marketshare}. In particular, Chinese smart speakers have occupied more than 51\% market share in 2019~\cite{chinamarketshare}, but Chinese \fw is less well studied. Mandarin Chinese and English are different as they belong to different language families. Chinese belongs to the Sino-Tibetan language family, while English belongs to the Indo-European language family. Unlike English, Chinese words are not made up with letters as in an alphabetic system, and the pronunciation of Chinese words cannot be inferred directly from the Chinese characters as Chinese is not a phonetic language. Moreover, Chinese and English vary greatly in pronunciation. Chinese is a tone language with four different tones, and pronouncing the same syllable in different tones indeed lead to different meanings. In contrast, English uses stress (rising or falling tones) to emphasize or express emotions, without changing the meaning of a word. In summary, Chinese and English have different word formations, pronunciation units and pronunciation rules, and we need to customize the generation framework for the two languages in appropriate ways. In the following subsection, we first present the design for the Chinese language, which is less well studied, then we present the design for the English language.



\subsection{Generating Chinese \FW}\label{3.3}
A Chinese word is made up of Chinese characters, also known as sinogram or "hanzi". Chinese characters are very different from English letters. Each Chinese character is both the smallest meaning unit and the smallest pronunciation unit. The pronunciation of a Chinese character is represented by \emph{pinyin}. The pinyin of a character consists of initials (e.g., x), finals (e.g., ai or iao) also known as vowels, and a tone. The pronunciation of a Chinese character is determined by its initial, final and tone. The pronunciation of a Chinese word is the combination of the pronunciation of each character. There is at most one initial in the pinyin of a character. Some characters do not have an initial, e.g., \`ai (爱). There are 23 initials in total~\cite{Pinyin-initial-final-full, Pinyin-initial-final}. There are one to two finals in the pinyin of a character. Some finals can be combined together, e.g., i and ao form iao, but some finals can not, e.g., i and ou. By considering valid final combinations as special finals, we have a total of 37 finals~\cite{Pinyin-initial-final-full, Pinyin-initial-final}. There are four tones in Chinese, denoted by a diacritical mark on the finals, i.e., \=a, \'a, \v a, \`a for tone $1, 2, 3, 4$ respectively. The same initial-final combination with different tones have different meanings, e.g., xi\v ao (小) and xi\`ao (笑) mean "small" and "laugh" respectively. Some initial-final combinations are invalid (cannot be pronounced), e.g., xang, no matter what the tone is. Some initial-final-tone combinations have no corresponding Chinese characters, e.g., jīng (精) is valid but there is no Chinese character that pronounces as j\'ing (the second tone). Such invalid combinations need to be culled during the evolution in the genetic algorithm.

\textbf{Encoding}. Since the pinyin of a Chinese character normally comprises of the initial, the final and the tone, we use three variables to encode a character. The range of each variable is the number of possible initials/finals/tones, which are 24 (23 initials and zero-initial), 37 and 4 respectively. Without loss of generality, we use the lexicographic order of initials/finals/tones~\cite{Pinyin-initial-final-full} as the value of the variable. As far as we know, all Chinese wake-up words are composed of four characters. Hence, we encode each individual as a 12-dimension vector. 
	
\textbf{Dissimilarity distance}. We cannot use the difference between encodings of two words to represent their dissimilarity, since the lexicographic order of initials/finals/tones does not reflect their pronunciation resemblance, e.g., "a", "o" and "an" are encoded as 1, 2 and 8 respectively, but "a" pronounces more closely to "an" than to "o". To capture the phonetic similarity between initials and finals, we leverage the high-dimensional embedding~\cite{dimsim}. Let $W_1=[c^{(1)}_1,...,$ $c^{(n)}_1]$ and $W_2=[c^{(1)}_2,..., c^{(n)}_2]$ denote two Chinese words, where $c^{(i)}$ is the $i$-th character of a word. The dissimilarity distance is calculated as $\operatorname{dist}(W_1, W_2)=\sum_i \operatorname{dist}(c^{(i)}_1, c^{(i)}_2)$, where $\operatorname{dist}(c^{(i)}_1, c^{(i)}_2)$ is the distance between two characters at the same offset. The distance calculated in this way increases with the number of characters, thus we normalize the distance to $[0,1)$ using $\operatorname{tanh}(\cdot)$, a commonly-used sigmoidal function that normalizes the activation of neural networks~\cite{tanh}. 
	
	\begin{equation}
	    \operatorname{dist}(W_1, W_2) = \frac{1}{n}\sum_{i} \operatorname{tanh}(\operatorname{dist}(c^{(i)}_1, c^{(i)}_2)/A).
	\end{equation}
The distance is divided by constant $A$ to attain a more evenly distribution. In our experiment, we set $A=100$.
	
  	\begin{table*}[t]
    	    \centering
    	    \caption{Generated \fw for Chinese and English smart speakers. We show the top-ten fuzzy word examples with the highest dissimilarity distance. }
    	    \vspace{-0.15in}
    	    \scalebox{0.75}{
        	\begin{tabular}{c|c c c c|c c c c}
            \toprule[1.5pt]
              \ &Baidu & Xiaomi & AliGenie & Tencent & Amazon Echo & Echo Dot & Google & Apple Siri 
               \\
            \midrule  
            Wake-up word & \makecell[c]{xi\v ao d\`u xi\v ao d\`u} & 
            \makecell[c]{xi\v ao \`ai t\'ong xu\'e} & \makecell[c]{ti\=an m\=ao jīng l\'ing} & 
            \makecell[c]{jiǔ s\`i \`er l\' ing} & Alexa & Alexa & Hey Google & Hey Siri  \\
            \midrule
            Total number & 63 & 108 & 322
            & 84 & 127 & 130 & 79 & 52 \\
            Mean dissimilarity & 5.35\% & 6.26\% & 2.37\%
            & 2.66\% & 15.28\% & 15.29\% & 11.98\% & 9.92\% \\
            
            \midrule \midrule 
            & \makecell[c]{xiǎo l\v ong xiǎo l\v ong} 
            & \makecell[c]{qiǎo bāi dōng h\`e} 
            & \makecell[c]{yān m\=en jīng l\'ing}
            & \makecell[c]{jiōng ni\`ao \`er l\'ing} 
            & \makecell[c]{ilebser} 
            & \makecell[c]{ureqssr} 
            & \makecell[c]{heii googerl} 
            & \makecell[c]{hey sserea} 
            \\
            
            & \cellcolor{gray!10}20\% 
            & \cellcolor{gray!10}90\% 
            & \cellcolor{gray!10}50\%
            & \cellcolor{gray!10}30\% 
            & \cellcolor{gray!10}70\%  
            & \cellcolor{gray!10}10\%  
            & \cellcolor{gray!10}60\%  
            & \cellcolor{gray!10}80\% \\

            & \makecell[c]{piǎo d\`ou piǎo d\`ou} 
            & \makecell[c]{qiāng bāi dōng s\`e} 
            & \makecell[c]{yān m\=ang jīng l\'ing}
            & \makecell[c]{jǐn sì ào líng} 
            & \makecell[c]{ileqsur} 
            & \makecell[c]{arleqsr} 
            & \makecell[c]{heiigoogaa} 
            & \makecell[c]{heai ssuree} 
            \\
            
            & \cellcolor{gray!10}10\% 
            & \cellcolor{gray!10}100\% 
            & \cellcolor{gray!10}100\%
            & \cellcolor{gray!10}10\% 
            & \cellcolor{gray!10}60\%  
            & \cellcolor{gray!10}100\%  
            & \cellcolor{gray!10}30\% 
            & \cellcolor{gray!10}50\%  \\
            
            
            & \makecell[c]{tiǎo dōu tiǎo dōu} 
            & \makecell[c]{qiào bāi tōu shè} 
            & \makecell[c]{w\'an m\=ang jīng l\'ing}
            & \makecell[c]{jiōng sì èr lián} 
            & \makecell[c]{ileqcer} 
            & \makecell[c]{ilekcer} 
            & \makecell[c]{heay gugal} 
            & \makecell[c]{hay scir e} 
            \\

            & \cellcolor{gray!10}60\% 
            & \cellcolor{gray!10}90\% 
            & \cellcolor{gray!10}40\%
            & \cellcolor{gray!10}10\% 
            & \cellcolor{gray!10}90\% 
            & \cellcolor{gray!10}100\%  
            & \cellcolor{gray!10}100\%  
            & \cellcolor{gray!10}60\%  \\
            
            & \makecell[c]{tiǎo dòng tiǎo dòng} 
            & \makecell[c]{qiāo cāi dōng sè} 
            & \makecell[c]{wáng mào jǐng lǐn}
            & \makecell[c]{jiǒng shì èr lián} 
            & \makecell[c]{ilekcer} 
            & \makecell[c]{ilexcer} 
            & \makecell[c]{hey gooogov a} 
            & \makecell[c]{haiiasciree} 
            \\
            
            & \cellcolor{gray!10}20\% 
            & \cellcolor{gray!10}100\% 
            & \cellcolor{gray!10}10\%
            & \cellcolor{gray!10}20\% 
            & \cellcolor{gray!10}70\%  
            & \cellcolor{gray!10}10\%  
            & \cellcolor{gray!10}100\%  
            & \cellcolor{gray!10}60\% \\
            
            Top-ten examples
            & \makecell[c]{shāo dōu shāo dōu} 
            & \makecell[c]{xiāo cāi dōng sè} 
            & \makecell[c]{yán māo jīng lǐng}
            & \makecell[c]{jiǒng sì èr lián} 
            & \makecell[c]{ileqsar} 
            & \makecell[c]{ilexsur} 
            & \makecell[c]{heii googurl} 
            & \makecell[c]{heyisyree} 
            \\
            
            \cellcolor{gray!10} wake-up rate 
            & \cellcolor{gray!10}20\% 
            & \cellcolor{gray!10}100\% 
            & \cellcolor{gray!10}90\%
            & \cellcolor{gray!10}100\% 
            & \cellcolor{gray!10}100\%  
            & \cellcolor{gray!10}100\% 
            & \cellcolor{gray!10}70\%  
            & \cellcolor{gray!10}20\%  \\
            
            & \makecell[c]{jiǎo dōu jiǎo dōu} 
            & \makecell[c]{qiāo ē dū sè} 
            & \makecell[c]{yān māo jīng líng}
            & \makecell[c]{jiǒng sì è rliáng} 
            & \makecell[c]{ilexsar} 
            & \makecell[c]{ileqsar} 
            & \makecell[c]{heii gugurl} 
            & \makecell[c]{heii sirea} 
            \\
            
            & \cellcolor{gray!10}40\% 
            & \cellcolor{gray!10}100\% 
            & \cellcolor{gray!10}100\%
            & \cellcolor{gray!10}30\% 
            & \cellcolor{gray!10}100\%  
            & \cellcolor{gray!10}100\% 
            & \cellcolor{gray!10}60\%  
            & \cellcolor{gray!10}70\%  \\
            
            & \makecell[c]{qiáo dōu qiáo dōu} 
            & \makecell[c]{qiāo āi dū sè} 
            & \makecell[c]{wān māo jīng líng}
            & \makecell[c]{jiǒng zì èr liáo} 
            & \makecell[c]{ilexsur} 
            & \makecell[c]{ileksar} 
            & \makecell[c]{hea gougll} 
            & \makecell[c]{haiy cire} 
            \\
            
            & \cellcolor{gray!10}30\% 
            & \cellcolor{gray!10}100\% 
            & \cellcolor{gray!10}90\%
            & \cellcolor{gray!10}90\% 
            & \cellcolor{gray!10}100\%  
            & \cellcolor{gray!10}100\% 
            & \cellcolor{gray!10}50\%  
            & \cellcolor{gray!10}80\%  \\
            
            & \makecell[c]{qiǎo dōu qiǎo dōu} 
            & \makecell[c]{qiàng āi dōng sàng} 
            & \makecell[c]{yán māo jīng líng}
            & \makecell[c]{jiōng sì èr mín} 
            & \makecell[c]{ileksur} 
            & \makecell[c]{ileksur} 
            & \makecell[c]{hei googll a} 
            & \makecell[c]{hey sirr e} 
            \\
            
            & \cellcolor{gray!10}50\% 
            & \cellcolor{gray!10}100\% 
            & \cellcolor{gray!10}90\%
            & \cellcolor{gray!10}10\% 
            & \cellcolor{gray!10}80\%  
            & \cellcolor{gray!10}100\% 
            & \cellcolor{gray!10}100\%  
            & \cellcolor{gray!10}100\%  \\
            
            & \makecell[c]{qiǎo dòng qiǎo dòng} 
            & \makecell[c]{qiàng āi chōu lè} 
            & \makecell[c]{wán māo jīng líng}
            & \makecell[c]{jiǒng sì èr mín} 
            & \makecell[c]{alexoer} 
            & \makecell[c]{alekcir} 
            & \makecell[c]{hei gooo r} 
            & \makecell[c]{hei suru r a} 
            \\
            
            & \cellcolor{gray!10}20\% 
            & \cellcolor{gray!10}100\% 
            & \cellcolor{gray!10}20\%
            & \cellcolor{gray!10}30\% 
            & \cellcolor{gray!10}60\%  
            & \cellcolor{gray!10}70\% 
            & \cellcolor{gray!10}20\%  
            & \cellcolor{gray!10}70\% \\
            
            & \makecell[c]{xiāo dōu xiāo dōu} 
            & \makecell[c]{qiǎo ā dǒng sà} 
            & \makecell[c]{wáng māo jīng líng}
            & \makecell[c]{jiǒng sì er liāo} 
            & \makecell[c]{ilexcer} 
            & \makecell[c]{ileqser} 
            & \makecell[c]{heiy googow l} 
            & \makecell[c]{hay syrrie e} 
            \\
            
            & \cellcolor{gray!10}50\% 
            & \cellcolor{gray!10}100\% 
            & \cellcolor{gray!10}10\%
            & \cellcolor{gray!10}30\% 
            & \cellcolor{gray!10}100\%  
            & \cellcolor{gray!10}100\% 
            & \cellcolor{gray!10}50\%  
            & \cellcolor{gray!10}100\% \\
          
            \bottomrule[1.5pt]
            \end{tabular}}
            
            \label{tab: fuzzy result}%
        \end{table*}

        \begin{figure*}[t]
    \centering
    	\begin{minipage}[b]{0.97\linewidth}
    		\centering
    		\includegraphics[trim=0mm 0mm 0mm 0mm, clip, width=0.3\textwidth]{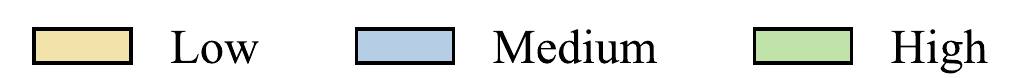}
    		\vspace{-0.2cm}
    	\end{minipage}
    \subfigure[Baidu: Volume]{
    	\begin{minipage}[b]{0.3\linewidth}
    		\centering
    		\includegraphics[trim=0mm 0mm 0mm 10mm, clip, width=0.95\textwidth]{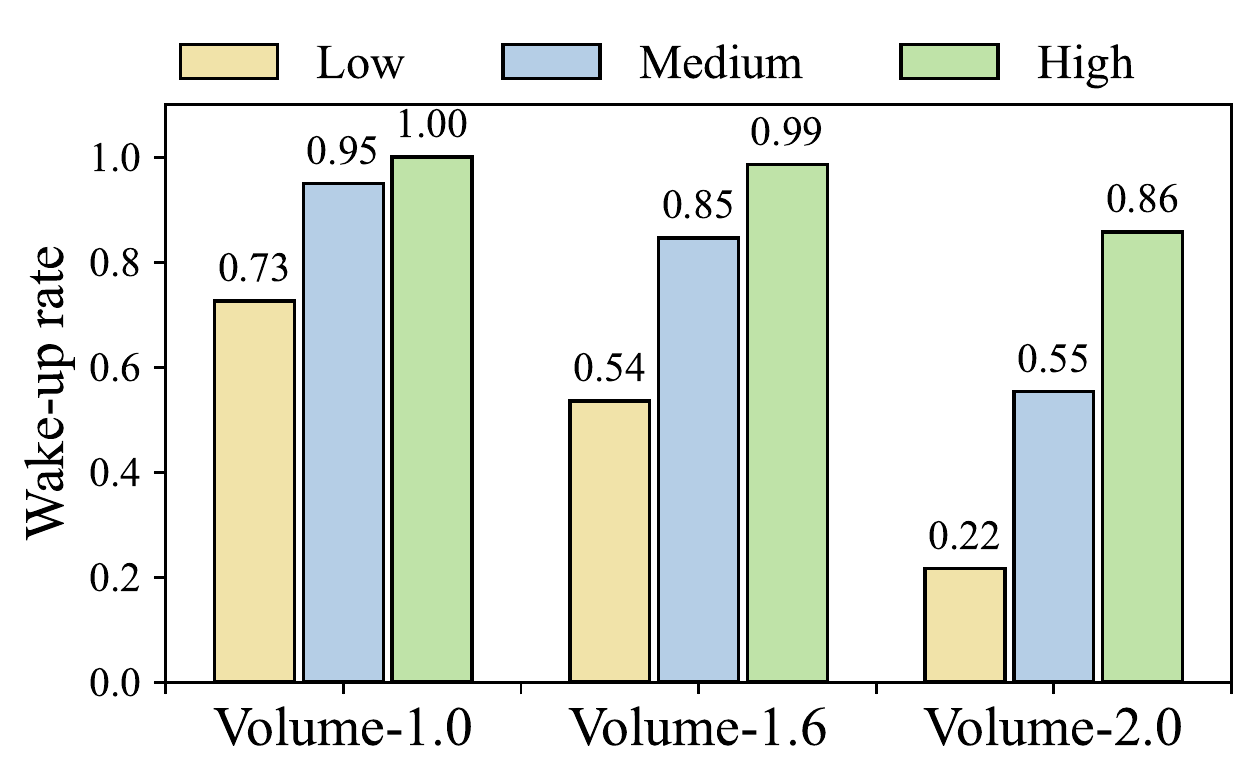}
    		\label{fig:ver_2figs_2cap_1}
    		\vspace{-0.2cm}
    	\end{minipage}
    }
    \vspace{-0.1cm}
    \subfigure[Baidu: Speed]{
    	\begin{minipage}[b]{0.3\linewidth}
    		\centering
    		\includegraphics[trim=0mm 0mm 0mm 10mm, clip,width=0.95\textwidth]{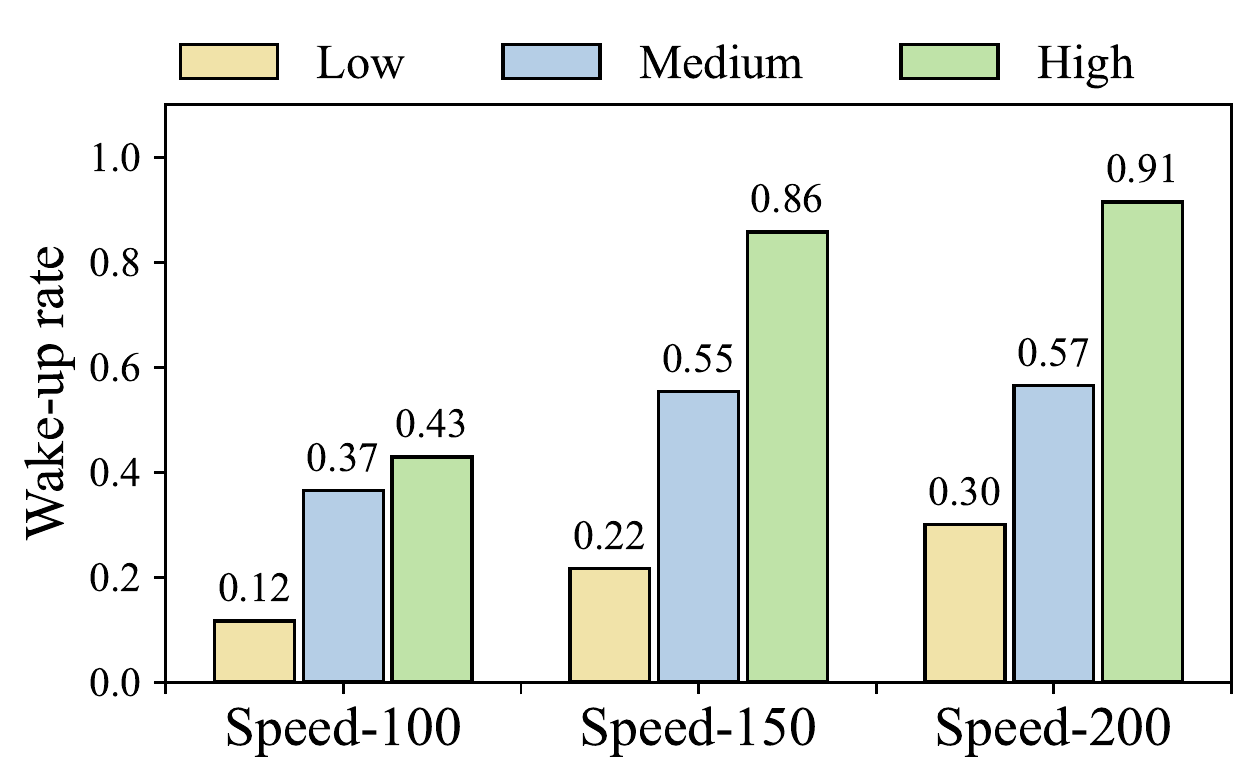}
    		\label{fig:ver_2figs_2cap_2}
    		\vspace{-0.2cm}
    	\end{minipage}
    }
    \vspace{-0.1cm}
    \subfigure[Baidu: Noise]{
    	\begin{minipage}[b]{0.3\linewidth}
    		\centering
    		\includegraphics[trim=0mm 0mm 0mm 10mm, clip,width=0.95\textwidth]{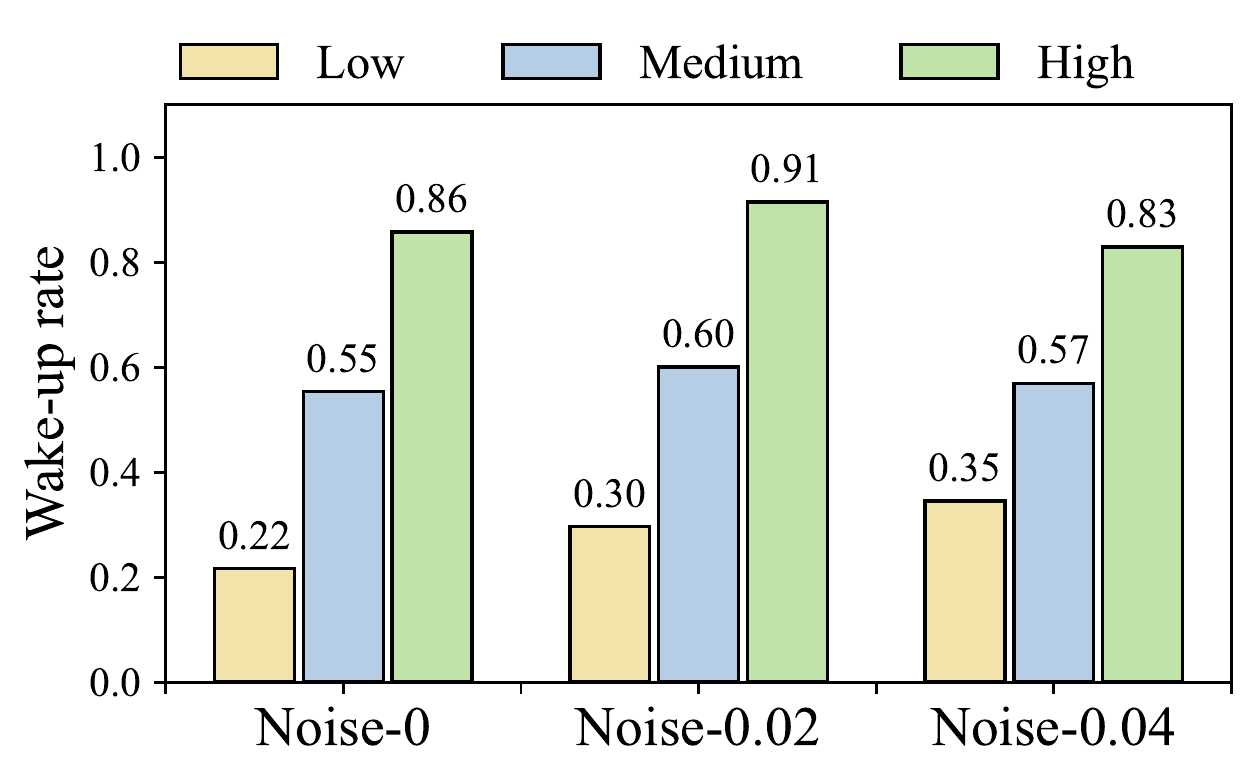}
    		\label{fig:ver_2figs_2cap_2}
    		\vspace{-0.2cm}
    	\end{minipage}
    }
    \vspace{-0.1cm}
    \subfigure[Xiaomi: Volume]{
    	\begin{minipage}[b]{0.3\linewidth}
    		\centering
    		\includegraphics[trim=0mm 0mm 0mm 10mm, clip, width=0.95\textwidth]{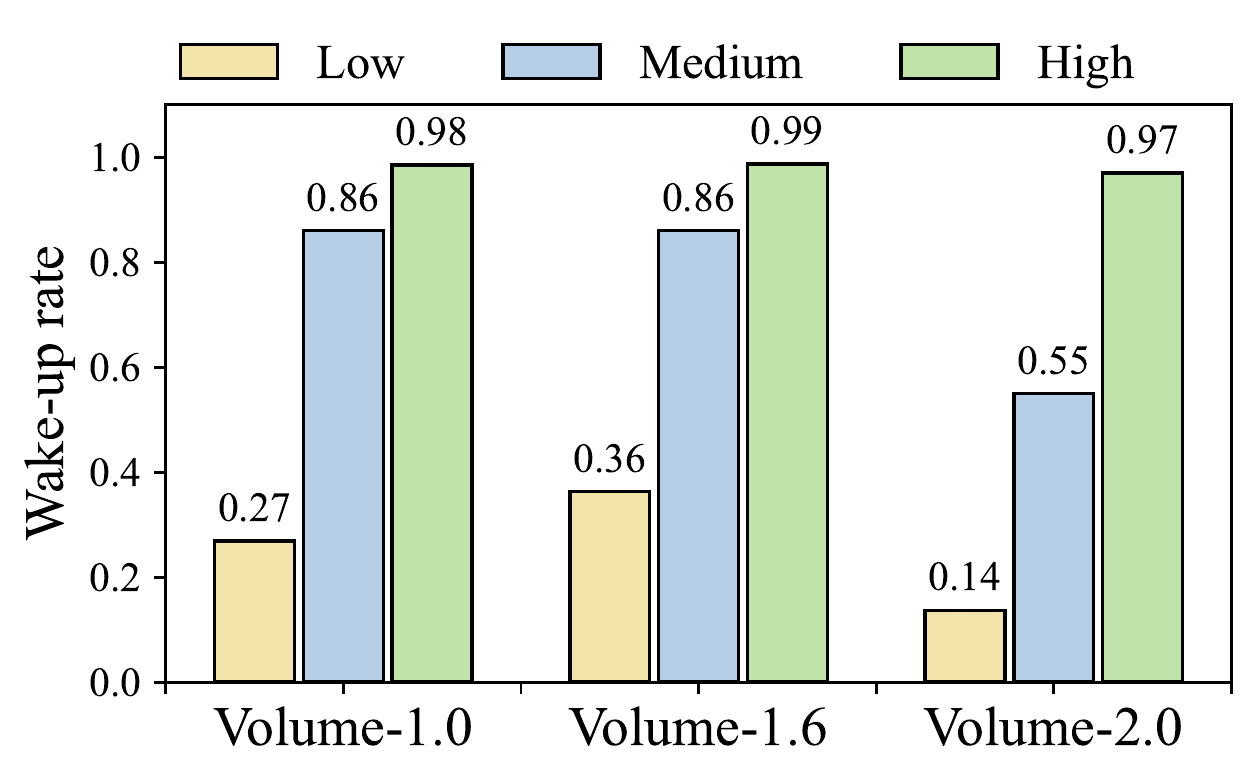}
    		\label{fig:ver_2figs_2cap_1}
    		\vspace{-0.2cm}
    	\end{minipage}
    }
    \vspace{-0.1cm}
    \subfigure[Xiaomi: Speed]{
    	\begin{minipage}[b]{0.3\linewidth}
    		\centering
    		\includegraphics[trim=0mm 0mm 0mm 10mm, clip,width=0.95\textwidth]{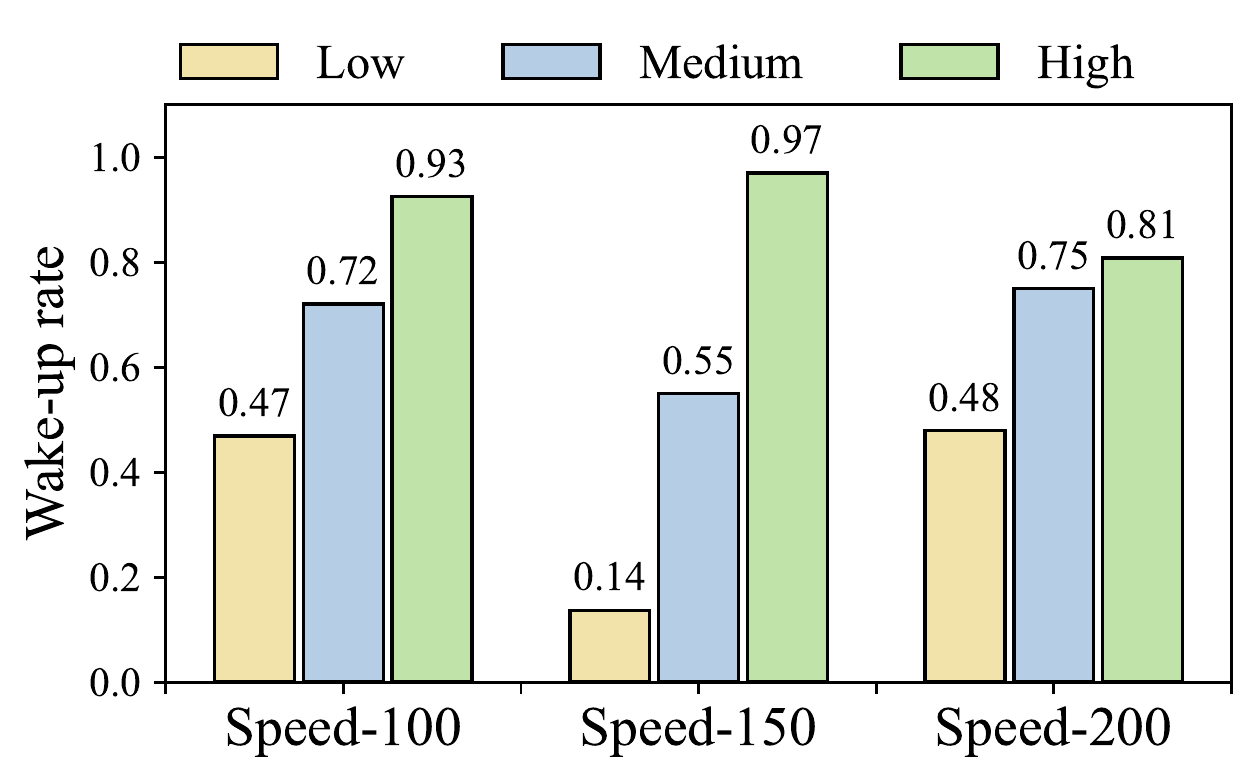}
    		\label{fig:ver_2figs_2cap_2}
    		\vspace{-0.2cm}
    	\end{minipage}
    }
    \vspace{-0.1cm}
    \subfigure[Xiaomi: Noise]{
    	\begin{minipage}[b]{0.3\linewidth}
    		\centering
    		\includegraphics[trim=0mm 0mm 0mm 10mm, clip,width=0.95\textwidth]{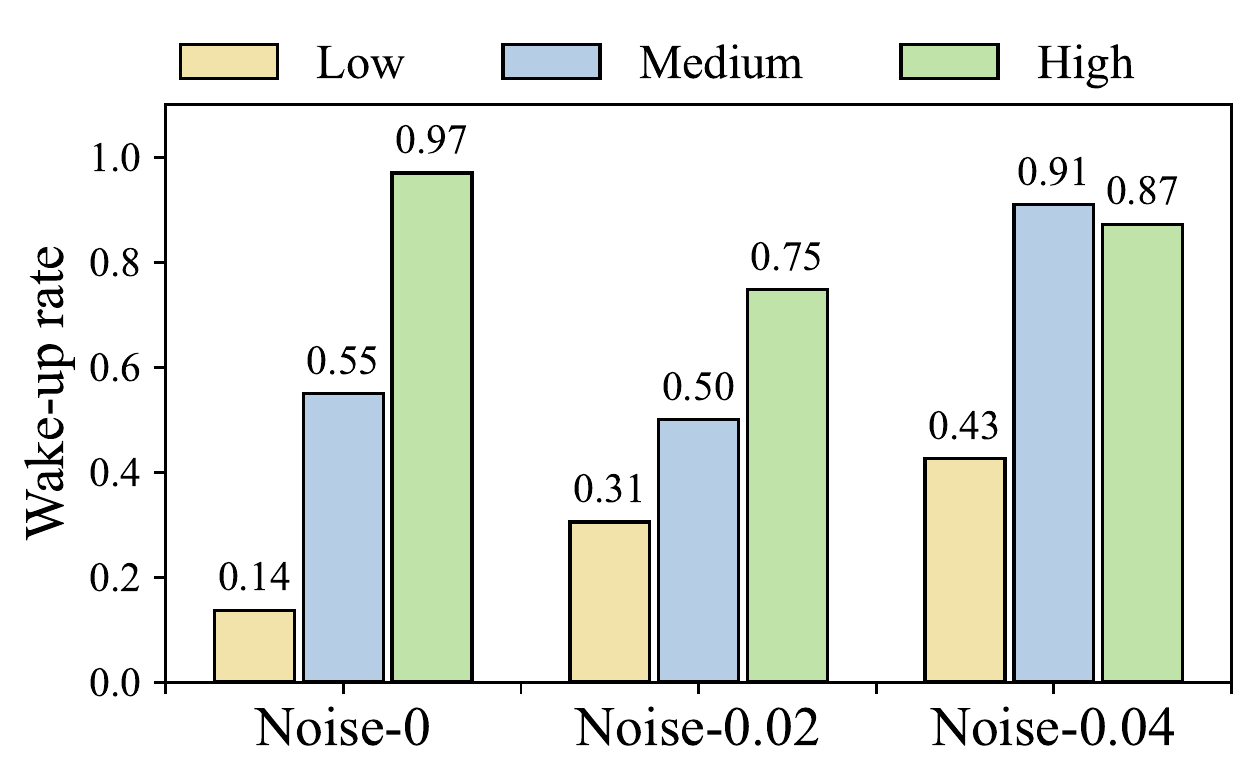}
    		\label{fig:ver_2figs_2cap_2}
    		\vspace{-0.2cm}
    	\end{minipage}
    }
    \vspace{-0.1cm}
    
    \subfigure[Alibaba: Volume]{
    	\begin{minipage}[b]{0.3\linewidth}
    		\centering
    		\includegraphics[trim=0mm 0mm 0mm 10mm, clip, width=0.95\textwidth]{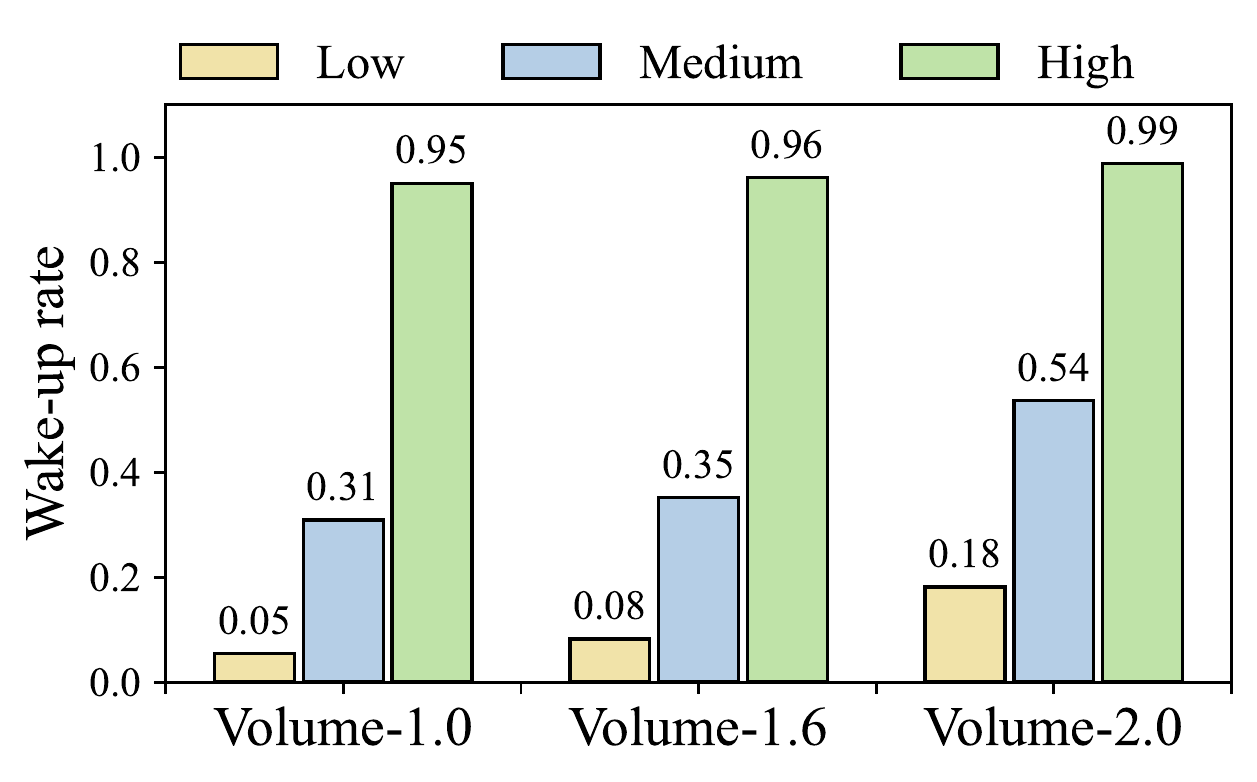}
    		\label{fig:ver_2figs_2cap_1}
    		\vspace{-0.2cm}
    	\end{minipage}
    }
    \vspace{-0.1cm}
    \subfigure[Alibaba: Speed]{
    	\begin{minipage}[b]{0.3\linewidth}
    		\centering
    		\includegraphics[trim=0mm 0mm 0mm 10mm, clip,width=0.95\textwidth]{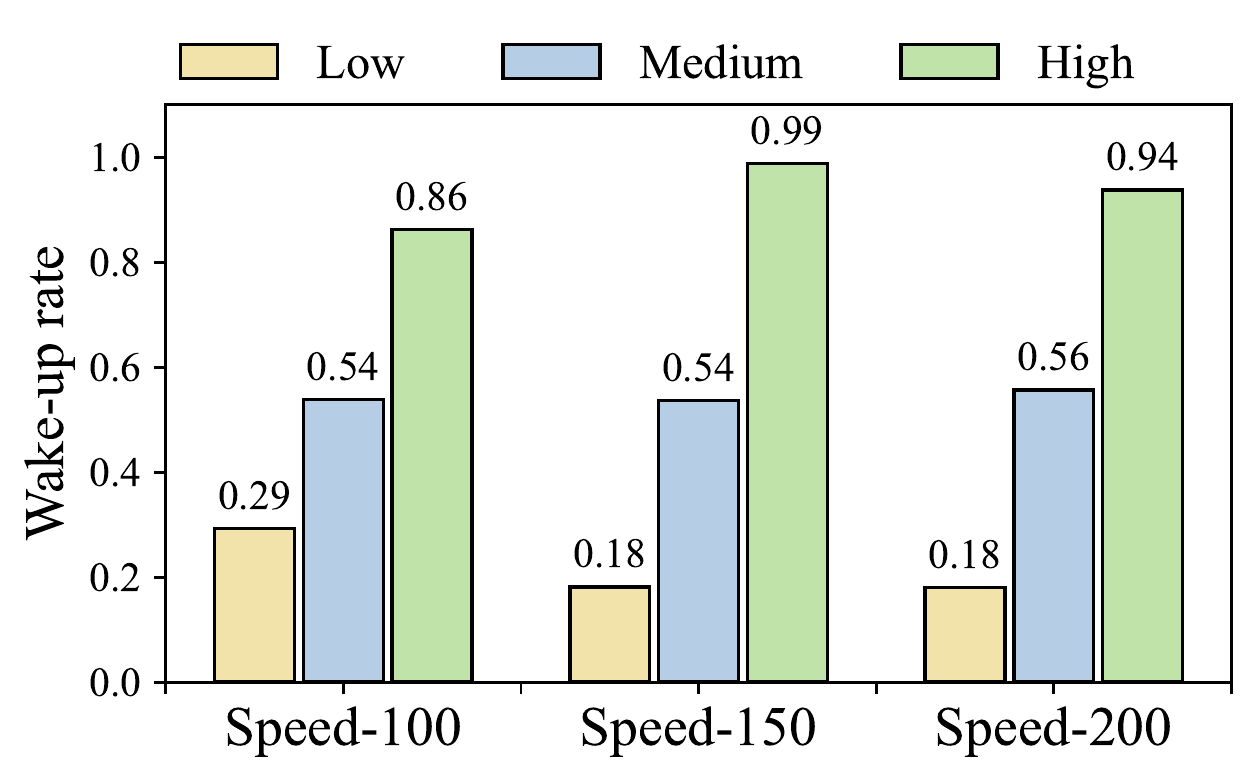}
    		\label{fig:ver_2figs_2cap_2}
    		\vspace{-0.2cm}
    	\end{minipage}
    }
    \vspace{-0.1cm}
    \subfigure[Alibaba: Noise]{
    	\begin{minipage}[b]{0.3\linewidth}
    		\centering
    		\includegraphics[trim=0mm 0mm 0mm 10mm, clip,width=0.95\textwidth]{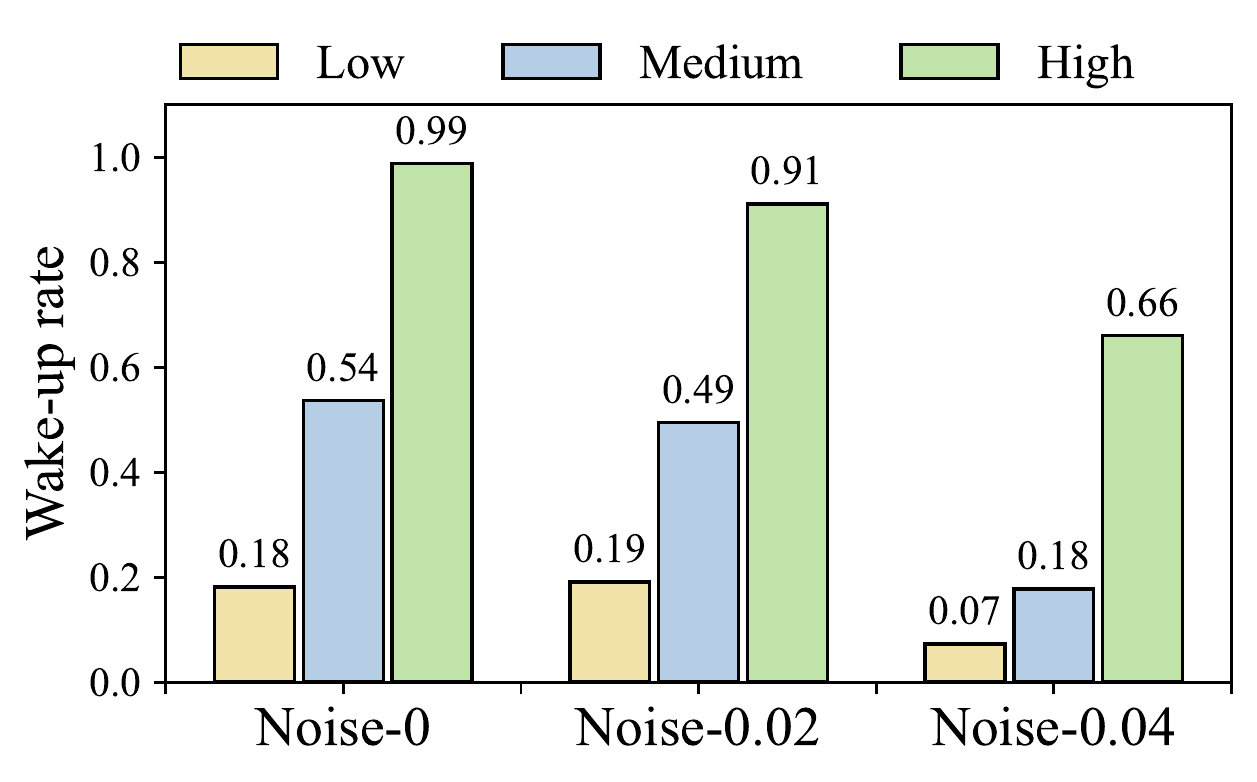}
    		\label{fig:ver_2figs_2cap_2}
    		\vspace{-0.2cm}
    	\end{minipage}
    }
    \vspace{-0.1cm}
    
    \subfigure[Tencent: Volume]{
    	\begin{minipage}[b]{0.3\linewidth}
    		\centering
    		\includegraphics[trim=0mm 0mm 0mm 10mm, clip, width=0.95\textwidth]{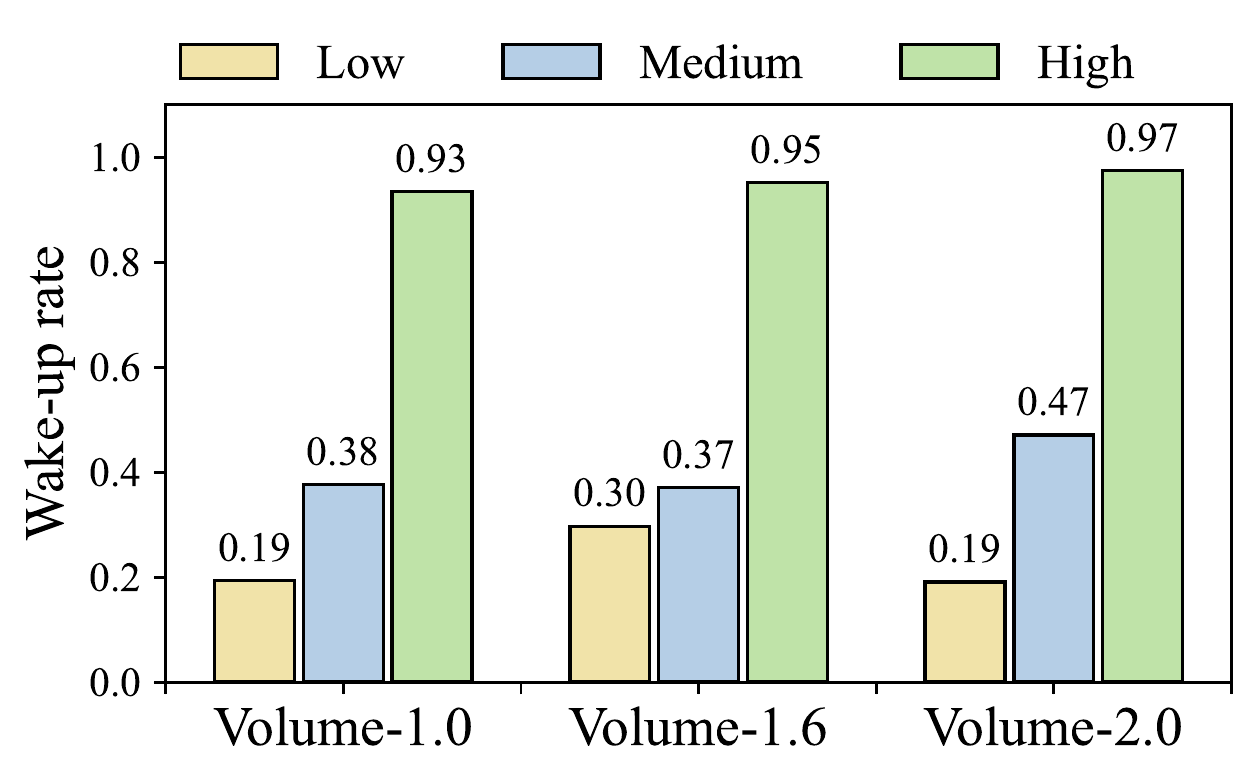}
    		\label{fig:ver_2figs_2cap_1}
    		\vspace{-0.2cm}
    	\end{minipage}
    }
    \vspace{-0.1cm}
    \subfigure[Tencent: Speed]{
    	\begin{minipage}[b]{0.3\linewidth}
    		\centering
    		\includegraphics[trim=0mm 0mm 0mm 10mm, clip,width=0.95\textwidth]{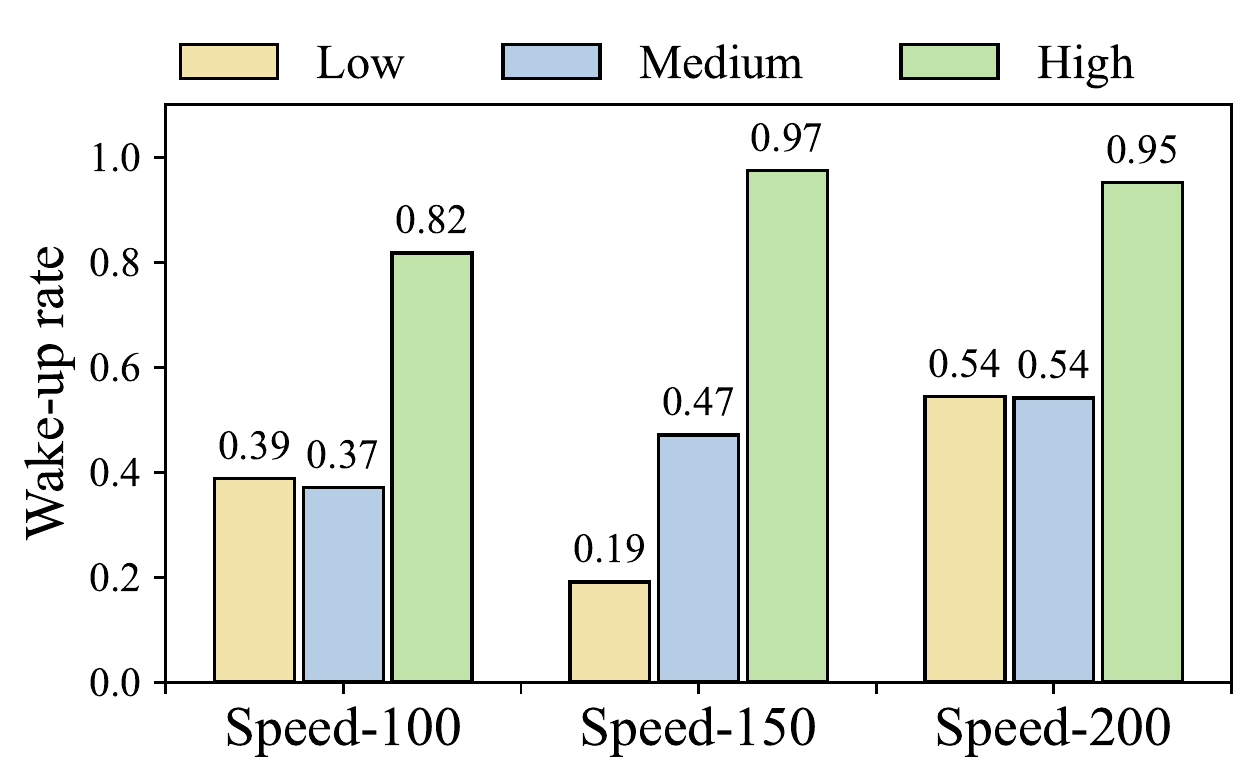}
    		\label{fig:ver_2figs_2cap_2}
    		\vspace{-0.2cm}
    	\end{minipage}
    }
    \vspace{-0.1cm}
    \subfigure[Tencent: Noise]{
    	\begin{minipage}[b]{0.3\linewidth}
    		\centering
    		\includegraphics[trim=0mm 0mm 0mm 10mm, clip,width=0.95\textwidth]{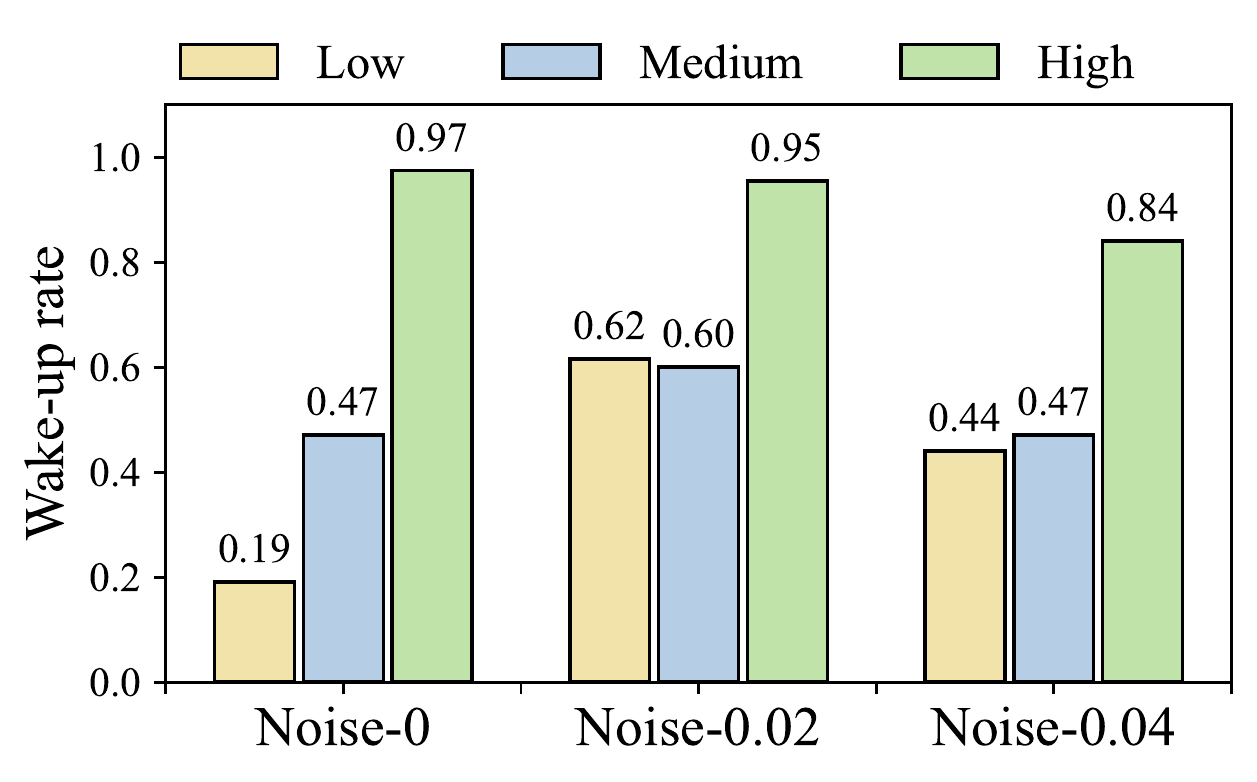}
    		\label{fig:ver_2figs_2cap_2}
    		\vspace{-0.2cm}
    	\end{minipage}
    }
    \vspace{-0.1cm}
	\caption{Wake-up rate of \fw under different environments for Chinese voice assistants.}
	\vspace{-0.15in}
	\label{fig:Ablation_results_Chinese}
\end{figure*}

        	\begin{figure}[tt]
    \centering
    \includegraphics[width=0.4\linewidth]{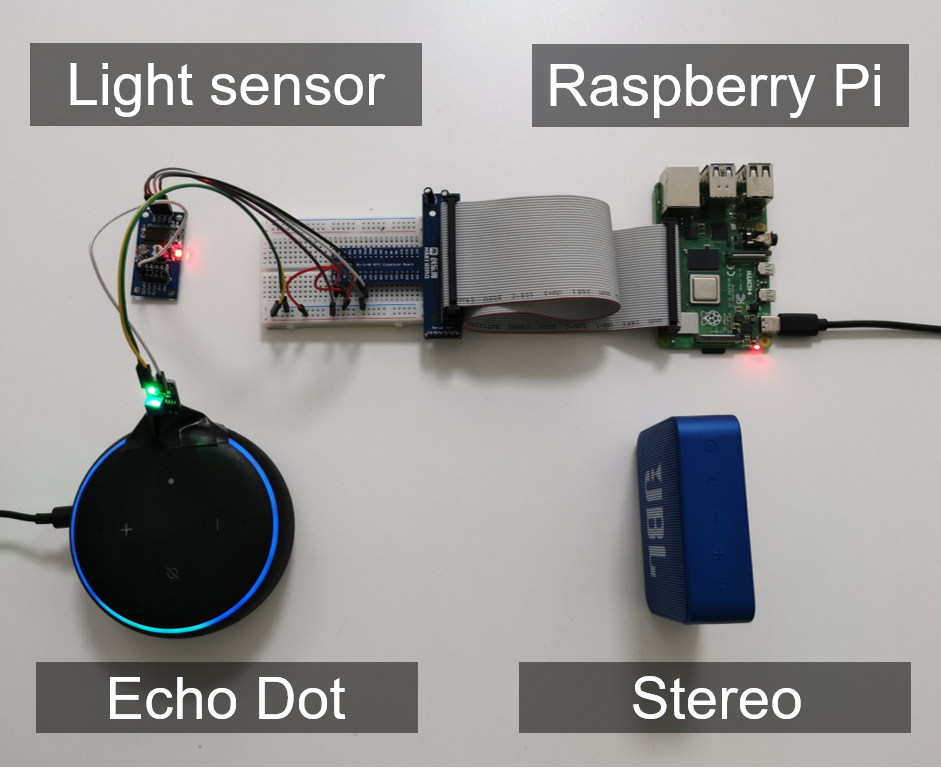}
    \caption{The experiment setup includes a laptop, a stereo, a Raspberry Pi, and a light sensor. The laptop runs the generation algorithm, and the stereo plays the generated \fw over the air to the smart speaker. The Raspberry Pi equipped with the light sensor detects the activation of the smart speaker, which is fed back to the laptop to calculate the wake-up rate. }
    \label{exsetup}
    \vspace{-0.2in}
    \end{figure}

\begin{figure*}[t]
    \centering
    	\begin{minipage}[b]{0.97\linewidth}
    		\centering
    		\includegraphics[trim=0mm 0mm 0mm 0mm, clip, width=0.3\textwidth]{Picture/Ablation_new/legend.pdf}
    		\vspace{-0.2cm}
    	\end{minipage}
    \subfigure[Amazon Echo: Volume]{
    	\begin{minipage}[b]{0.3\linewidth}
    		\centering
    		\includegraphics[trim=0mm 0mm 0mm 10mm, clip, width=0.95\textwidth]{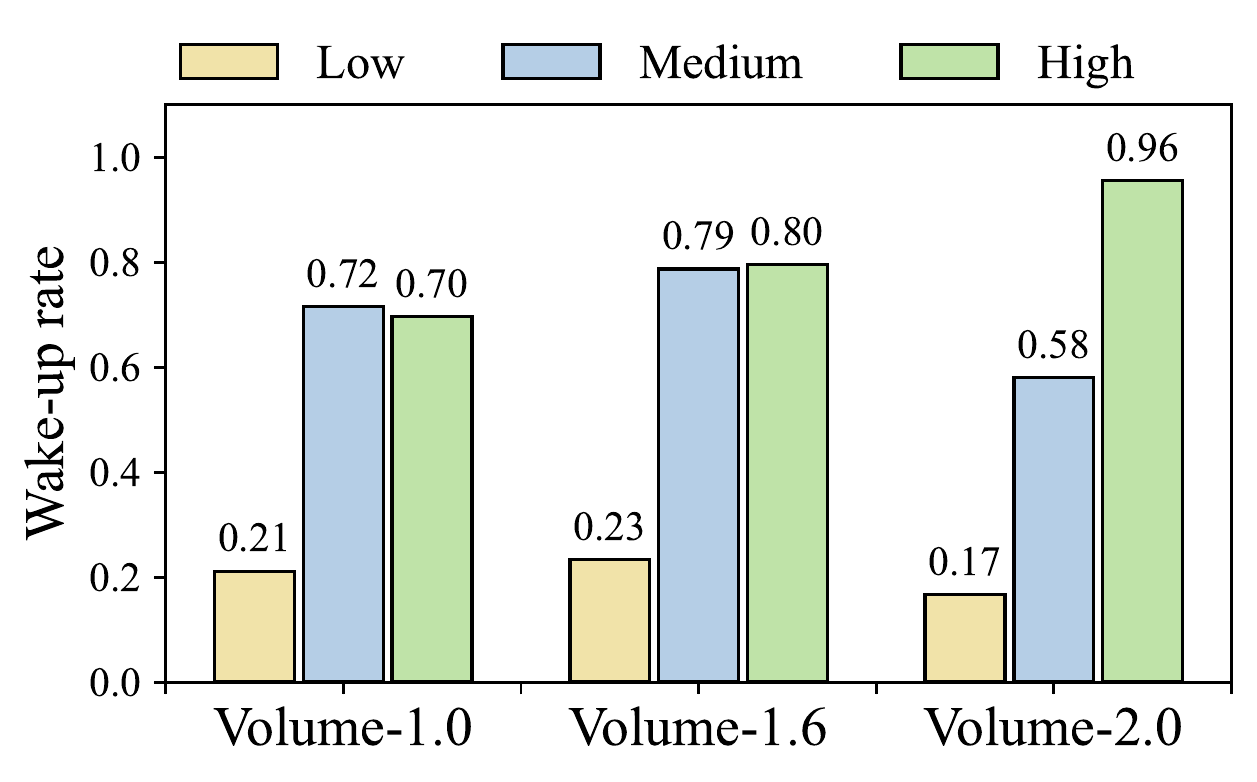}
    		\label{fig:ver_2figs_2cap_1}
    		\vspace{-0.2cm}
    	\end{minipage}
    }
    \vspace{-0.1cm}
    \subfigure[Amazon Echo: Speed]{
    	\begin{minipage}[b]{0.3\linewidth}
    		\centering
    		\includegraphics[trim=0mm 0mm 0mm 10mm, clip,width=0.95\textwidth]{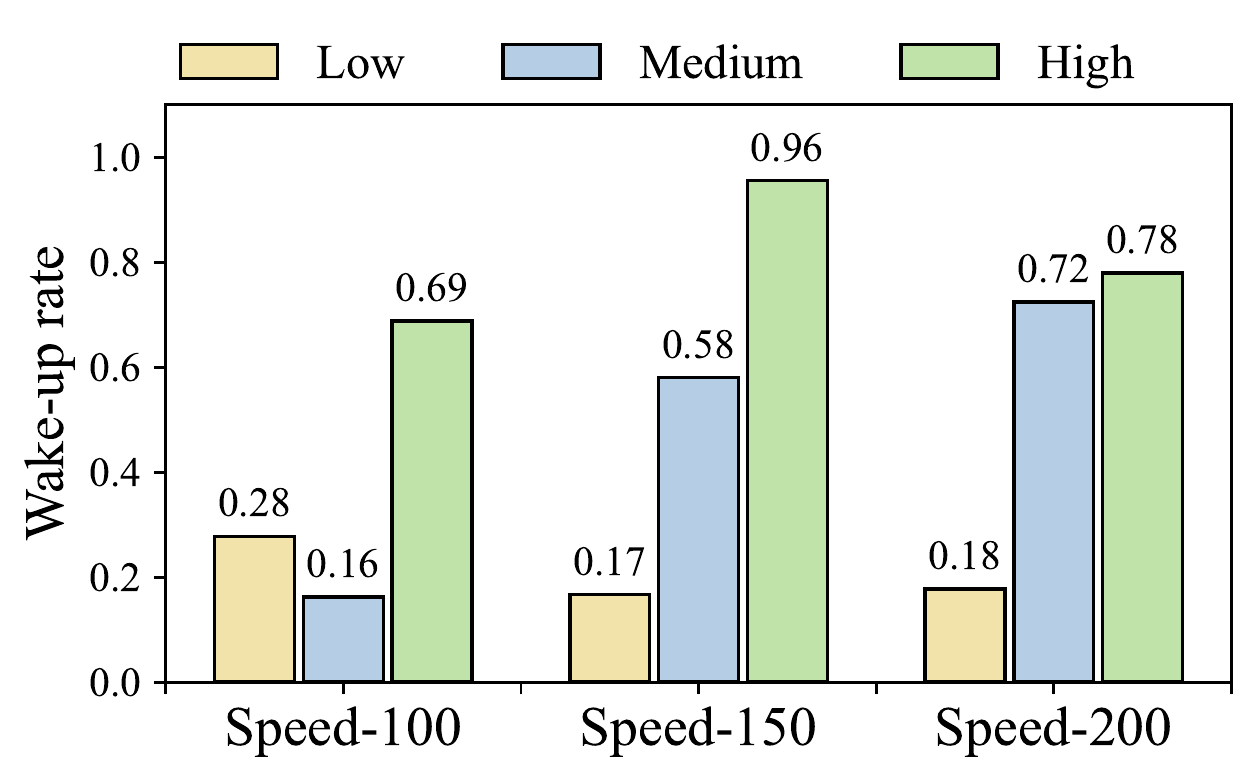}
    		\label{fig:ver_2figs_2cap_2}
    		\vspace{-0.2cm}
    	\end{minipage}
    }
    \vspace{-0.1cm}
    \subfigure[Amazon Echo: Noise]{
    	\begin{minipage}[b]{0.3\linewidth}
    		\centering
    		\includegraphics[trim=0mm 0mm 0mm 10mm, clip,width=0.95\textwidth]{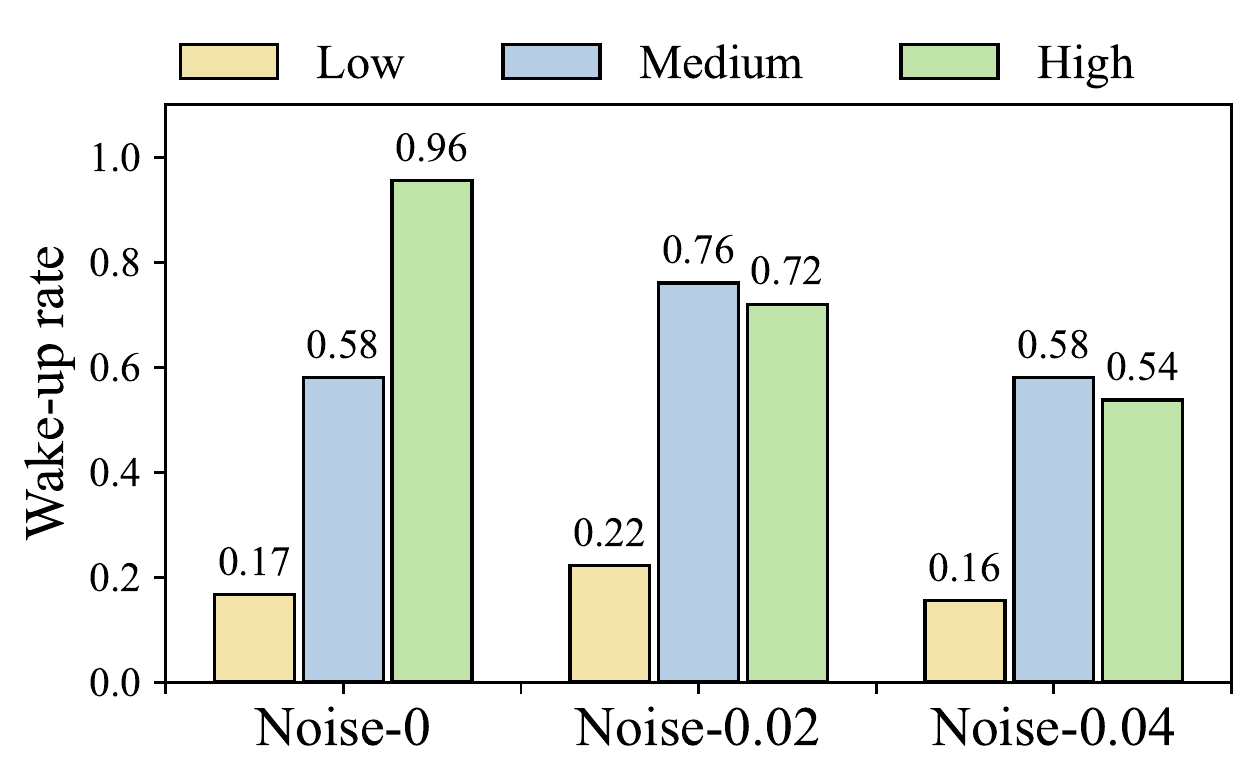}
    		\label{fig:ver_2figs_2cap_2}
    		\vspace{-0.2cm}
    	\end{minipage}
    }
    \vspace{-0.1cm}
    \subfigure[Echo Dot: Volume]{
    	\begin{minipage}[b]{0.3\linewidth}
    		\centering
    		\includegraphics[trim=0mm 0mm 0mm 10mm, clip, width=0.95\textwidth]{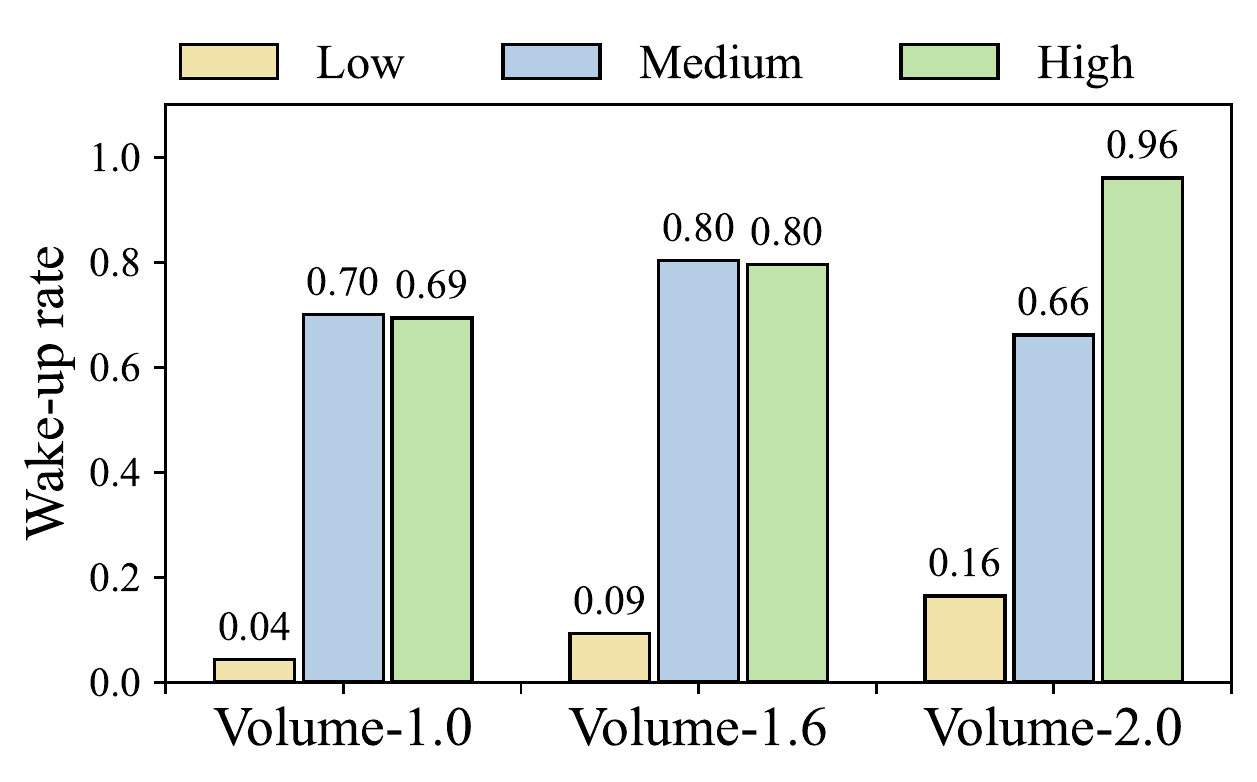}
    		\label{fig:ver_2figs_2cap_1}
    		\vspace{-0.2cm}
    	\end{minipage}
    }
    \vspace{-0.1cm}
    \subfigure[Echo Dot: Speed]{
    	\begin{minipage}[b]{0.3\linewidth}
    		\centering
    		\includegraphics[trim=0mm 0mm 0mm 10mm, clip,width=0.95\textwidth]{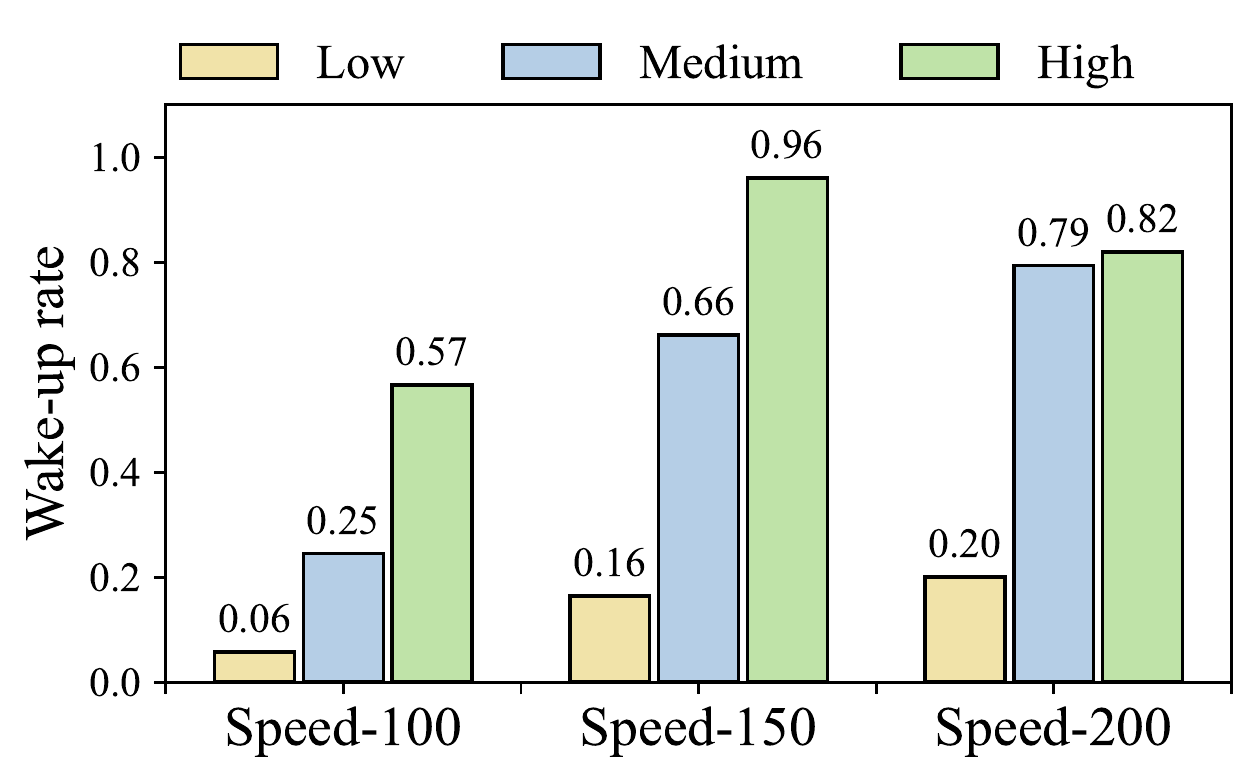}
    		\label{fig:ver_2figs_2cap_2}
    		\vspace{-0.2cm}
    	\end{minipage}
    }
    \vspace{-0.1cm}
    \subfigure[Echo Dot: Noise]{
    	\begin{minipage}[b]{0.3\linewidth}
    		\centering
    		\includegraphics[trim=0mm 0mm 0mm 10mm, clip,width=0.95\textwidth]{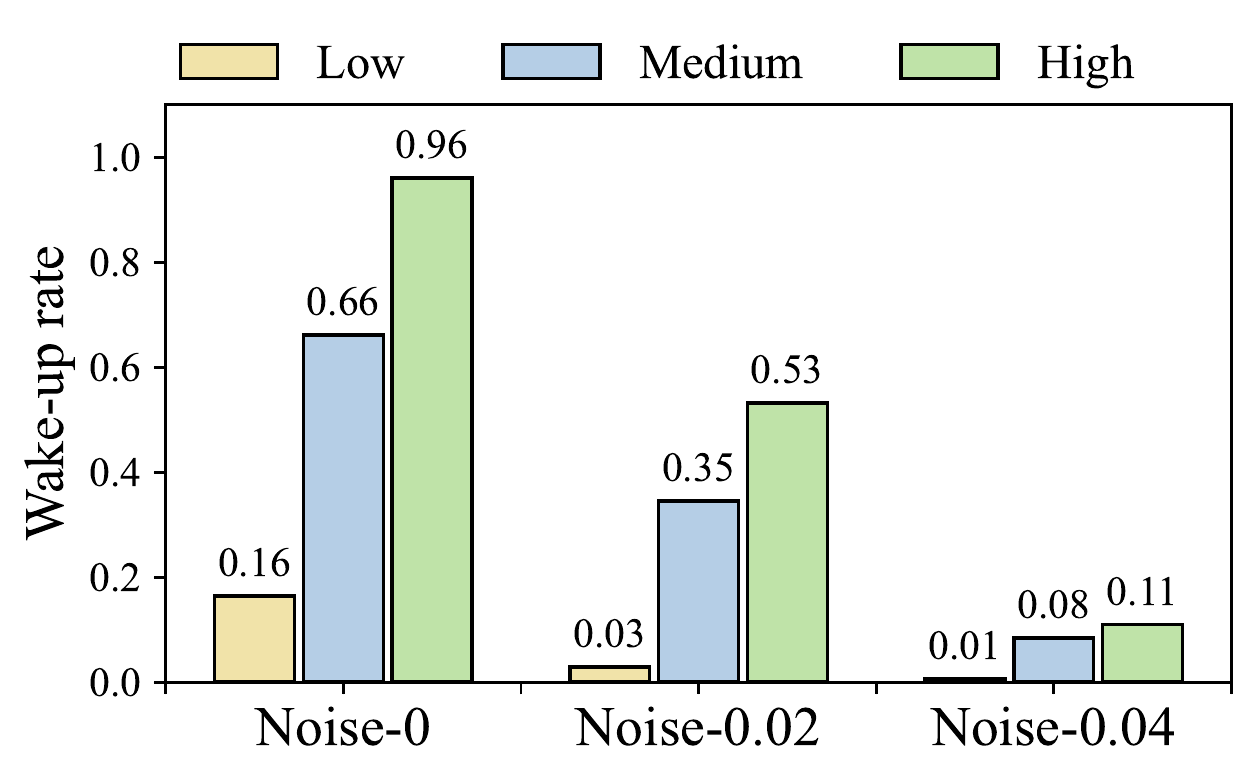}
    		\label{fig:ver_2figs_2cap_2}
    		\vspace{-0.2cm}
    	\end{minipage}
    }
    \vspace{-0.1cm}
    
    \subfigure[Google: Volume]{
    	\begin{minipage}[b]{0.3\linewidth}
    		\centering
    		\includegraphics[trim=0mm 0mm 0mm 10mm, clip, width=0.95\textwidth]{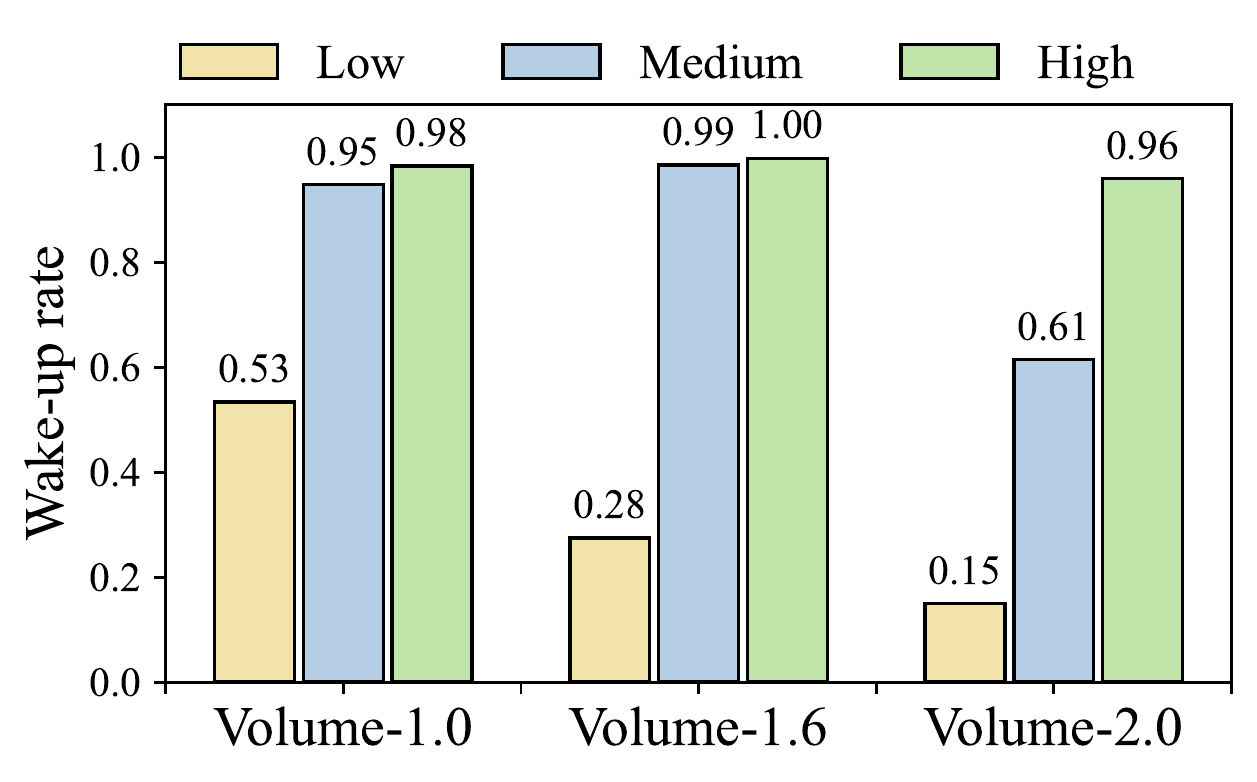}
    		\label{fig:ver_2figs_2cap_1}
    		\vspace{-0.2cm}
    	\end{minipage}
    }
    \vspace{-0.1cm}
    \subfigure[Google: Speed]{
    	\begin{minipage}[b]{0.3\linewidth}
    		\centering
    		\includegraphics[trim=0mm 0mm 0mm 10mm, clip,width=0.95\textwidth]{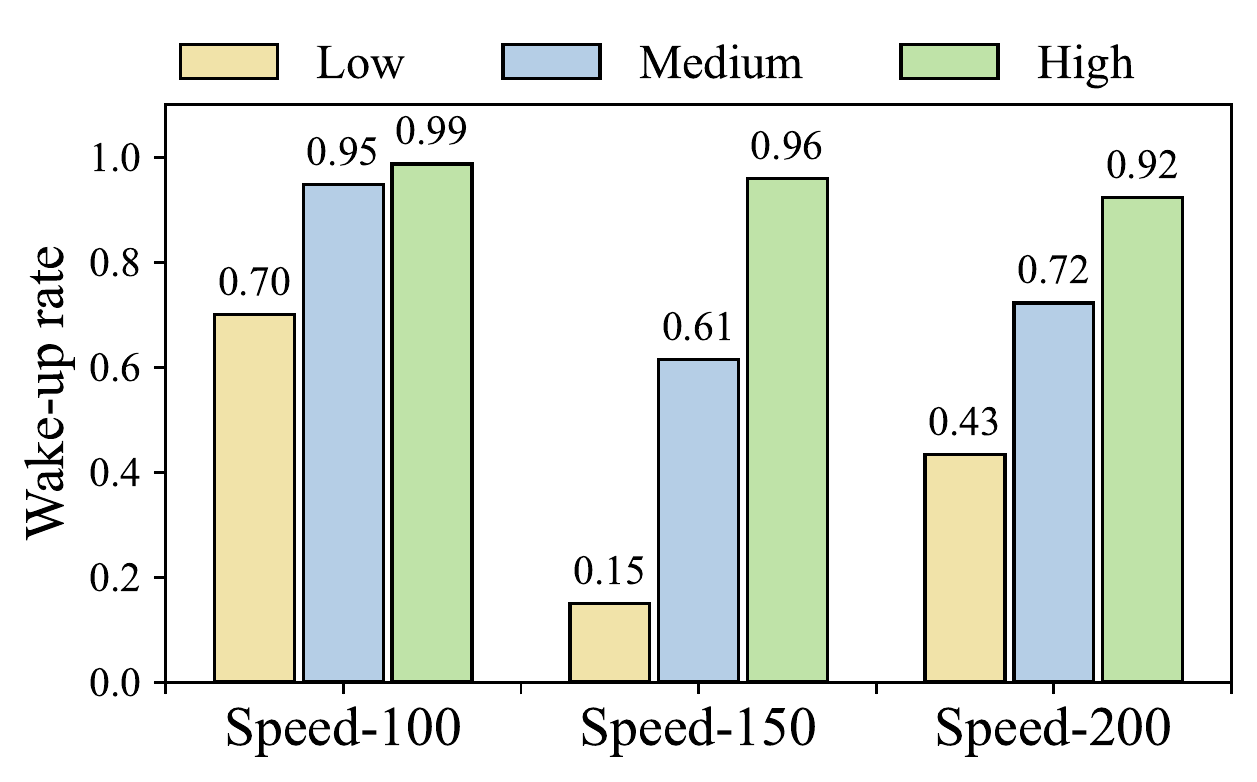}
    		\label{fig:ver_2figs_2cap_2}
    		\vspace{-0.1cm}
    	\end{minipage}
    }
    \vspace{-0.2cm}
    \subfigure[Google: Noise]{
    	\begin{minipage}[b]{0.3\linewidth}
    		\centering
    		\includegraphics[trim=0mm 0mm 0mm 10mm, clip,width=0.95\textwidth]{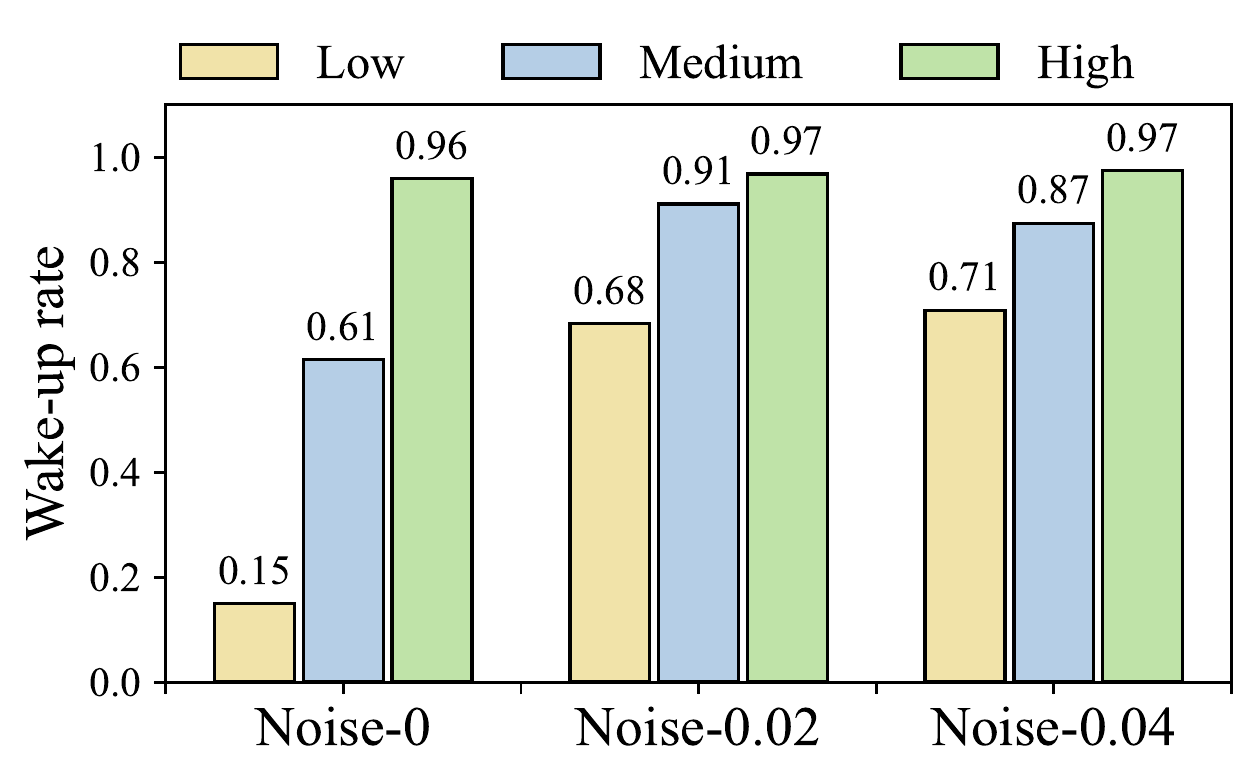}
    		\label{fig:ver_2figs_2cap_2}
    		\vspace{-0.2cm}
    	\end{minipage}
    }
    \vspace{-0.1cm}
    
    \subfigure[Apple: Volume]{
    	\begin{minipage}[b]{0.3\linewidth}
    		\centering
    		\includegraphics[trim=0mm 0mm 0mm 10mm, clip, width=0.95\textwidth]{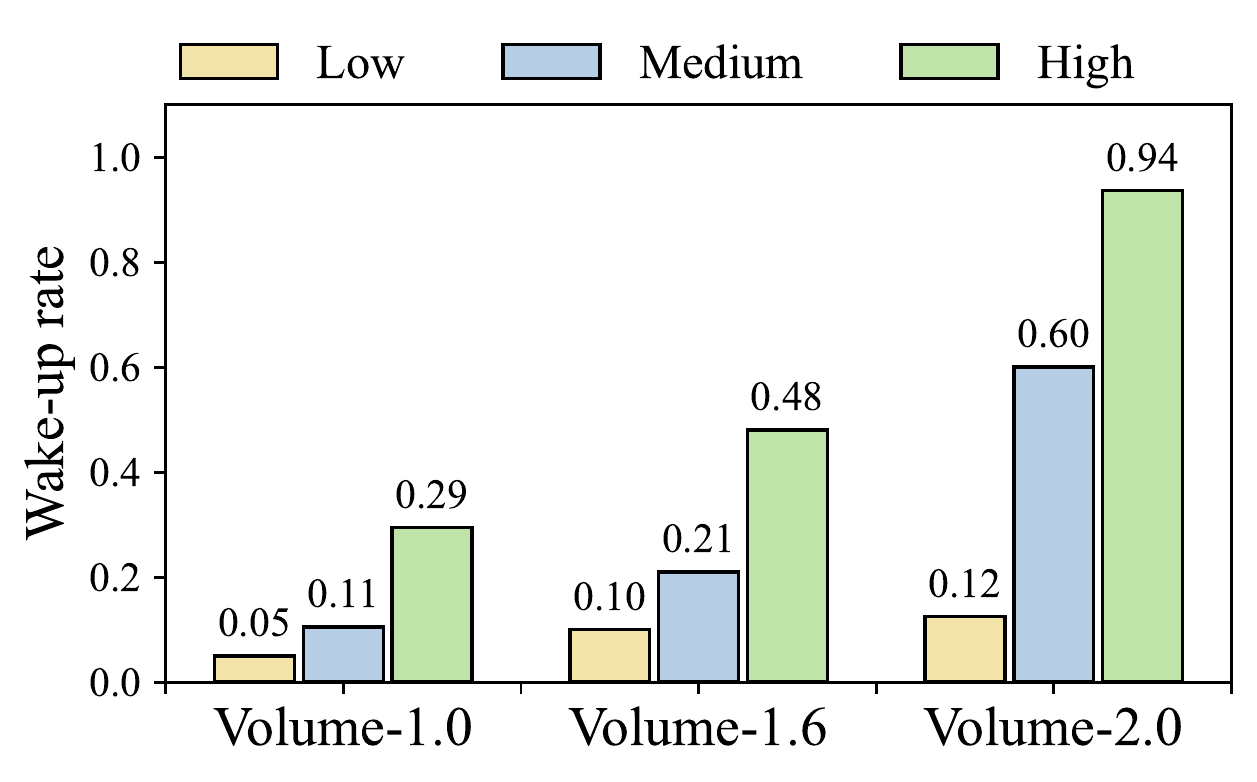}
    		\label{fig:ver_2figs_2cap_1}
    		\vspace{-0.2cm}
    	\end{minipage}
    }
    \vspace{-0.2cm}
    \subfigure[Apple: Speed]{
    	\begin{minipage}[b]{0.3\linewidth}
    		\centering
    		\includegraphics[trim=0mm 0mm 0mm 10mm, clip,width=0.95\textwidth]{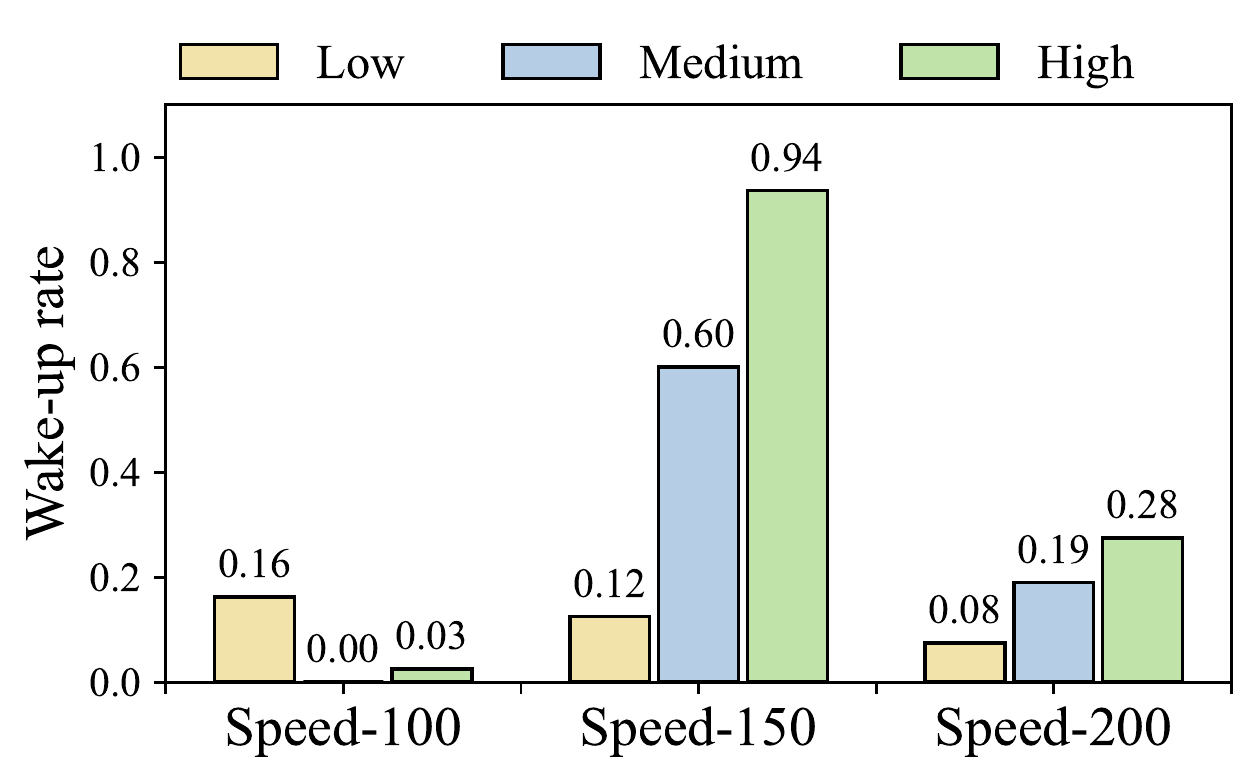}
    		\label{fig:ver_2figs_2cap_2}
    		\vspace{-0.2cm}
    	\end{minipage}
    }
    \vspace{-0.2cm}
    \subfigure[Apple: Noise]{
    	\begin{minipage}[b]{0.3\linewidth}
    		\centering
    		\includegraphics[trim=0mm 0mm 0mm 10mm, clip,width=0.95\textwidth]{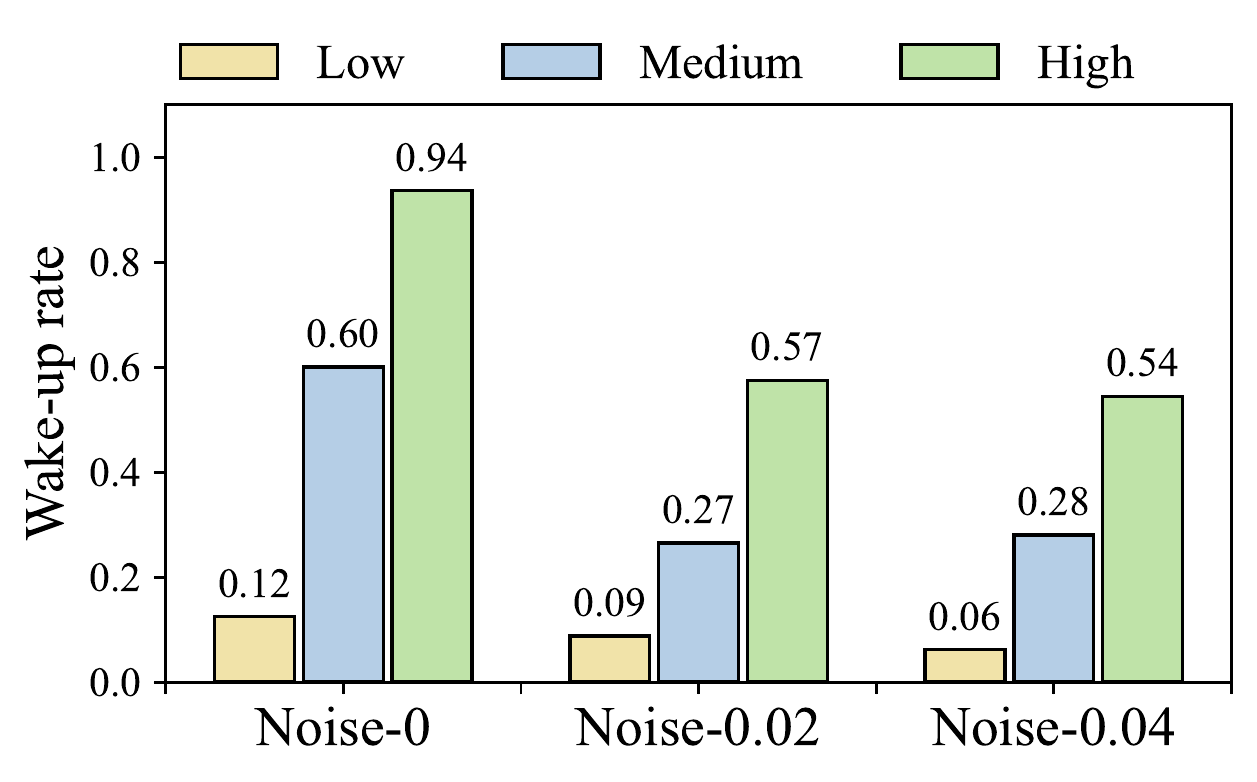}
    		\label{fig:ver_2figs_2cap_2}
    		\vspace{-0.2cm}
    	\end{minipage}
    }
    \vspace{-0.2cm}
	\caption{Wake-up rate of \fw under different environments for English voice assistants.}
	\vspace{-0.15in}
	\label{fig:Ablation_results_English}
\end{figure*}
	
\subsection{Generating English \FW}
	
	
An English word can be divided into graphemes (a letter or a letter combination) which correspond to different pronunciation units, i.e., phonemes.
It is worth noting that people can pronounce words that they have never seen (e.g., foreign names) based on experience, and Text to Speech (TTS) engines can pronounce "non-dictionary words"~\cite{g2p}, which increases the space of words for an attacker to generate \fw. 

\textbf{Encoding}. There are two special challenges facing the design for the English language. First, we need to decide whether to encode an English word based on its letter composition or phoneme composition. An intuitive thought is to encode an English word according to its phoneme composition. Nevertheless, we have found that this is not applicable due to several reasons. Firstly, it is not always possible to convert a combination of phonemes into a word in letter, which makes it difficult to produce meaningful \fw. Secondly, existing TTS services cannot pronounce phoneme combinations as naturally as letter combinations. The pronunciation of phoneme combinations sounds mechanical and incoherent, and cannot even wake up the voice assistants by saying the phoneme combinations of their real wake-up words. Therefore, we choose to encode English words according to their letter compositions. 
	
	
The second challenge is that English words have varied length. 
An English word can be as short as 1 letter and as long as 17-18 letters. Two English words with different lengths may sound similar, e.g., loose and lose. Moreover, a combination of two English words may sound like one English word, e.g., a lot and allot. Among the top three English voice assistants~\cite{market}, Amazon uses one word as the wake-up word, while Google and Apple use two. To address this problem, for a voice assistant with a specific wake-up word, we first use one variable to represent one letter, and then insert spaces (" ") between letters as a place holder to increase the overall length of an individual. The length of an individual is set as $r$ times that of the original wake-up word (with a length of $n$), where $r>1$ is generally set to 1.5. In this way, we not only address the problem of encoding wake-up words that are composed of two words, but also increase the diversity of the generated \fw. To sum up, we use an $r\times n$-dimension vector to represent an individual, where each variable represents a letter or a space, and the variable value ranges from 1 to 27. 
	
\textbf{Dissimilarity distance}. Similar to the Chinese language, the encoding of English words also does not carry pronunciation information, thus we choose phonemes instead of letters to quantify dissimilarity distance. Two same-length English words may have different numbers of phonemes, e.g., animal and beauty. To tackle this difficulty, we use Levenshtein distance between the phoneme composition of two English words to calculate their dissimilarity distance. Let $W_1=[p^{(1)}_1,...p^{(m)}_1]$ and $W_2=[p^{(1)}_2,...p^{(n)}_2]$ denote two English words, where $p^{(i)}$ is the $i$-th phoneme of a word. 
The conventional Levenshtein distance~\cite{leven} assumes that the distance between any pair of different phonemes is 1, while some phonemes sound similar and some phonemes sound far apart. Therefore, we integrate phonetic dissimilarity of phonemes~\cite{panphon} into the Levenshtein distance to quantify the dissimilarity distance between two words. Let $\operatorname{dist}(p^{(i)}, p^{(j)}) \in [0,1]$ denote the dissimilarity of two phonemes. We have 
	
	\begin{equation}
	    \operatorname{dist}(W_1, W_2) = \frac{D+I+2\sum\limits_{(i,j) \in \mathcal{S}} \operatorname{dis}(p^{(i)}_1, p^{(j)}_2)}{m+n} ,
	\end{equation}
where $D$ and $I$ denote the number of deletions and insertions respectively, $\mathcal{S}$ is the set of substitutions which replaces phoneme $p^{(i)}_1$ with $p^{(j)}_2$, and $m$ and $n$ are the length of $W_1$ and $W_2$ respectively. The dissimilarity between space and any phoneme is set as 1.
    

\subsection{Experiment Results}\label{sec:exp}

We conduct extensive experiments to answer the following questions:
\begin{itemize}[leftmargin=*]
    \item \textbf{(Q1)} How does our proposed generation framework perform in producing \fw for different Chinese and English smart speakers?
    \item \textbf{(Q2)} How does environmental factors, including volume, speed, noise levels and gender speaker, influence the robustness of the generated \fw?
    \item \textbf{(Q3)} Do the generated \fw sound different from the real wake-up words from the human perspective?
\end{itemize}
We will answer these questions after presenting the experiment settings.

\textbf{Evaluated voice assistants}.
For English voice assistants, we conduct experiments on Amazon Echo, Amazon Echo Dot, Google Nest Mini and Apple HomePod. Amazon Echo and Amazon Echo Dot can be woken up by "Alexa", "Amazon", "Echo" or "Computer", and we focus on generating \fw of "Alexa". Google Nest Mini can be woken up by "Hey Google" or "Ok Google", and we focus on generating \fw of "Hey Google". Apple HomePod's wake-up word is "Hey Siri", the same as iPhone and iPad. 
	
For Chinese voice assistants, we conduct experiments on Baidu, Xiaomi, AliGenie, and Tencent. Baidu smart speakers can be woken up by "xiǎo dù xiǎo dù" (小度小度), a repetition of the nickname of Baidu (百度). Xiaomi smart speakers can be triggered by "xiǎo ài tóng xué" (小爱同学), the Chinese name of Xiaomi's virtual assistant. AliGenie, also known as Tmall Genie, can be woken up by "tiān māo jīng líng" (天猫精灵), the Chinese name for the smart speaker. AliGenie can also be woken up by "nǐ hǎo tiān māo" (你好天猫), where "nǐ hǎo" means "hello" in Chinese.  Tencent smart speakers can be activated by "jiǔ sì èr líng" (九四二零), which sounds similar to "I just love you" (就是爱你) in Chinese.
	
\textbf{Experiment setup}. As shown in Figure~\ref{exsetup}, our experiment setup consists of a laptop, a Raspberry Pi, a stereo and a light sensor. The laptop runs the generation algorithm to produce \fw, and the stereo plays the audio of each generated fuzzy word to test its wake-up rate. The Raspberry Pi is equipped with a light sensor to detect whether the smart speaker is activated or not. The Raspberry Pi returns the wake-up rate of the \fw to the generation algorithm to evaluate their fitness.


Our experiments are carried out in a quiet laboratory room. We employ pyttsx3 to generate the audio samples of \fw by using TTS, which can articulate non-dictionary words~\cite{g2p}. The distance between the stereo and the tested smart speaker is 20 centimeters. We play the audio samples with the default male voice at a moderate volume. The play speed is set to 150 in pyttsx3 by default, which approximates the average speed of human speakers. Each word is played 10 times for wake-up rate computation. 


\textbf{Performance of generation framework (Q1)}.
We display the results of fuzzy word generation in Table~\ref{tab: fuzzy result}. A full list of generated \fw is in the Appendix. Note that we consider the \fw generated by the genetic algorithm as out of distribution, since subjective tests show that users perceive the \fw as different from the real wake-up word. Furthermore, we leverage the generated \fw to strengthen the wake-up word detector, which improves its performance regarding both \fw and non-\fw.   


For Chinese voice assistants, Baidu has the fewest \fw while AliGenie has the most \fw. This indicates that repetition in xi\v ao d\`u xi\v ao d\`u may indeed mitigate the \sys phenomena. Tencent also has a small number of \fw since the wake-up word jiǔ sì èr líng contains rich combinations of initials, finals and tones. AliGenie has a significantly larger number of \fw since its wake-up word detector relies heavily on \emph{ian} to detect the wake-up word, which we will explain in Section~\ref{sec:understand}. For English voice assistants, longer wake-up words, e.g., Hey Google and Hey Siri, have fewer \fw. Also, Siri is a less commonly-used word with distinctive pronunciation, making it more difficult to produce its \fw.

\begin{figure}[t]
    \centering
    \begin{minipage}[b]{0.97\linewidth}
    \centering
    \includegraphics[width=0.9\linewidth,bb= 0 0 6.46in 0.32in]{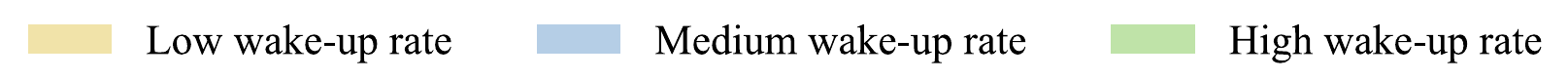}
    \end{minipage}
    \begin{minipage}[b]{0.8\linewidth}
    \centering
    \includegraphics[width=0.8\linewidth, bb=0 0 7.76in 3.33in]{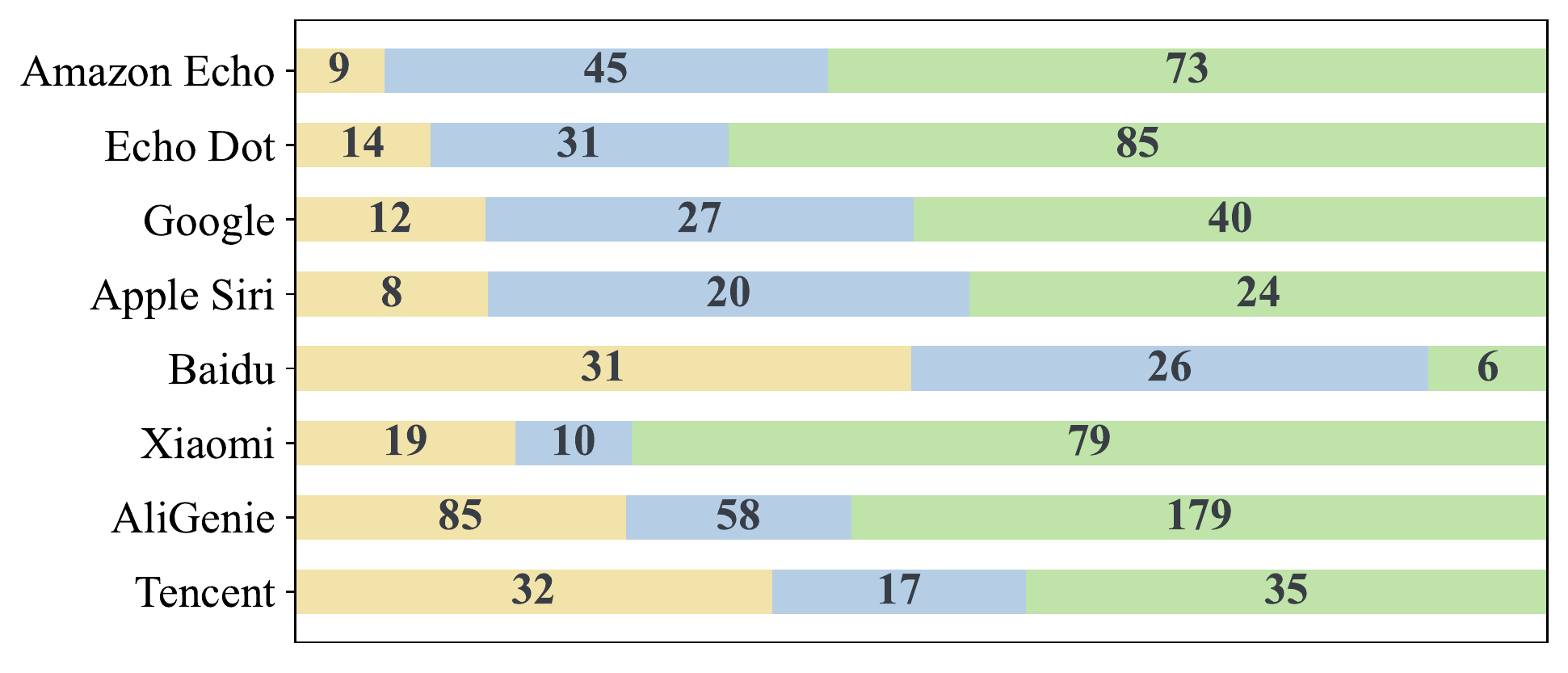}
    \end{minipage}
    \caption{Distribution of \fw for different voice assistants. Wake-up rate ranges: low (0-0.3), medium (0.4-0.7) and high (0.8-1.0). The numbers of \fw are marked. }
    \vspace{-0.15in}
    \label{Wake rate proportion}
    \end{figure}

We show the wake-up rate distribution of \fw for different voice assistants in Figure~\ref{Wake rate proportion}, where we divide the wake-up rate into three ranges: low (0.1$\sim$0.3), medium (0.4$\sim$0.7), and high (0.8$\sim$1.0). More than 40\% \fw have high wake-up rate except for Baidu, which may be another benefit of word repetition.




\textbf{Environmental impact on \fw (Q2)}. We investigate the impact of volume, speed, noise level and speaker gender on the wake-up rate of generated \fw. For volume, we change the volume to 1.6 and 2 times of the original volume. For speed, the original speed parameter is 150, and we change this parameter to 100 (slower) and 200 (faster). For noise level, we add Gaussian noise to audio samples to simulate environmental noises. The parameter of the Gaussian noise is set to 0.02 and 0.04. At 0.02, the signal-to-noise ratio (SNR) is around 40db (office, library), and at 0.04, the SNR is around 30db (soft music, whisper)~\cite{noise, commonnoise}. For speaker gender, the original audio samples use a male voice, and we change the voice to female. As the TTS we use does not support Chinese female voice, we test the impact of speaker gender only on English voice assistants.

\begin{tcolorbox}[enhanced, fontupper=\normalsize, left=2pt,right=2pt,top=0pt,bottom=0pt, title =\textbf{Insight 1}]
 Fuzzy words with high wake-up rate mostly maintain their wake-up rate when the environment changes. 
\end{tcolorbox}

We demonstrate the experiment results of Chinese and English voice assistants in Figure \ref{fig:Ablation_results_Chinese} and Figure~\ref{fig:Ablation_results_English} respectively. 

\begin{itemize}[leftmargin=*]
  \item  \emph{Volume}. With increased volume, the wake-up rate rises for most \fw of most voice assistants. But for Baidu and Google, the wake-up rate decreases with a higher volume. 
    
  \item   \emph{Speed}. Speed has a mixed influence on the wake-up rate. Generally, as speed increases, the wake-up rate first goes up then goes down, especially for the English voice assistants, e.g., Echo and Apple Siri. This may be because a slightly faster speed boosts coherence of the TTS, but a super fast speed makes the speech intelligible for the voice assistants.
    
 \item    \emph{Noise level}. In general, noises degrade the wake-up rate. But for Xiaomi, the wake-up rate for \fw with medium wake-up rate grows to as high as 0.91 at a high noise level. Similar trend is also observed in Google.  
    
 \item    \emph{Speaker gender}. After changing the speaker voice from male to female, the mean wake-up rate for \fw with high wake-up rate decreases, while the mean wake-up rate for \fw with medium and low wake-up rate increases. 
 Due to page limitation, the results of the influence of speaker gender is in the Appendix. 
    \end{itemize}

\textbf{Perceptual difference of \fw (Q3)}. We conduct a subjective test to investigate whether the generated \fw with high dissimilarity distance according to the generation algorithm indeed sound different from the real wake-up words to human ears. 

We have recruited 33 volunteers (5 females and 28 males). For each evaluated smart speaker, we choose the top 20 \fw with the largest dissimilarity distance. We ask each volunteer to listen to the audio of each fuzzy word for 3 times and then evaluate the dissimilarity between the fuzzy word and the real wake-up word on a scale from 1 (very similar) to 5 (very dissimilar). We also ask each volunteer to score whether the fuzzy word is common in daily life on a scale from 1 (not common) to 5 (very common). We show the results of the top 10 \fw in Figure~\ref{expdist}, and the results of the remaining \fw are in the Appendix.

\begin{tcolorbox}[enhanced, fontupper=\normalsize, left=2pt,right=2pt,top=0pt,bottom=0pt, title = \textbf{Insight 2}]
  Fuzzy words with high dissimilarity distance also have high perceptual difference from the real wake-up words. 
\end{tcolorbox}
As shown in Figure~\ref{expdist}, for most voice assistants, the generated \fw have an average perceptual difference of more than 3.0. Specifically, the \fw for Xiaomi have high perceptual difference (more than 4.0), while the \fw for Echo have relatively low perceptual difference (around 3.0). The \fw with high perceptual difference tend to be regarded as less common in daily life. There is large variance for certain \fw due to individual differences in hearing experiences.

	\begin{figure*}[tt]
    \centering
    \begin{minipage}[b]{0.97\linewidth}
    		\centering
    		\includegraphics[trim=0mm 0mm 0mm 0mm, clip, width=0.3\textwidth]{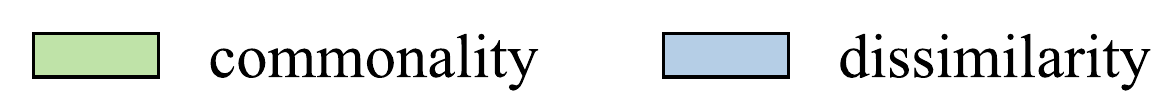}
    	\end{minipage}
    \subfigure[Amazon Echo]{
	\includegraphics[trim = 3mm 4mm 8mm 7mm, clip, width=0.23\linewidth]{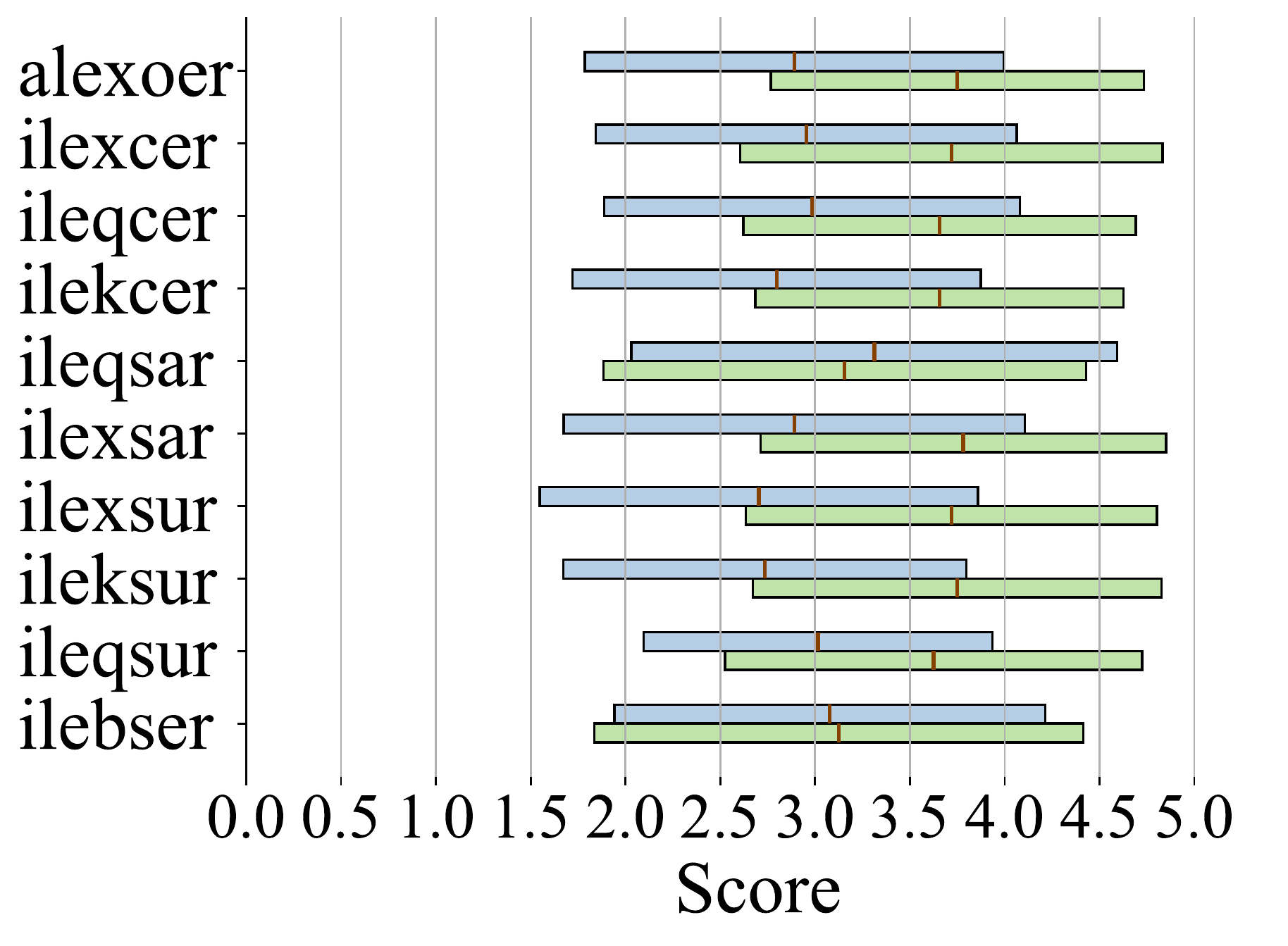}
	}
	\subfigure[Echo Dot]{
	\includegraphics[trim = 3mm 4mm 8mm 7mm, clip,width=0.23\linewidth]{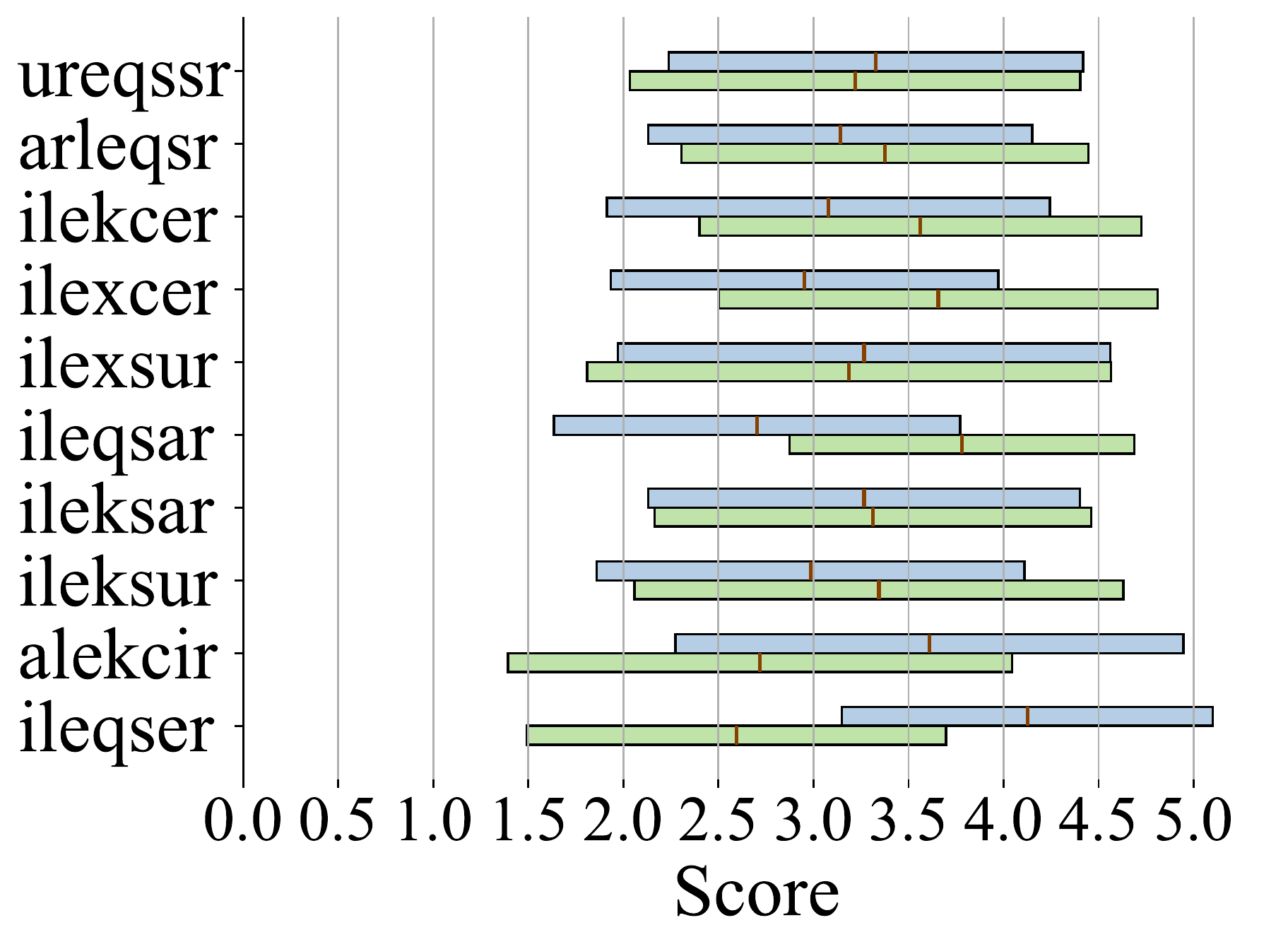}
	}
	\subfigure[Google]{
	\includegraphics[trim = 3mm 4mm 8mm 7mm, clip,width=0.23\linewidth]{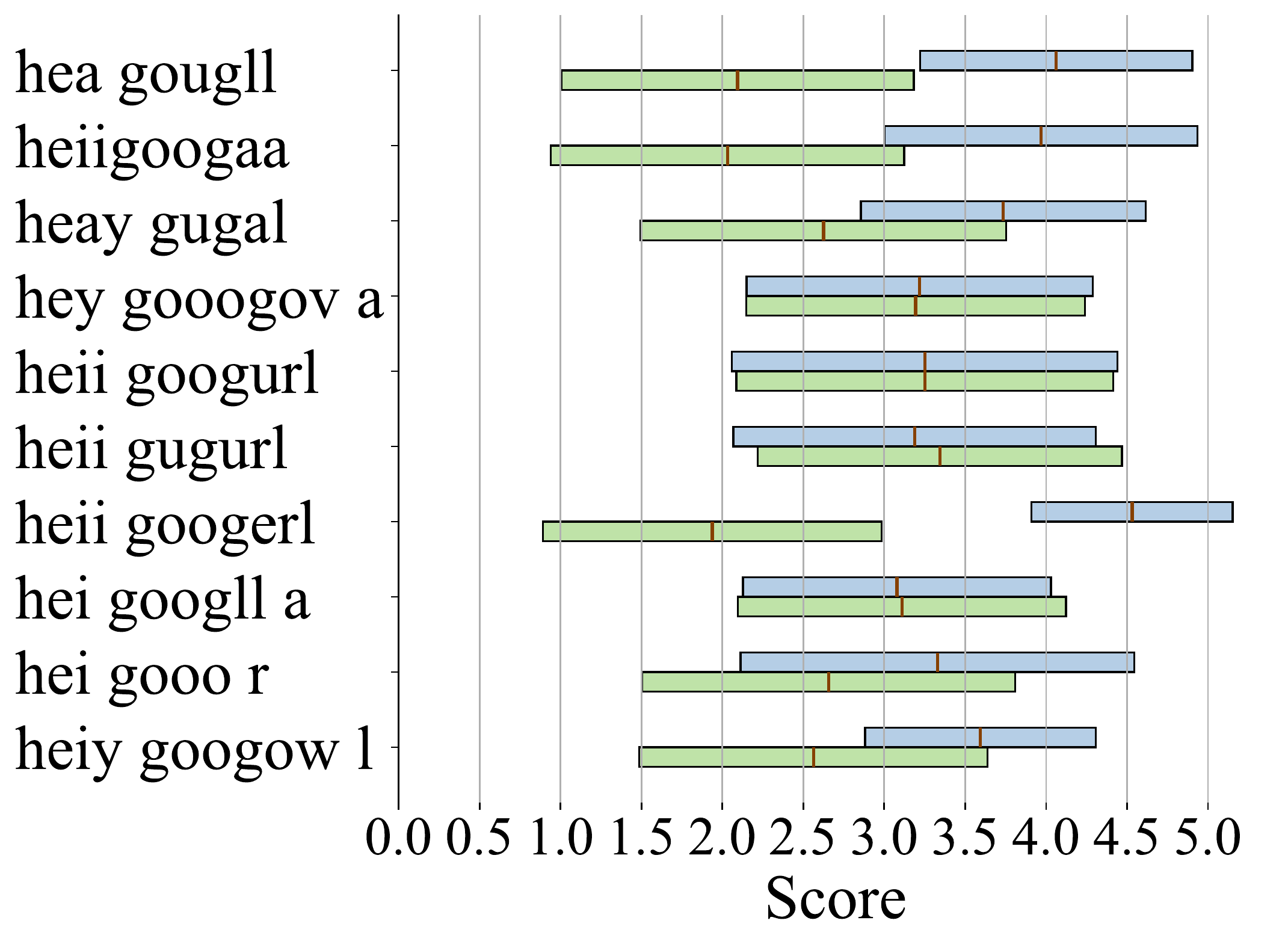}
	}
	\subfigure[Apple Siri]{
	\includegraphics[trim = 3mm 4mm 8mm 7mm, clip,width=0.235\linewidth]{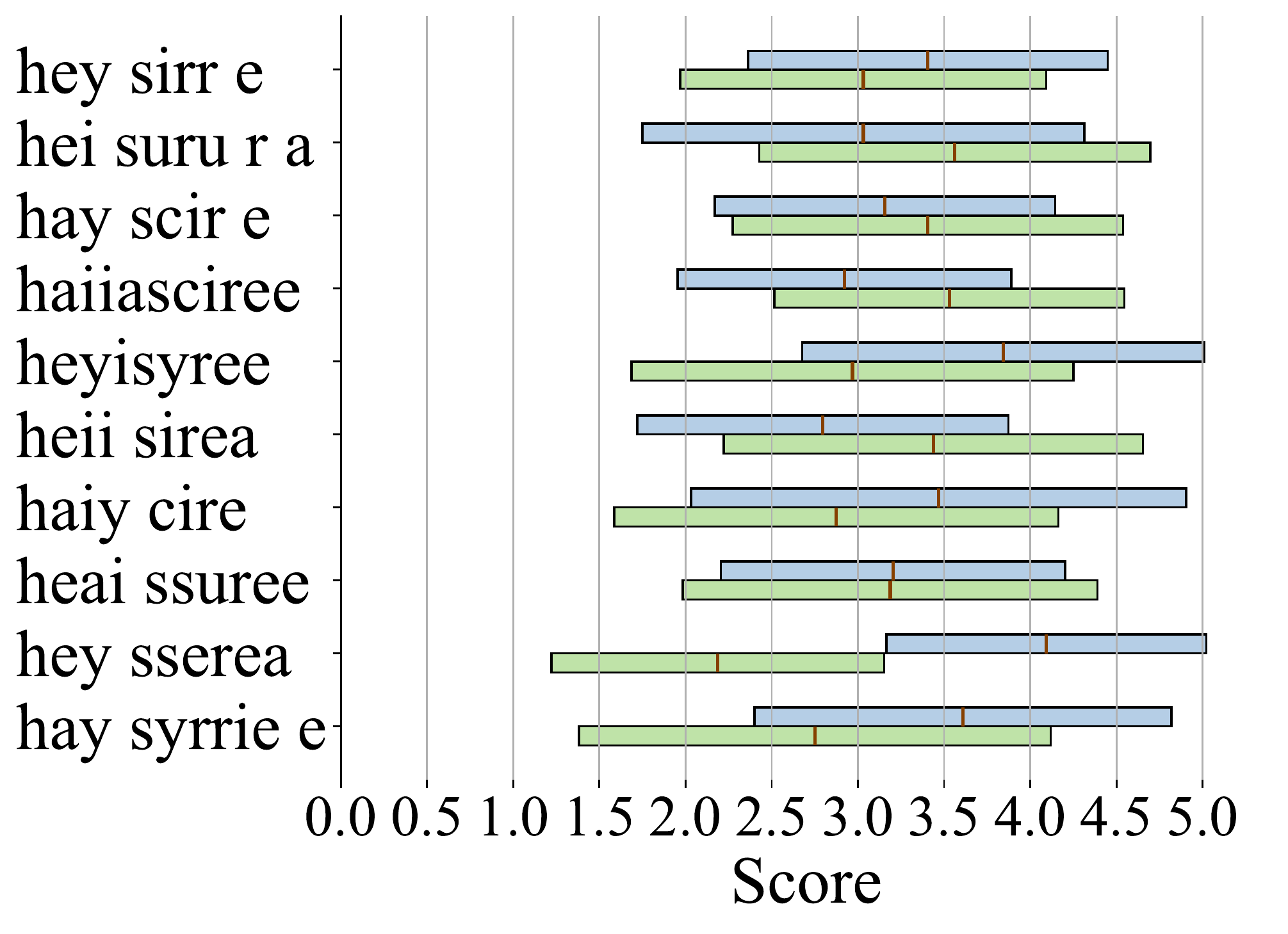}
	}
	\subfigure[Baidu]{
	\includegraphics[trim = 3mm 4mm 8mm 7mm, clip,width=0.23\linewidth]{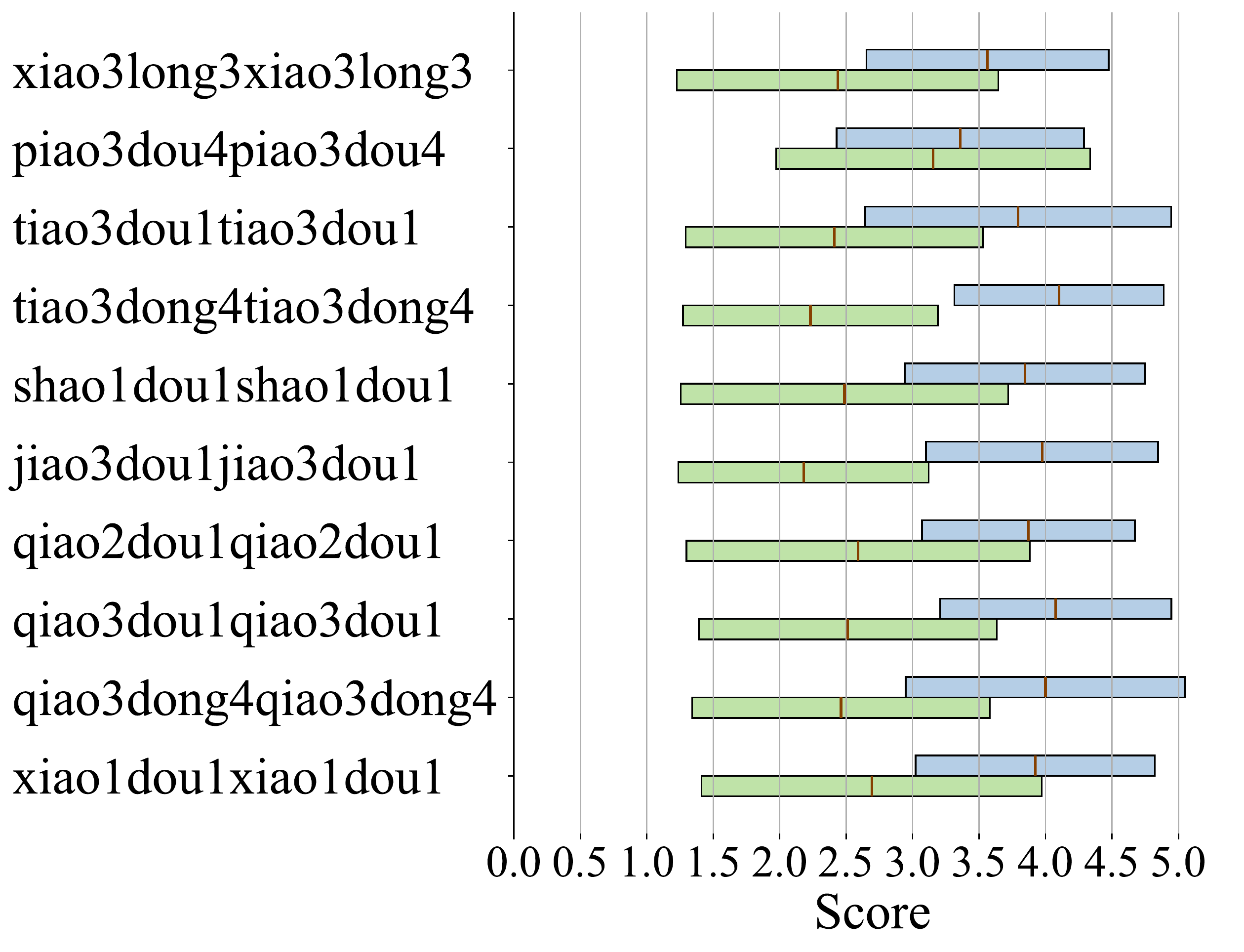}
	}
	\subfigure[Xiaomi]{
	\includegraphics[trim = 3mm 4mm 8mm 7mm, clip,width=0.23\linewidth]{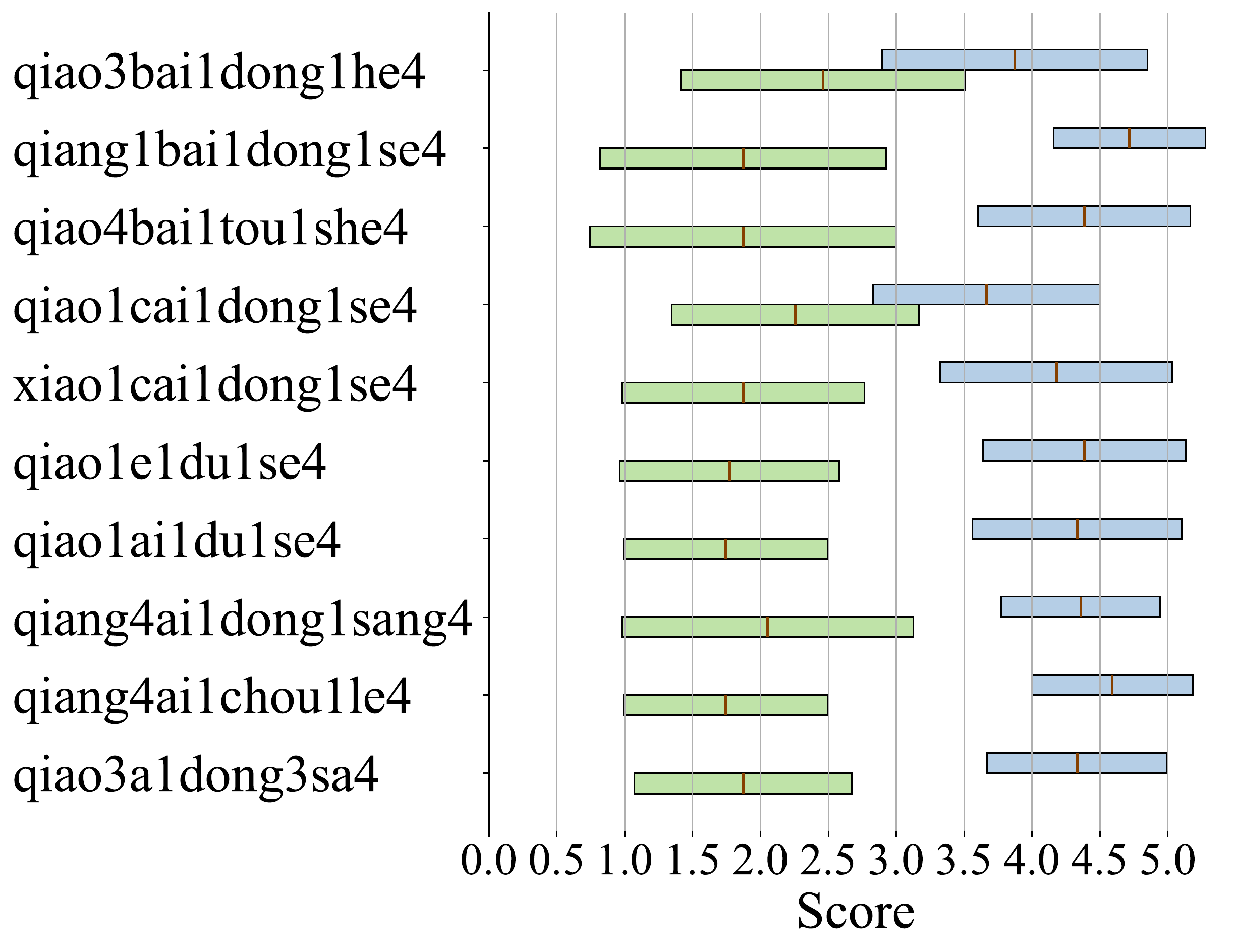}
	}
	\subfigure[AliGenie]{
	\includegraphics[trim = 3mm 4mm 8mm 7mm, clip,width=0.23\linewidth]{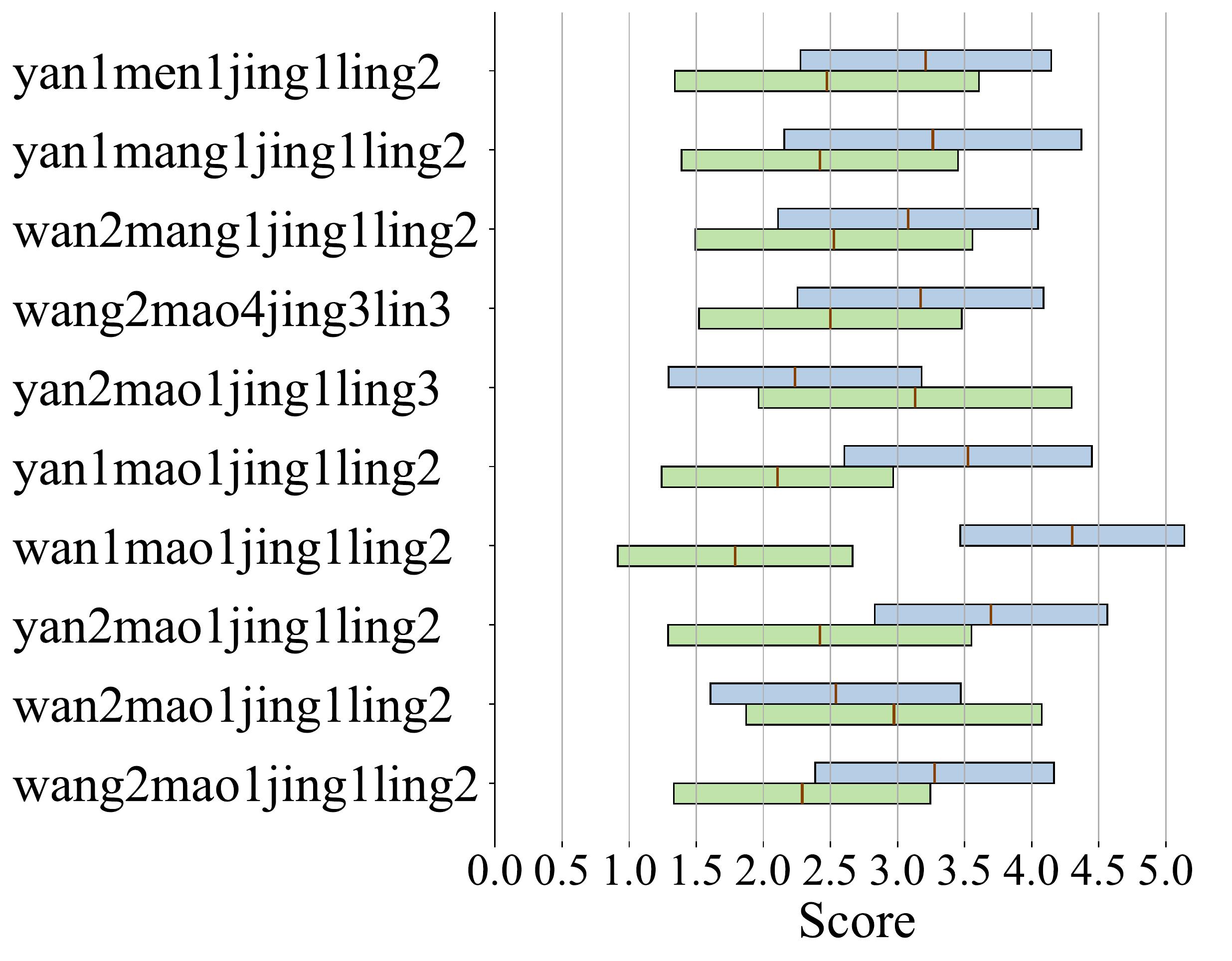}
	}
	\subfigure[Tencent]{
	\includegraphics[trim = 3mm 4mm 8mm 7mm, clip,width=0.23\linewidth]{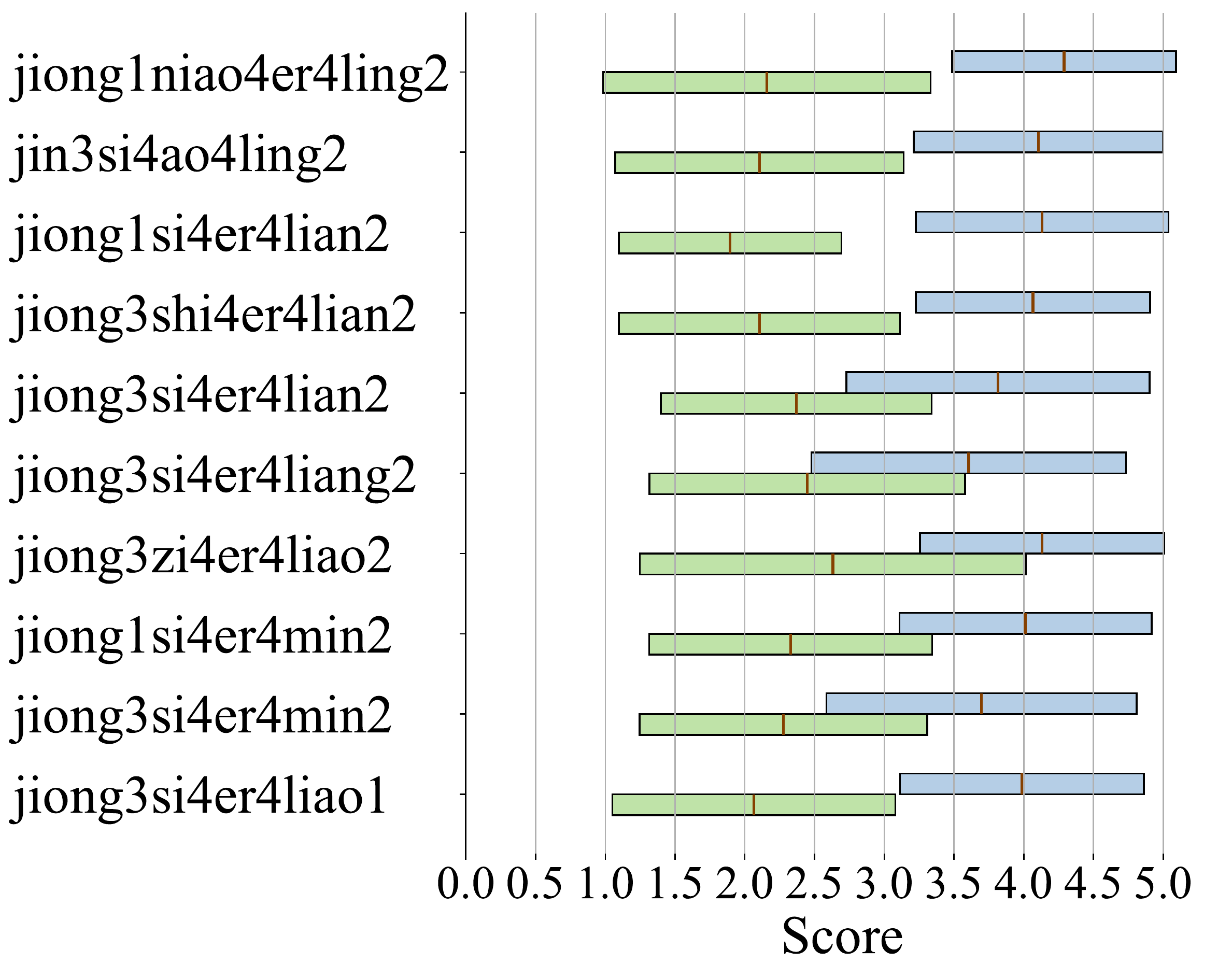}
	}

	\caption{The results of subjective tests for the top-10 \fw. The perceptual dissimilarity between \fw and the real wake-up words ranges from 1 (very similar) to 5 (very dissimilar). The commonality of a fuzzy word in daily life ranges from 1 (not common) to 5 (very common). }
	\label{expdist}
    \end{figure*}

\section{Understanding \FW}\label{sec:understand}
	\label{explanation}
To understand the cause of the \sys phenomena, we aim to find the decisive factors that lead to false acceptance of \fw. Nonetheless, we have no access to the black-box wake-up word detector, which makes it impossible to analyze the causes of false acceptance at the model level. Even if we have side information of the wake-up word detectors, they usually use deep learning models, which are notorious for its non-interpretable nature due to its highly nonlinear structures and massive parameters. Therefore, we try to explain the \sys phenomena by analyzing the generated \fw. 

	\begin{figure*}[tt]
	\includegraphics[width=0.7\linewidth]{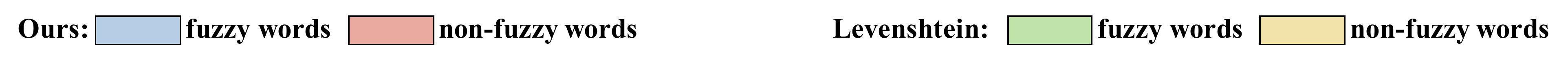}
	\vspace{-1.5em}
	\end{figure*}
	\begin{figure*}[tt]
    \centering
    \subfigure[Amazon Echo]{
	\includegraphics[trim = 2mm 0mm 3mm 0mm, clip, width=0.23\linewidth]{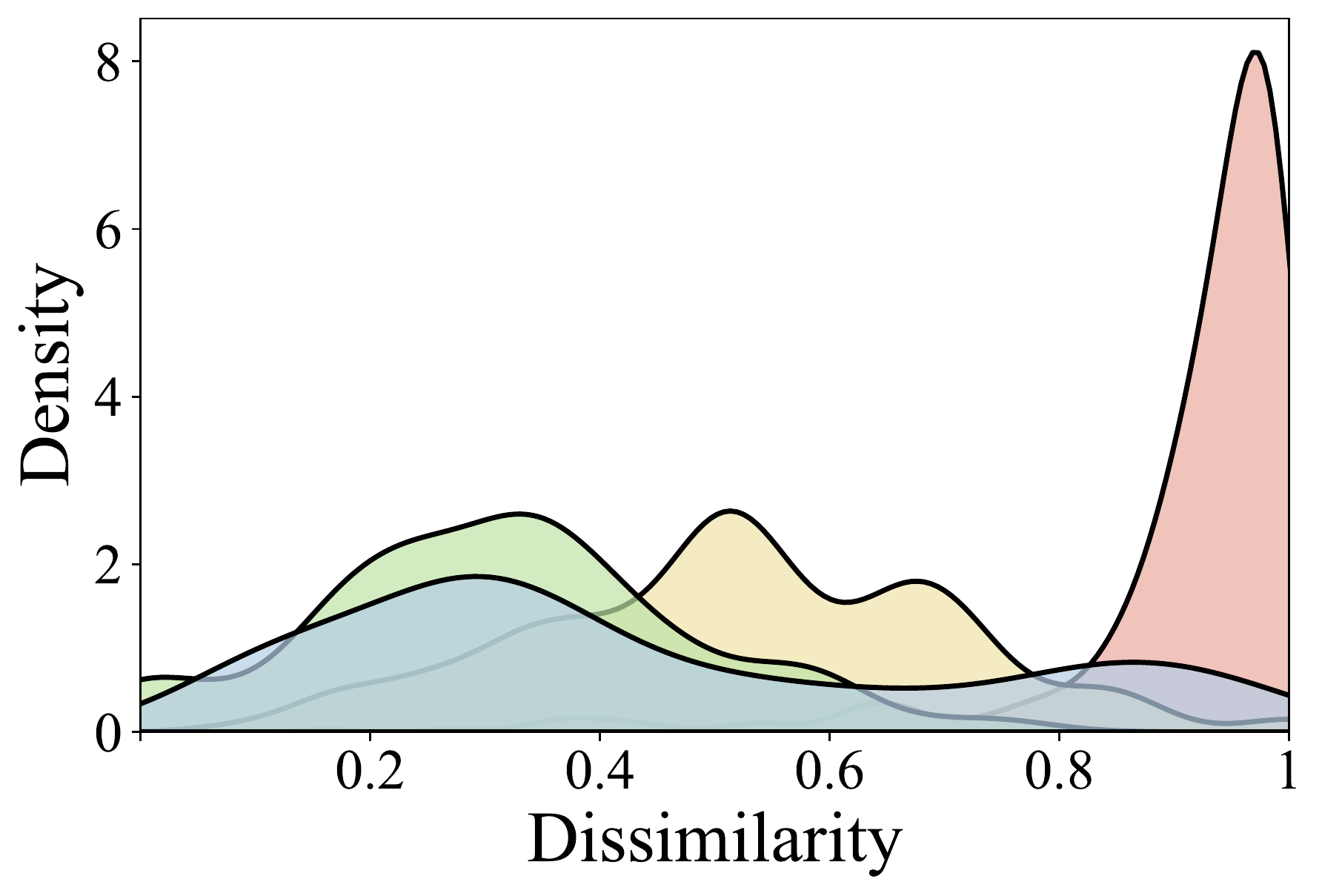}
	}
	\vspace{-0.1cm}
	\subfigure[Echo Dot]{
	\includegraphics[trim = 2mm 0mm 3mm 0mm, clip,width=0.23\linewidth]{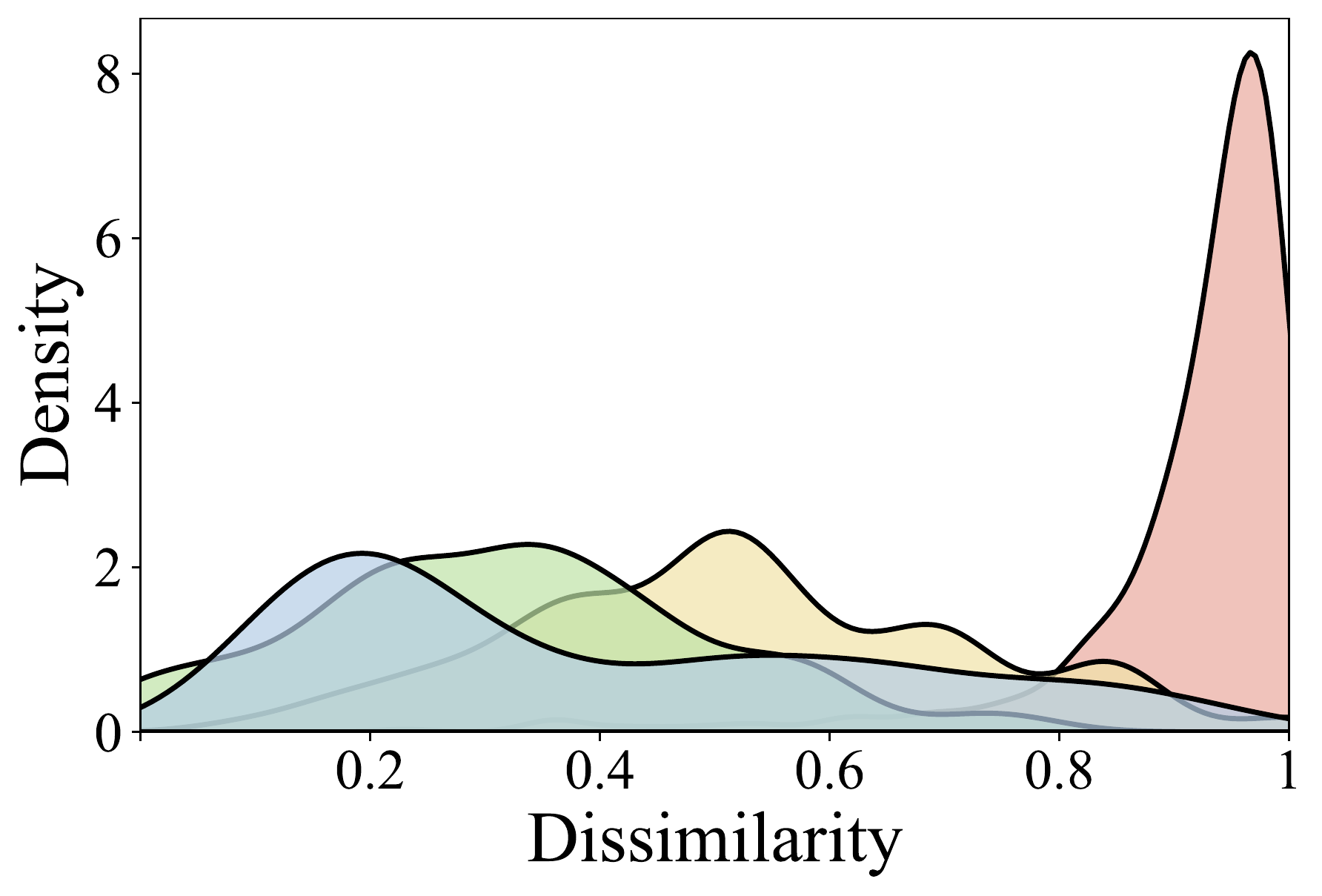}
	}
	\vspace{-0.1cm}
	\subfigure[Google]{
	\includegraphics[trim = 2mm 0mm 3mm 0mm, clip,width=0.23\linewidth]{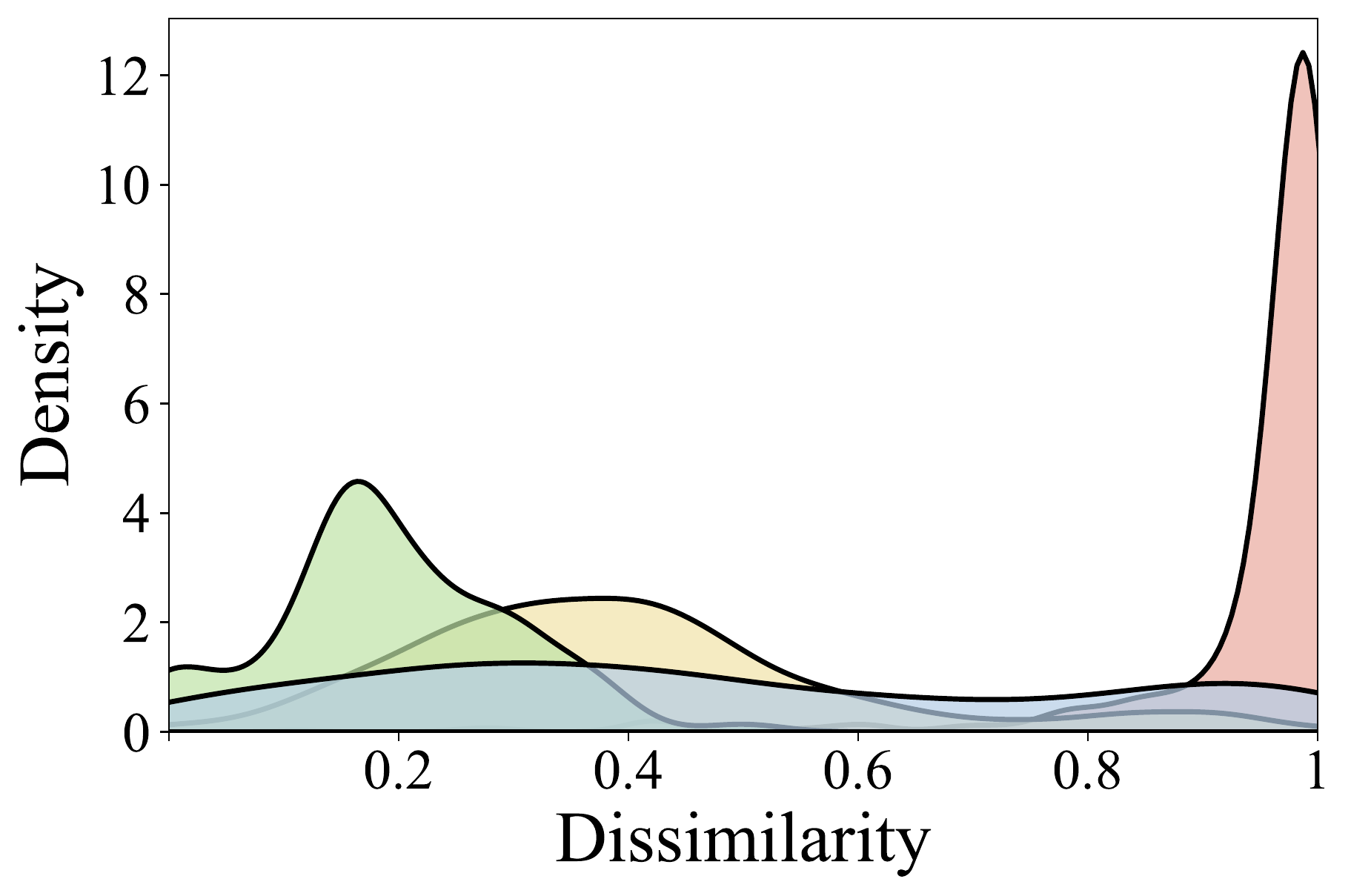}
	}
	\vspace{-0.1cm}
	\subfigure[Apple Siri]{
	\includegraphics[trim = 2mm 0mm 3mm 0mm, clip,width=0.23\linewidth]{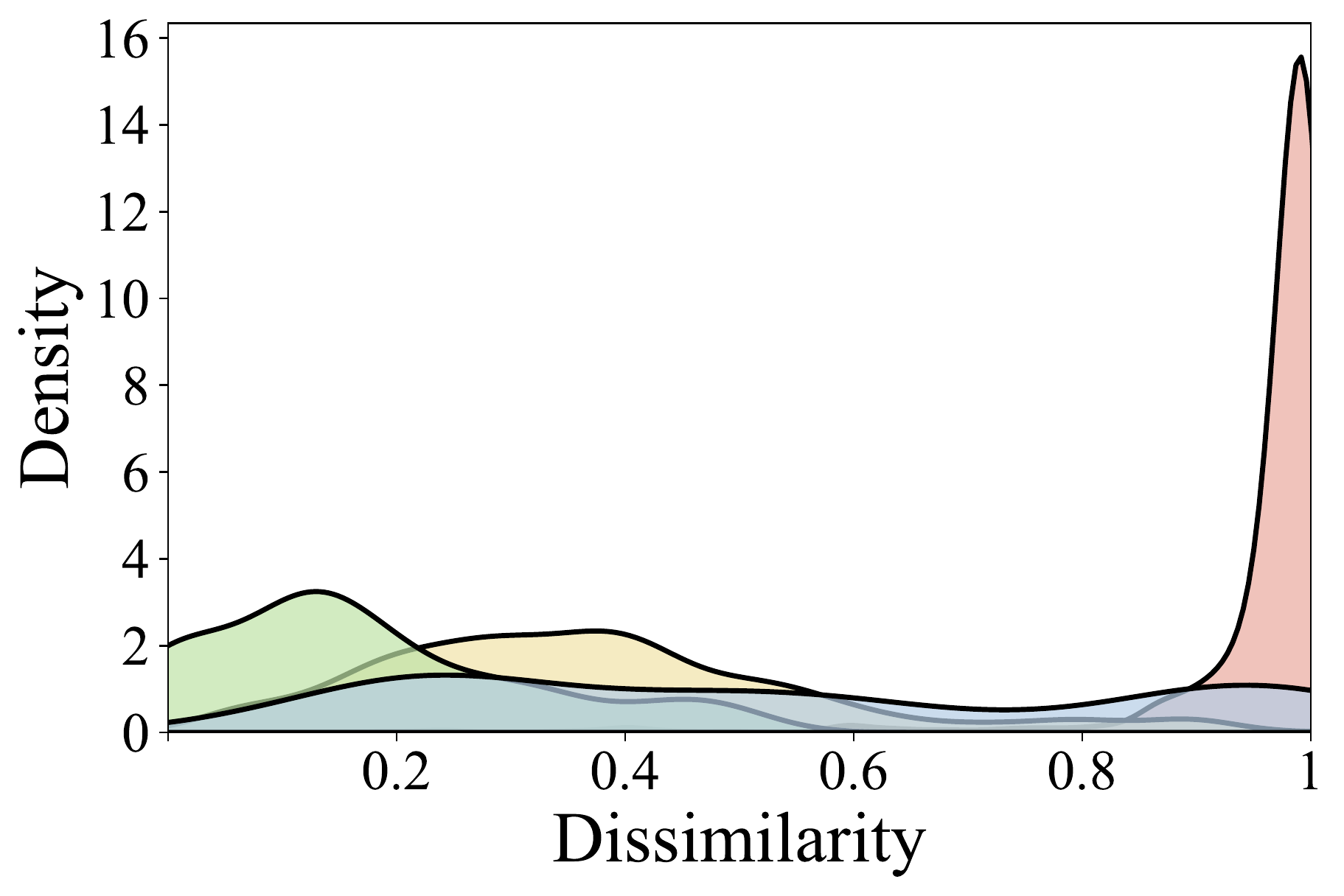}
	}
	\vspace{-0.1cm}
	\subfigure[Baidu]{
	\includegraphics[trim = 3mm 0mm 3mm 0mm, clip,width=0.23\linewidth]{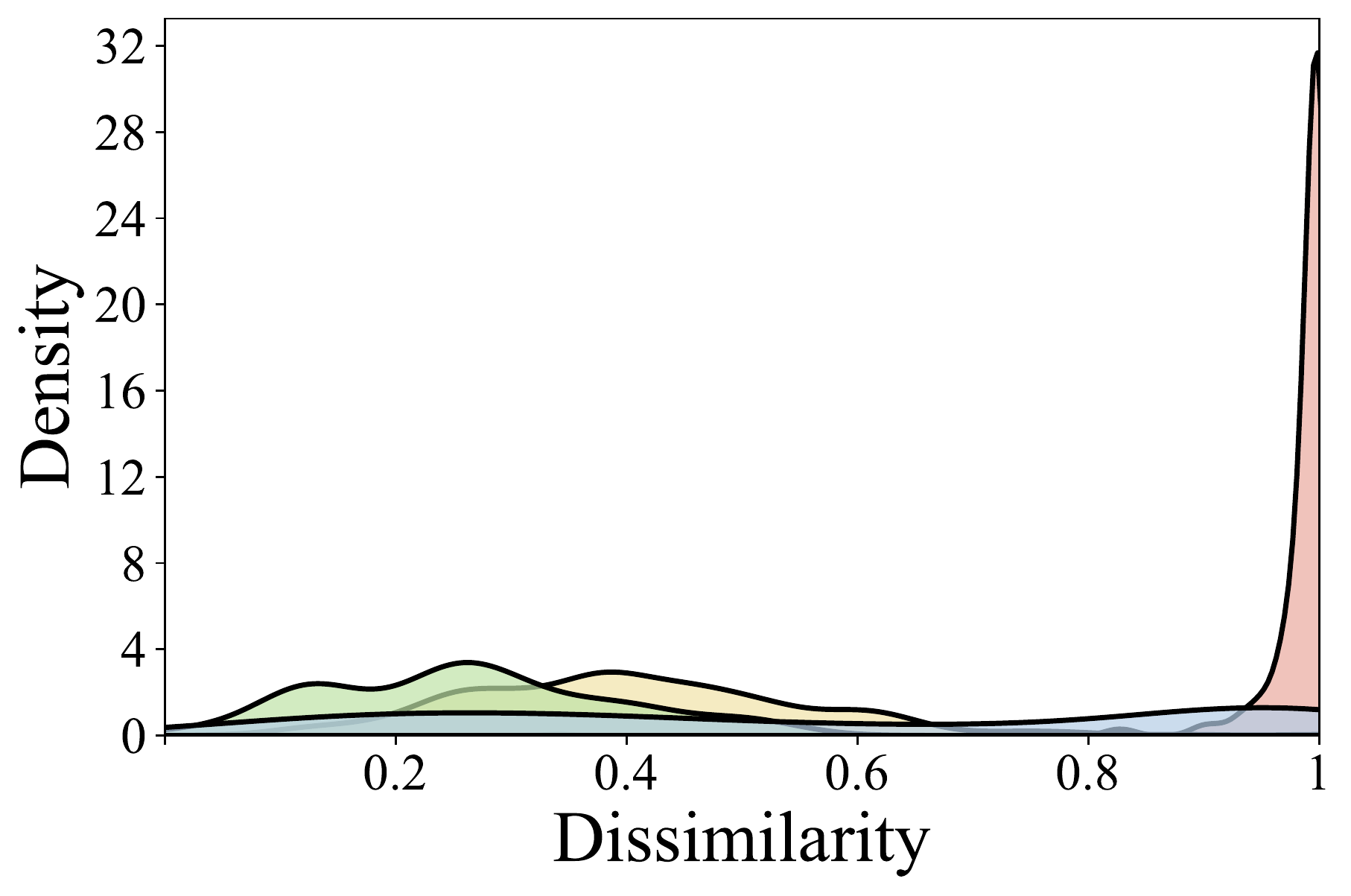}
	}
	\vspace{-0.1cm}
	\subfigure[Xiaomi]{
	\includegraphics[trim = 3mm 0mm 3mm 0mm, clip,width=0.23\linewidth]{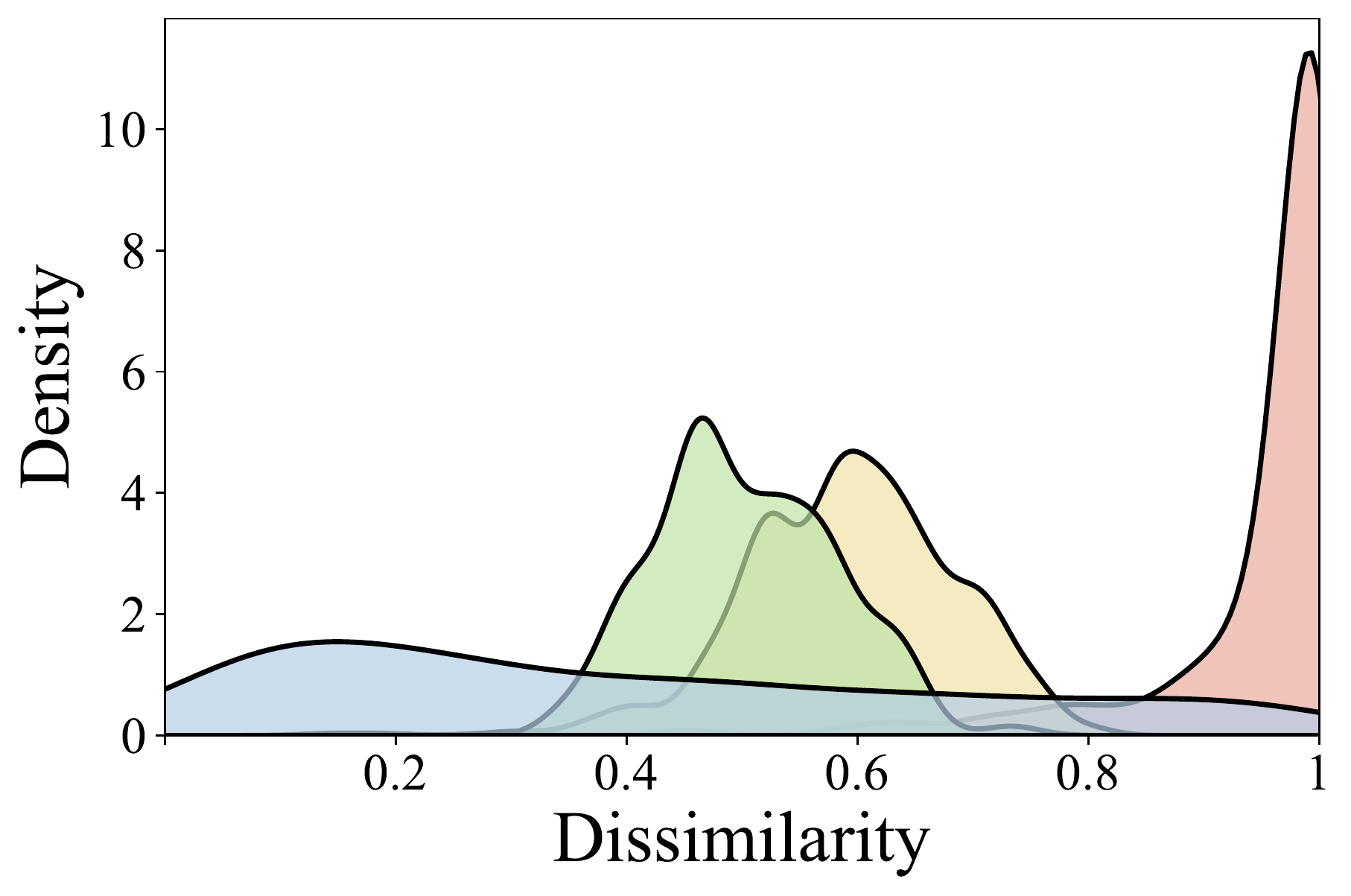}
	}
	\vspace{-0.1cm}
	\subfigure[AliGenie]{
	\includegraphics[trim = 3mm 0mm 3mm 0mm, clip,width=0.23\linewidth]{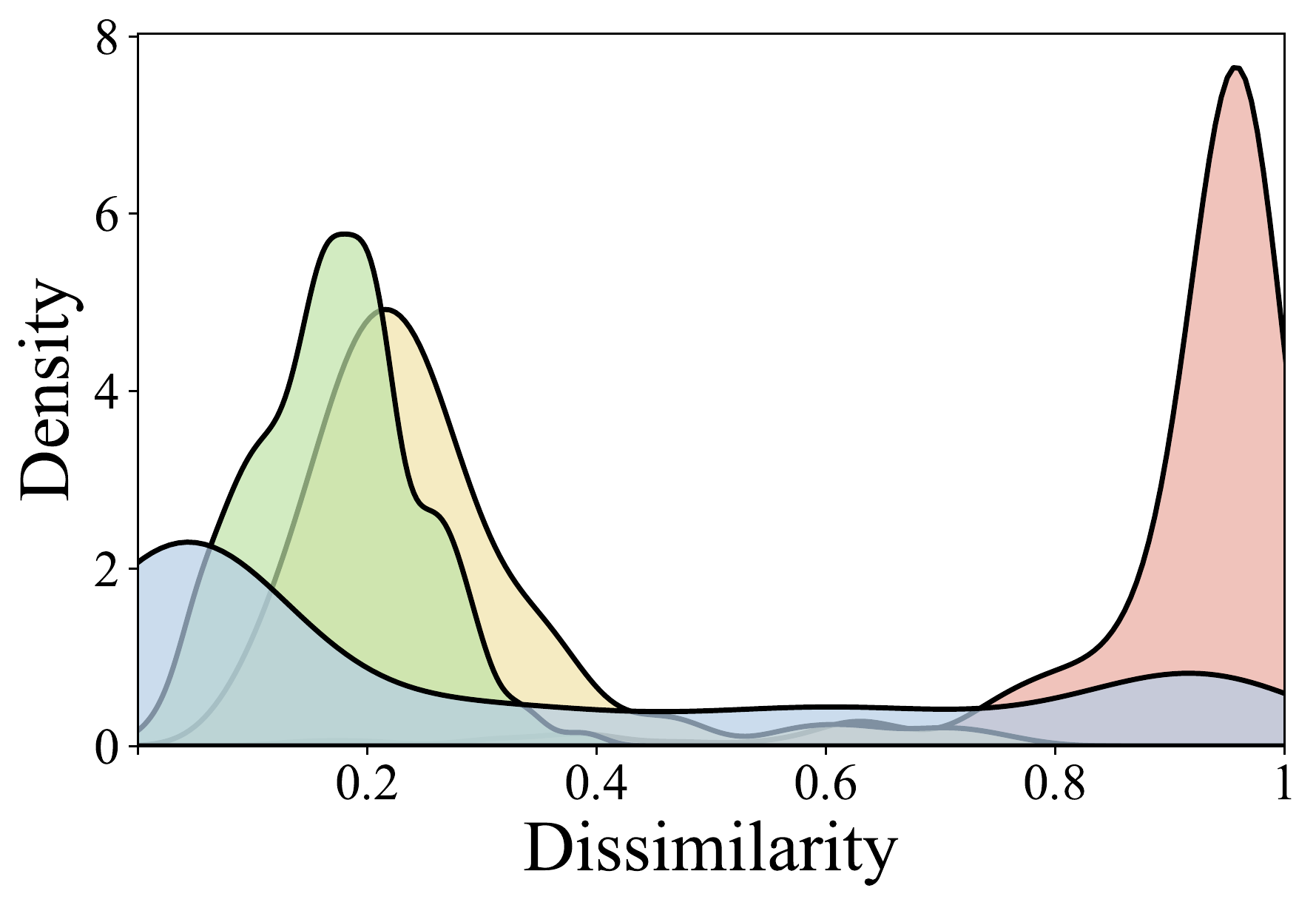}
	}
	\vspace{-0.1cm}
	\subfigure[Tencent]{
	\includegraphics[trim = 3mm 0mm 3mm 0mm, clip,width=0.23\linewidth]{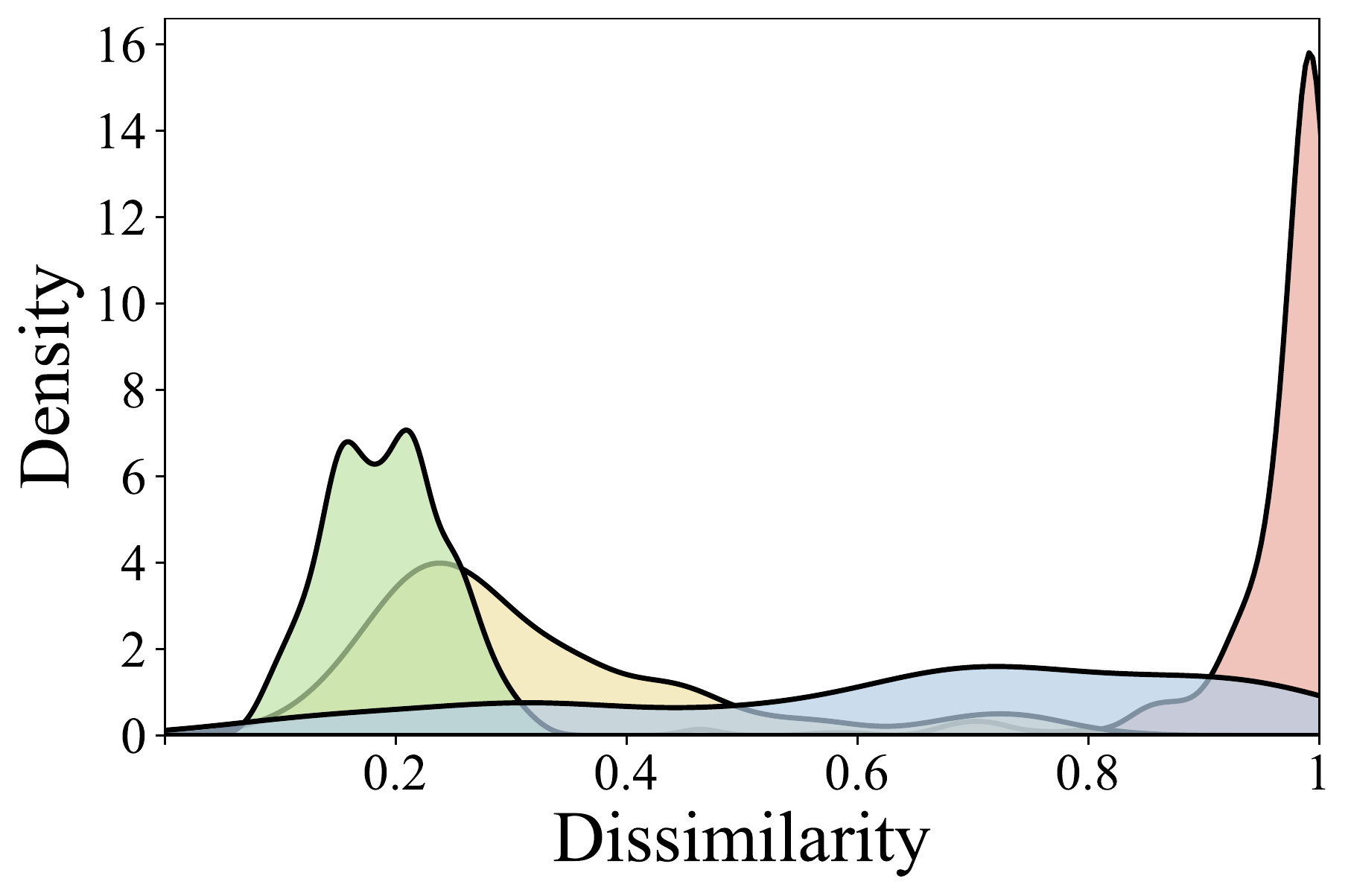}
	}
	\vspace{-0.1cm}

	\caption{The distribution of dissimilarity between fuzzy-words/non-fuzzy-words and real wake-up words. Both \fw and non-\fw have similar distribution of Levenshtein distance, but have distinctively different distributions of our proposed dissimilarity score. }
	\vspace{-0.15in}
	\label{dis}
    \end{figure*}
    
    \begin{figure*}[t]
	\includegraphics[width=0.45\linewidth]{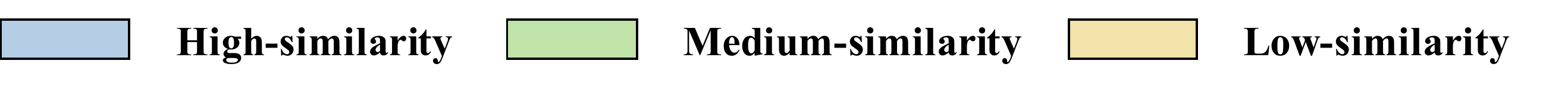}
	\vspace{-1.5em}
	\end{figure*}
	\begin{figure*}[tt]
    \centering
    \subfigure[Amazon Echo]{
	\includegraphics[width=0.23\linewidth]{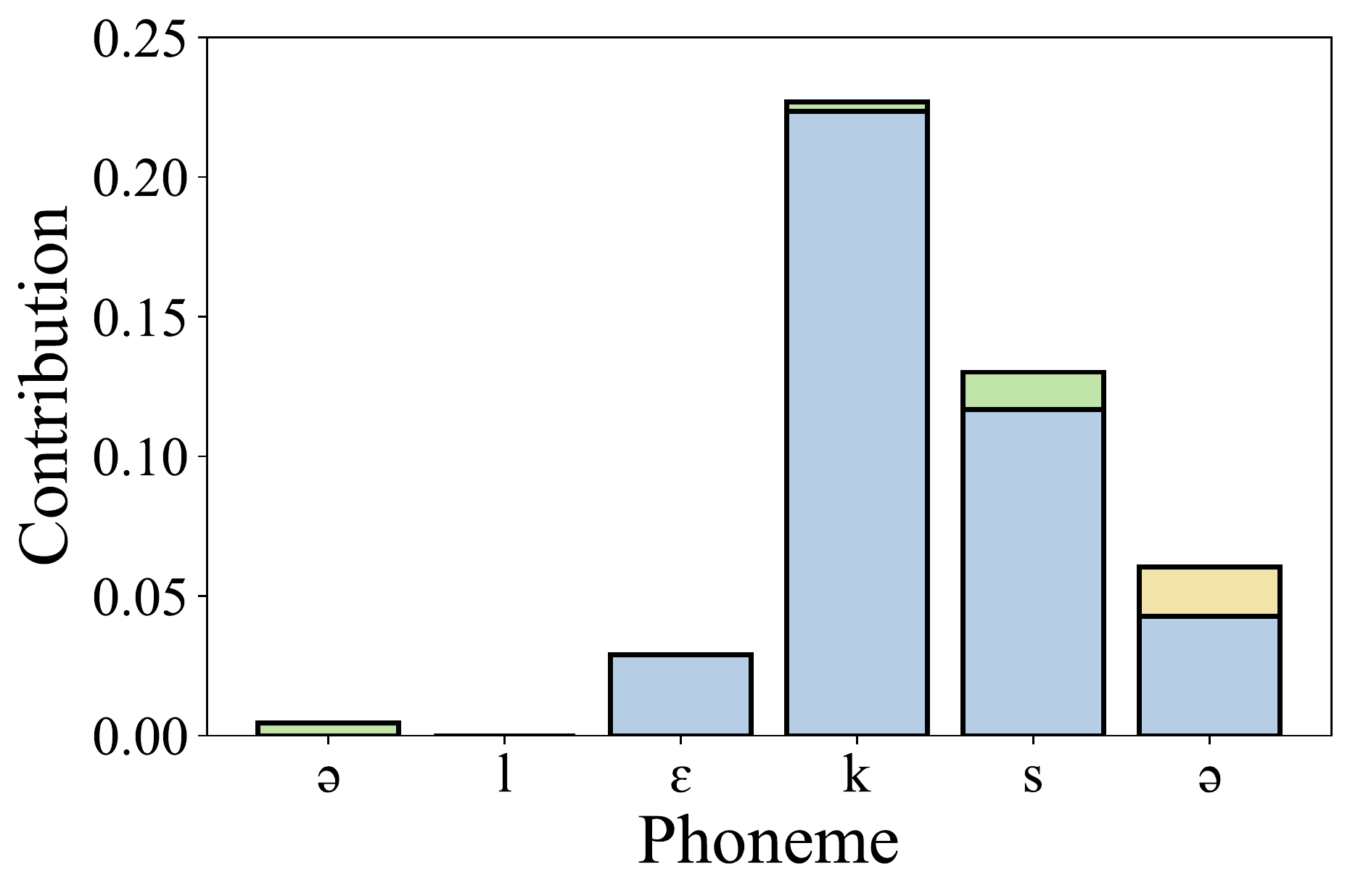}
	\label{expe}
	}
	\vspace{-0.1cm}
	\subfigure[Echo Dot]{
	\includegraphics[width=0.23\linewidth]{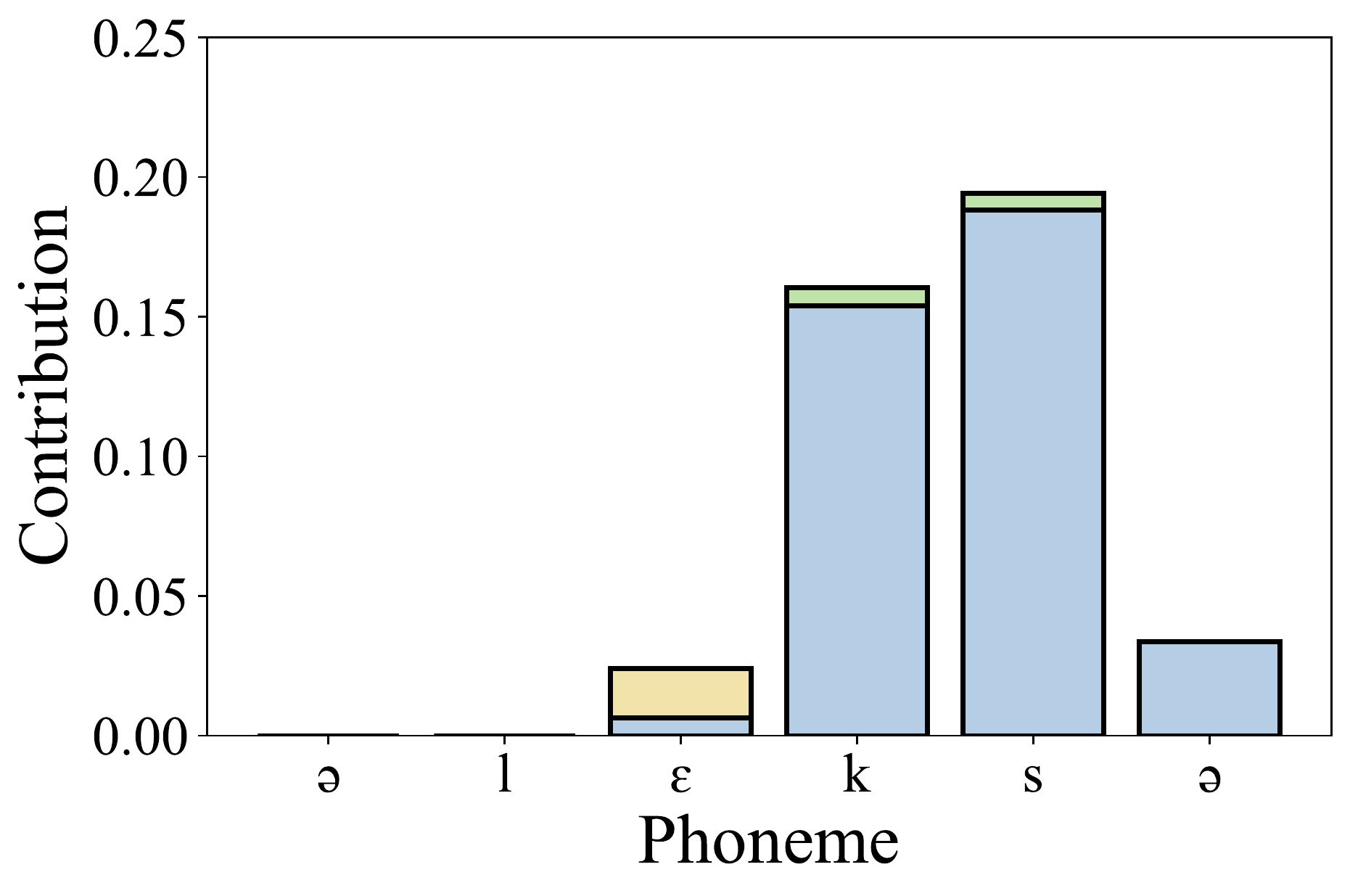}
	}
	\vspace{-0.1cm}
	\subfigure[Google]{
	\includegraphics[width=0.23\linewidth]{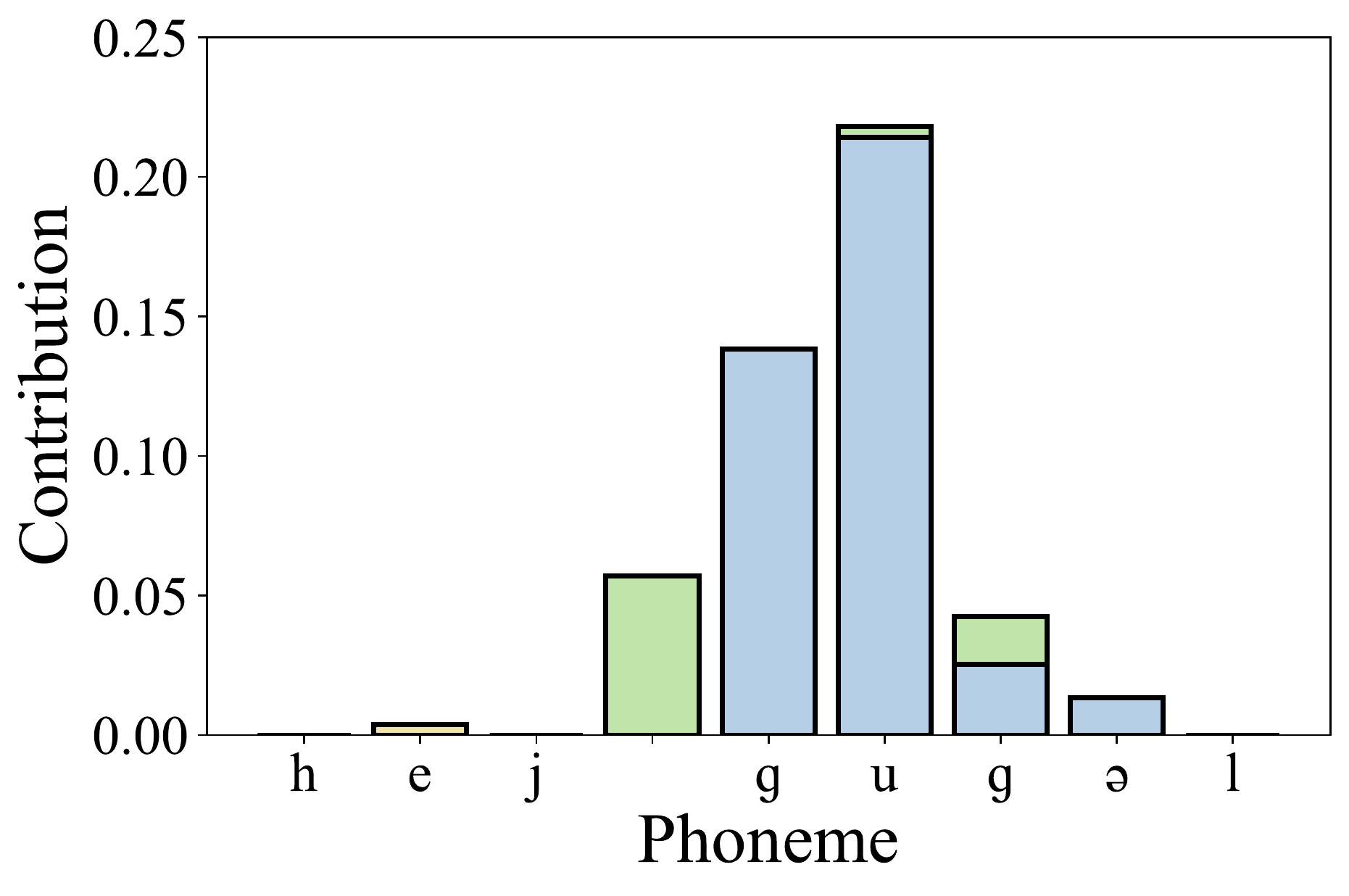}
	}
	\vspace{-0.1cm}
	\subfigure[Apple Siri]{
	\includegraphics[width=0.23\linewidth]{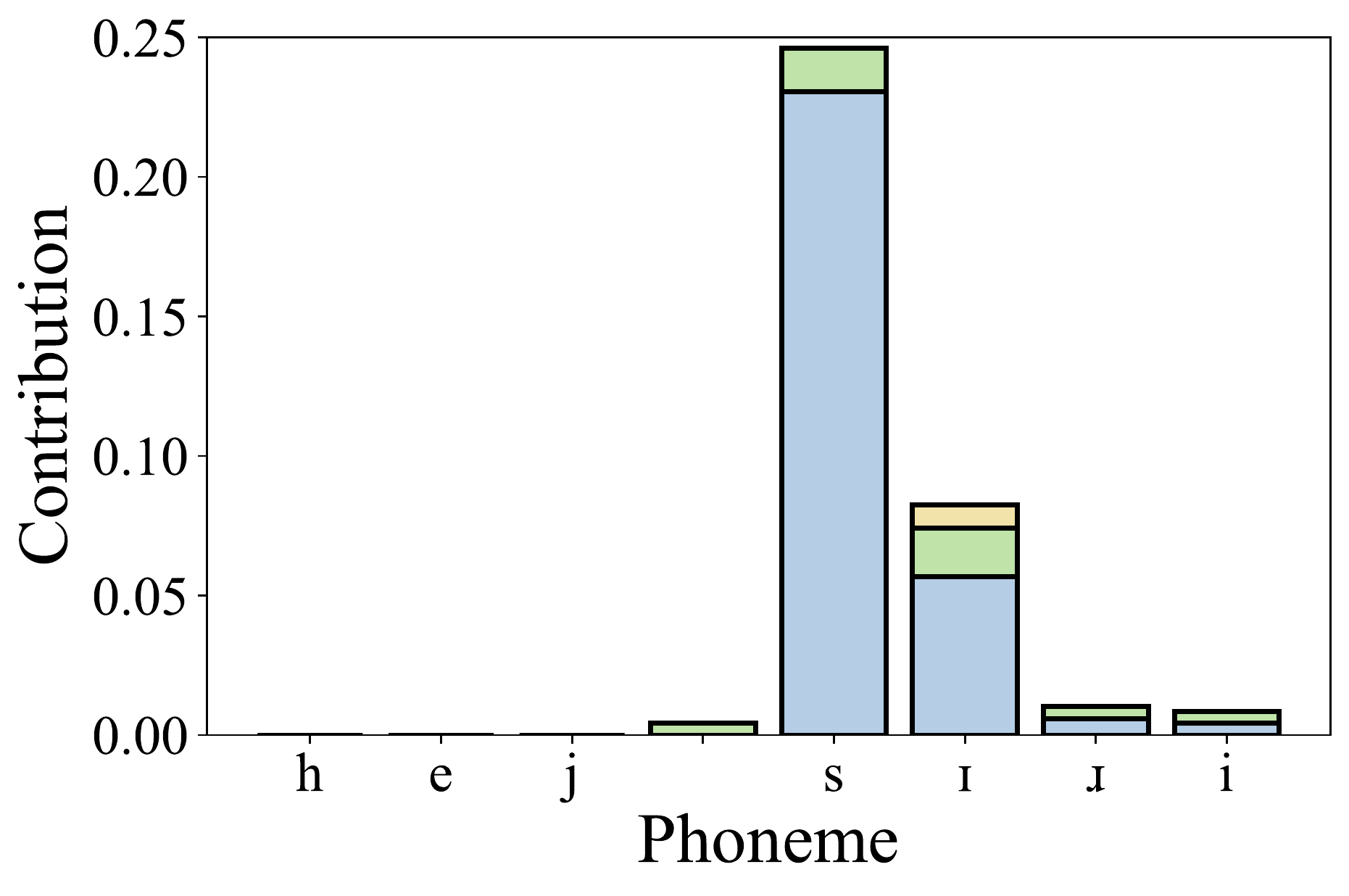}
	}
	\vspace{-0.1cm}
	\subfigure[Baidu]{
	\includegraphics[width=0.23\linewidth]{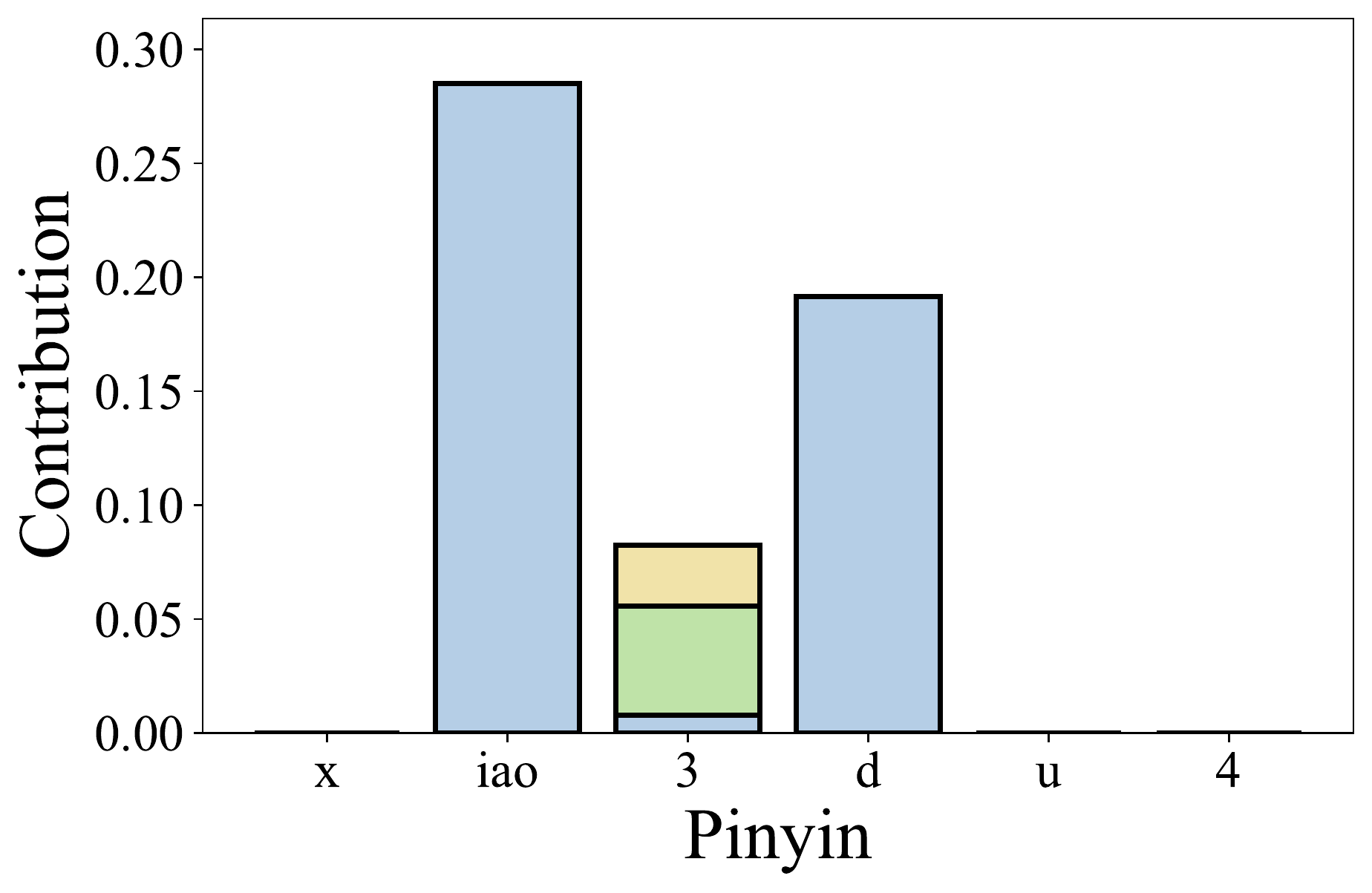}
	\label{expc}
	}
	\vspace{-0.1cm}
	\subfigure[Xiaomi]{
	\includegraphics[width=0.23\linewidth]{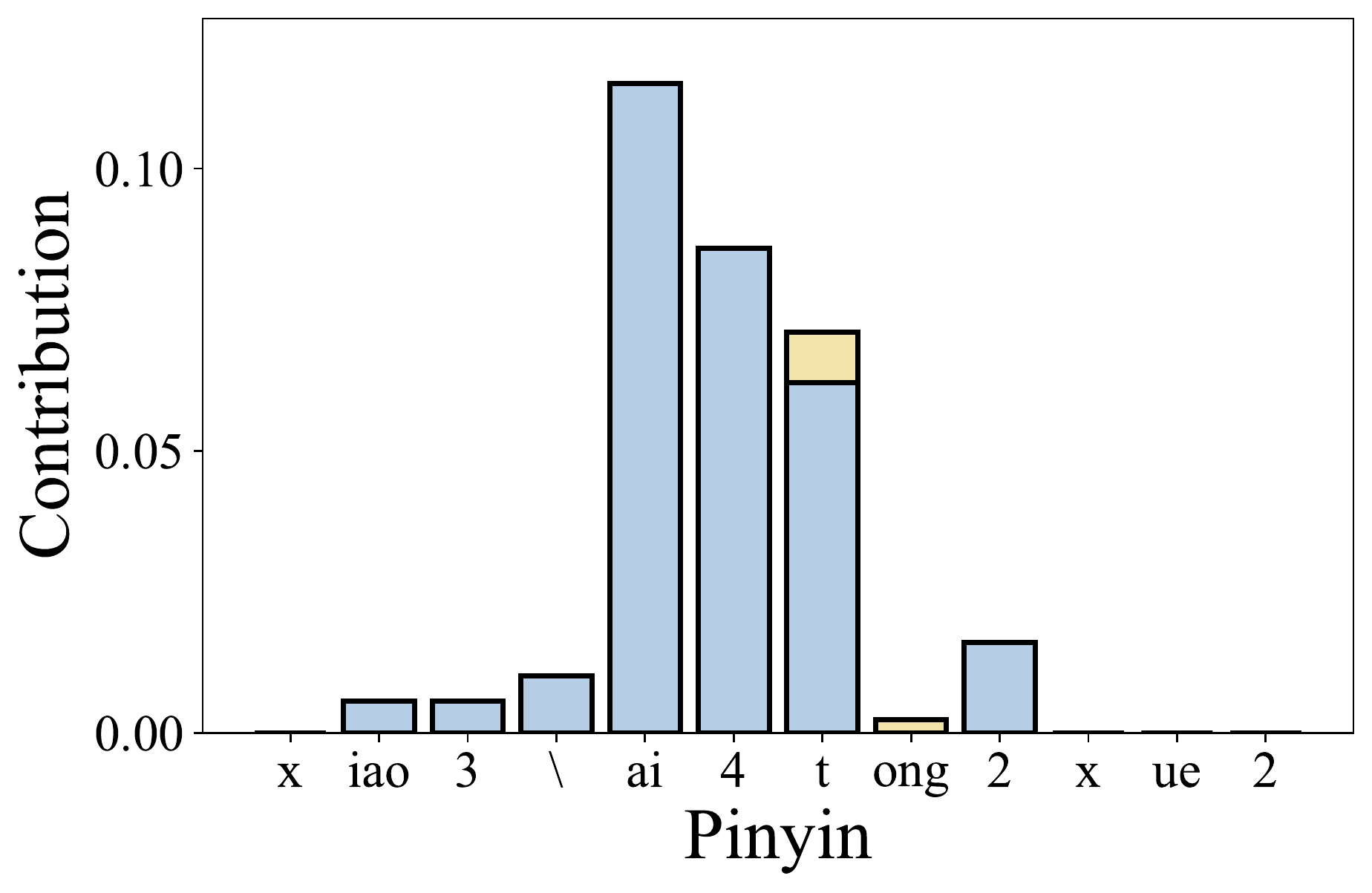}
	}
	\vspace{-0.1cm}
	\subfigure[AliGenie]{
	\includegraphics[width=0.23\linewidth]{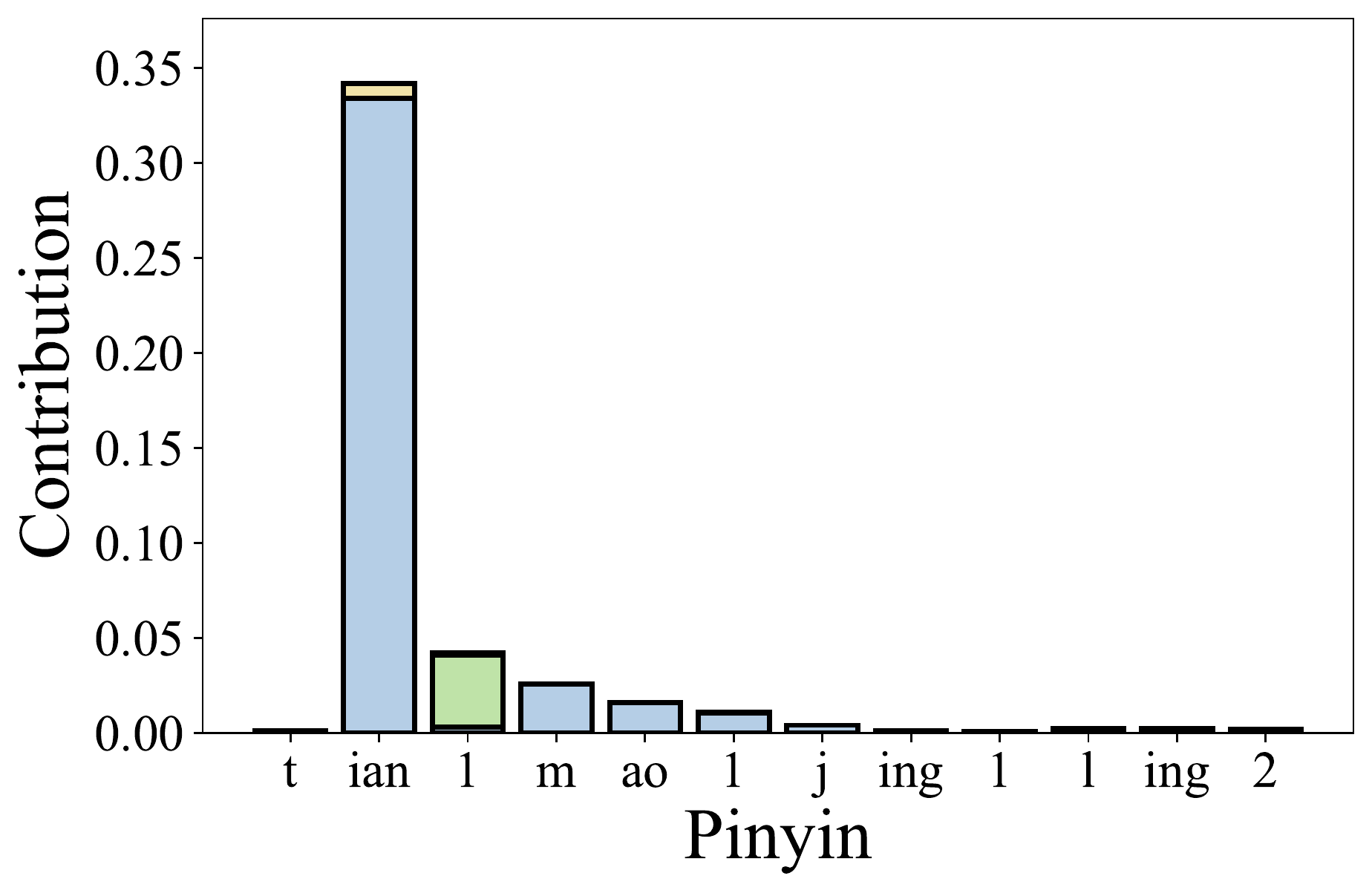}
	}
	\vspace{-0.1cm}
	\subfigure[Tencent]{
	\includegraphics[width=0.23\linewidth]{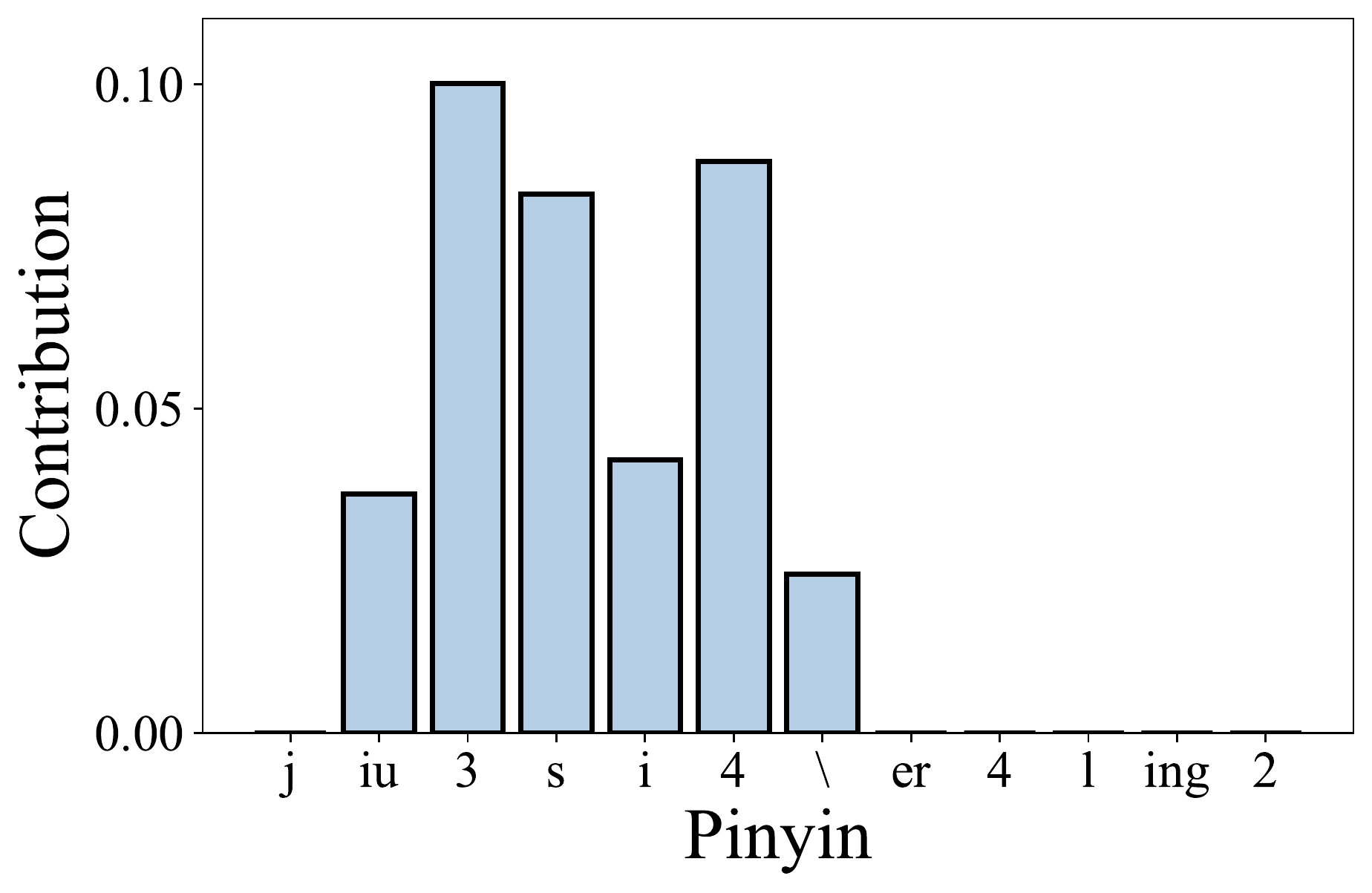}
	}
	\vspace{-0.1cm}

	\caption{The contributions of decisive factors at different positions of the real wake-up word. The legend \emph{High-similarity} refers to decisive factors that are phonetically similar to their counterparts in the real wake-up word. \emph{Medium-similarity} and \emph{Low-similarity} are defined accordingly. }
	\vspace{-0.15in}
	\label{expall}
    \end{figure*}

A fuzzy word is considered as the wake-up word because the wake-up word detector is fooled by certain "similarities" between the two words. Measuring the similarities based on conventional metrics (e.g., calculating the Levenshtein distance between two words~\cite{schonherr2020unacceptable}) may not be appropriate since our generated \fw have an expanded distribution of Levenshtein distance from the real wake-up words (as shown in Figure~\ref{dis}). Therefore, we need to dig deeper into the root causes of false acceptance of \fw. 
	
To address this challenge, we propose to build an interpretable tree-based classifier as a proxy to the inaccessible and inexplicable black-box wake-up word detector. Based on the classifier, we propose a dissimilarity score that can predict whether a word is likely to be accepted by the wake-up word detector or not. After that, we pinpoint phonetic features with the highest contributions to the classification results as decisive factors that have the most influence on the decision of wake-up word detectors.

\textbf{Dissimilarity score}. We train an interpretable model with the generated \fw as positive samples and non-\fw as negative samples, which simulates the behavior of the wake-up word detector. To extract features of each fuzzy word as the input to the model, we do not use the encoding of the genetic algorithm, since it consists only of numbers that do not carry pronunciation information. For Chinese, we leverage the high-dimensional embedding~\cite{dimsim}, which encodes each initial and each final into a two-dimensional vector. This embedding characterizes the phonetic features of a Chinese word. For English, since there is no high-dimensional embedding for phonemes, we use multi-dimensional scaling~\cite{mds} to find an encoding that preserves the dissimilarity between phonemes~\cite{panphon}. We also encode each phoneme into a two-dimensional vector. In summary, each initial/final/phoneme is encoded into two features.
	    
	    
We use gradient boosting classifier~\cite{gbc}, an ensemble of weak classifiers (we use decision tree as the weak classifier), as the interpretable model. We perform 10-fold cross validation, and the results show that the interpretable models have a prediction accuracy of more than 90\% for most voice assistants. The prediction accuracy of models for different voice assistants is in the Appendix. We use the output confidence score of a certain fuzzy word as its similarity score $\gamma$. The dissimilarity score equals $1-\gamma$. 

\begin{tcolorbox}[enhanced, fontupper=\normalsize, left=2pt,right=2pt,top=0pt,bottom=0pt, title = \textbf{Insight 3}]
 Our constructed dissimilarity score can better separate \fw and non-\fw than Levenshtein distance. 
\end{tcolorbox}

Figure~\ref{dis} displays the distribution of dissimilarity scores between fuzzy-words/non-fuzzy-words and real wake-up words. We can observe that \fw and non-\fw have similar expanded distributions in terms of Levenshtein distance, which indicates that Levenshtein distance is not a proper metric to predict whether a word is similar to the real wake-up word. In contrast, with our approach, non-\fw have significantly higher dissimilarity scores than \fw, which means that our constructed dissimilarity score can better predict whether a word is likely to be accepted by the wake-up word detector.   

\textbf{Decisive factors.} In order to evaluate the contribution of each feature, we calculate their SHAP values~\cite{shap}. We regard each feature as a "contributor", and add up contributions of all features to obtain the prediction result. 
	    
	    \begin{equation}
	        y_i = y_0 + \sum_j f(\alpha_i^{(j)}),
	        \label{sp}
	    \end{equation}
where $y_i$ is the prediction result for sample $x_i$, $y_0$ is the mean prediction results of all samples, $\alpha_i^{(j)}$ is the $j$-th feature of $x_i$, and $f(\alpha_i^{(j)})$ is the contribution of $\alpha_i^{(j)}$ to the prediction result. 
	    
Note that the contribution $f(\alpha_{i}^{(j)})$ can be positive or negative. A positive contribution means that the feature is helpful for accurate prediction of whether a sample is a fuzzy word or not. For a certain fuzzy word $x_i$, let $\mathcal{A}^+_i$ denote the set of features with positive contributions. We rank the elements in $\mathcal{A}^+_i$ according to their contributions in a non-increasing order. We construct a set $\mathcal{D}_i$ to represent important features for $x_i$, and then add elements in $\mathcal{A}^+_i$ to $\mathcal{D}_i$ sequentially until the ratio of the contribution of all elements in $\mathcal{D}_i$ to the contribution of all elements in $\mathcal{A}^+_i$ is more than a threshold $\beta$. 

	    \begin{equation}
	       \frac{ \sum_{k \in \mathcal{D}_i} f(\alpha_{i}^{(k)})}{\sum_{k' \in \mathcal{A}^+_i} f(\alpha_{i}^{(k')})} \ge \beta.
	        \label{df}
	    \end{equation}
We consider a(n) initial/final/phoneme as a decisive factor for $x_i$ if at least one of its two features is in $\mathcal{D}_i$. In this way, we can obtain the set of decisive factors and their contributions. The contribution of a(n) initial/final/phoneme is the sum of contributions of its features in $\mathcal{D}_i$.

We construct the set of decisive factors for each fuzzy word that is correctly classified by the interpretable model. We compare the decisive factors and their counterparts in the original wake-up word by computing the absolute differences of their encoding. We obtain mean and variance $\delta$ of all computed differences for normalization. According to the similarity measurement, we category the decisive factors into three groups: high-similarity (the normalized difference is within $\delta$), medium-similarity (the normalized difference is within $2\delta$), and low-similarity (the normalized difference is beyond $2\delta$). After that, we compute the average contribution of each group of decisive factors at each position of the original wake-up word, as shown in Figure~\ref{expall}. The threshold $\beta$ is set as 0.8. 

\begin{tcolorbox}[enhanced, fontupper=\normalsize, left=2pt,right=2pt,top=0pt,bottom=0pt, title = \textbf{Insight 4}]
  Wake-up words with fewer decisive factors have more \fw. 
\end{tcolorbox}

Figure~\ref{expall} demonstrate that decisive factors that determine false acceptance of \fw usually concentrate on a short snippet of the wake-up word, e.g., \emph{ks} for Alexa and \emph{\`ai} for xiǎo ài tóng xué. By keeping the decisive factors and alter other parts of the wake-up word may help create a new fuzzy word. If a wake-up word has fewer decisive factors, there is more space for altering other parts to generate \fw. For instance, AliGenie relies disproportionately on \emph{ian} to recognize the wake-up word, thus the number of its \fw is the highest among all voice assistants.

\section{Mitigating \FW}\label{sec:defend}
In this section, we present two potential approaches to strengthen wake-up word detectors against \fw. The first approach is to scrutinize words that contain decisive factors. In Section \ref{sec:understand}, we have revealed decisive factors that contribute the most to false acceptance of \fw by wake-up word detectors. We quantify the proportion of generated \fw that contain decisive factors, as shown in Table~\ref{exp}. We can observe that for most voice assistants, more than 90\% of \fw contain at least one of the top-3 decisive factors. In particular, 100\% of the \fw of Google and Baidu contain top-3 decisive factors. The wake-up word detector can send audio samples that involve decisive factors to more discreet examination by complicated speech recognition models.

	\begin{table}[tt]
	\centering
	\footnotesize
	\caption{Fraction of \fw containing decisive factors. The column \emph{top-n} presents the proportion of \fw that comprise at least one of the top-n decisive factors with the highest contribution.}
	\begin{tabular}{cccc||cccc}
		\hline
	 & top-1 & top-2 & top-3&& top-1 & top-2 & top-3 \\
		\hline
		 Echo & 85.6\% & 89.9\% & 90.7\% &Baidu & 68.8\% & 76.6\% & 100.0\% \\
		Echo Dot & 83.7\% & 88.9\% & 88.9\% & Xiaomi & 65.7\% & 65.7\% & 65.7\% \\
		Google & 90.7\% & 90.7\% & 100.0\% &AliGenie & 61.8\% & 82.6\% & 98.8\%\\
		Apple & 87.5\% & 87.5\% & 95.3\% &
        Tencent & 60.7\% & 83.3\% & 98.8\% \\
		\hline
		\end{tabular}
		\label{exp}
		
	\end{table}
	
The second remedy is to strengthen wake-up word detectors by retraining them with the generated \fw. Since we have no access to wake-up word detectors of commercial voice assistants, we test our idea on a lightweight, open-source keyword-spotting model, \textsc{Mycroft Precise}~\cite{mycroftprecise}. We also experiment with \textsc{TC-ResNet}~\cite{choi2019temporal}, another open-source keyword-spotting model, but the trained model has poor generalization ability on our datasets. The prediction accuracy on the test dataset is as low as 85.64\%.
For a specific wake-up word, we collect three types of datasets: a conventional dataset, a fuzzy word dataset, and a collective dataset. The conventional dataset consists of samples of the wake-up word (positive) and non-\fw (negative). We divide the conventional dataset into a training dataset and a test dataset. To begin with, we train an original wake-up word detector using the conventional training dataset on \textsc{Precise}. 
The fuzzy word dataset consists of the generated \fw of the wake-up word. We use the fuzzy word dataset to retrain the original model to obtain the strengthened model. The collective dataset consists of a large number of words from a dictionary with potential \fw. The collective dataset has no overlap with the conventional dataset and the fuzzy word dataset. We test the original model and the strengthened model on the conventional test dataset to obtain the false positive rate, false negative rate, and accuracy. We further test the original model and the strengthened model on the collective dataset to measure the fuzzy rate, which is defined as as the ratio of wrongly accepted words among all words in the collective dataset. A larger fuzzy rate means that more \fw in the dictionary are accepted by the detector, which indicates that the model is more vulnerable to \fw.

Since we do not have access to the training datasets of the wake-up word detectors of commercial voice assistants, we use public datasets as the conventional dataset to train wake-up word detectors. We obtain five public datasets of English words, including "Alexa", "Athena", "Computer", "Hi Xiaowen" and "Hi Mia". Unfortunately, we did not find public datasets of Chinese words. We split each conventional dataset into a training dataset with 3/4 of the samples and a test dataset with the remaining samples. 
We keep a relatively balanced ratio of negative samples to positive samples in the training dataset to avoid bias. For Alexa's training dataset, we have 296 positive samples and 399 negative samples. The positive samples are collected from kaggle~\cite{alexa-dataset} and the negative samples are collected from the human noise and the environment noise in the dataset of  mycroft-precise~\cite{mycroftprecise}. For Athena's training dataset, we have 386 positive samples and 839 negative samples. For Computer's training dataset, we have 87 positive samples and 161 negative samples. We collect the data of Athena and Computer from mycroft-precise~\cite{mycroftprecise}. For Hi Xiaowen's training dataset, we have 380 positive samples and 693 negative samples, all collected from MobvoiHotwords~\cite{DBLP:journals/spl/HouSOHX19}. For Hi Mia's training dataset, we have 800 positive samples and 1,000 negative samples, all collected from AISHELL~\cite{qin2020hi}. For the fuzzy word dataset, we search for \fw using a dictionary of 498 words from the Google TTS library~\cite{pypi}. The collective dataset is also from the Google TTS library with 49,822 words. 

As shown in Table~\ref{tab: mitigation result}, the fuzzy rate of the original model can be as high as more than 20\%, but drops to almost 0 after being strengthened. Take Alexa as an example. 22.89\% of the words in the collective dataset are wrongly accepted by the original wake-up word detector. In contrast, the strengthened model only accepts 0.25\% of the words in the collective dataset. Surprisingly, for all strengthened models, their accuracy is even higher than the original model on the conventional test dataset. This shows that our mitigation approach not only defends against \fw but also improves the performance of the original model. The possible reason is that the \fw are close to the decision boundary, and using \fw to train the detector provides a more effective way to learn the decision boundary between wake-up words and other words.
In the future, we aim to evaluate our mitigation method on commercial wake-up word detectors that are trained with large-scale audio samples to unveil potential limitations of this mitigation approach.

	\begin{table}[tt]
	\footnotesize
	\caption{Mitigation results}
	\vspace{-0.1in}
	\centering
	\renewcommand\arraystretch{1.2}
	\begin{tabular}{cccc}
		\toprule[1.5pt]
		Wake-up & Evaluation & Original & Strengthened \\
		word & metric & model & model \\
		\midrule
		\midrule
		\multirow{4}[0]{*}{Alexa} & False positive rate & 0.00\% & 0.00\% \\
		& False negative rate & 6.85\% & 5.48\% \\
		& Detection accuracy & 96.73\% & 97.39\% 
		\\
		\cellcolor{white} & \cellcolor{gray!10} \textbf{Fuzzy rate} & \cellcolor{gray!10} \textbf{22.89\%} & \cellcolor{gray!10} \textbf{0.25\%} \\
		\midrule
		\multirow{4}[0]{*}{Computer} & False Positive rate& 0.00\% & 0.00\% \\
		& False negative rate & 5.00\% & 0.00\% \\
		& Detection accuracy & 97.96\% & 100\% 
		\\
		\cellcolor{white} & \cellcolor{gray!10} \textbf{Fuzzy rate} & \cellcolor{gray!10} \textbf{10.04\%} & \cellcolor{gray!10} \textbf{0.02\%} \\
		\midrule
		\multirow{4}[0]{*}{Athena} & False Positive rate& 1.37\% & 0.88\% \\
		& False negative rate& 1.04\% & 1.04\% \\
		& Detection accuracy & 98.73\% & 99.07\% 
		\\
		\cellcolor{white} & \cellcolor{gray!10} \textbf{Fuzzy rate} & \cellcolor{gray!10} \textbf{17.67\%} & \cellcolor{gray!10} \textbf{1.18\%} \\
		\midrule
		\multirow{4}[0]{*}{Hi Xiaowen} & False Positive rate& 0.00\% & 0.00\% \\
		& False negative rate& 2.13\% & 2.13\% \\
		& Detection accuracy & 99.36\% & 99.36\% 
		\\
		\cellcolor{white} & \cellcolor{gray!10} \textbf{Fuzzy rate} & \cellcolor{gray!10} \textbf{11.65\%} & \cellcolor{gray!10} \textbf{0.20\%} \\
		\midrule
		\multirow{4}[0]{*}{Hi Mia} & False Positive rate& 0.00\% & 0.00\% \\
		& False negative rate& 0.00\% & 0.00\% \\
		& Detection accuracy & 100\% & 100\% 
		\\
		\cellcolor{white} & \cellcolor{gray!10} \textbf{Fuzzy rate} & \cellcolor{gray!10} \textbf{5.62\%} & \cellcolor{gray!10} \textbf{0.00\%} \\
		\bottomrule[1.5pt]
	\end{tabular}
	\vspace{-0.1in}
	\label{tab: mitigation result}%
\end{table}%

	
	\section{Related Work}
	
	
	\textbf{Wake-up word detection}.
	Various speech recognition or keyword-spotting models have been proposed. Earlier works leverage Hidden Markov Models (HMM) in connection with Gaussian Mixture Models (GMM) for acoustic modeling~\cite{garcia2006keyword, zhang2009unsupervised}, which suffers from high computational complexity. Hence, the traditional HMM-GMM models are now replaced by more efficient Deep Neural Networks (DNN)~\cite{panchapagesan2016multi, sun2017compressed}, Convolutional Neural Networks (CNN)~\cite{chen2014small, sainath2015convolutional}, Recurrent Neural Networks (RNN)~\cite{fernandez2007application}, Convolutional Recurrent Neural Networks (RCNN)~\cite{arik2017convolutional}, Gated Recurrent Units (GRUs)~\cite{woellmer2013keyword}, and Long Short-Term Memory
	(LSTM) units~\cite{baljekar2014online}. 
	
	
	\textbf{Wake-up word security}.
    As far as we know, there is only a few works on wake-up word security. Schönherr et al.~\cite{schonherr2020unacceptable} investigated \emph{accidental triggers} for English, Chinese and German smart speakers. Accidental triggers were exhaustively searched by playing a large corpus of media audios to smart speakers, following which a dictionary is used to find more accidental triggers based on Levenshtein distance. The Chinese speakers are tested with English audio samples, and the influence of distance, volume, speed and ambient noises are not considered. Similarly, Dubois et al.~\cite{WhenSpeakersAreAllEarsCharacterizingMisactivationsofIoTSmartSpeakers} played UK and US television shows to voice assistants to characterize the sources and the number of mis-activations. Mitev et al.~\cite{mitev2020leakypick} utilized a phoneme dictionary to fuzz voice assistants with words of a similar phoneme length and proposed to scan network traffic for mis-activated devices. All existing methods seek for \fw by exhaustive search, which takes days or even weeks to find a limited set of \fw.

    \textbf{Covert commands}.
	Covert command attacks aim to inject malicious commands for the voice assistant to execute certain operations without users noticing. Vaidya et al.~\cite{vaidya2015cocaine} proposed to modify the MFCCs of voice samples (e.g., OK Google) so that the mangled samples are unintelligible to humans (e.g., electronic-sounding noise) but will be identified as a command by voice assistants. Carlini et al.~\cite{carlini2016hidden} built on top of this work by presenting an attack on a voice recognition system where the underlying mechanics are known, resulting in a more precise attack. Carlini et al.~\cite{carlini2018audio} created an audio sample from another one containing a spoken sentence, preserving similar waveforms but misleading voice recognition algorithms to yield false interpretations of a different sentence. As electronic-sounding noises may irritate legitimate users, Schönherr et al.~\cite{schonherr2018adversarial} and Yuan et al.~\cite{yuan2018commandersong} proposed methods to insert intended commands inside an innocent audio sample, e.g., a  music file, which is only recognizable by voice recognition algorithms but not by humans. Another approach is proposed by Zhang et al.~\cite{zhang2017dolphinattack} which used ultrasonic audio to inaudibly inject commands into voice assistants.
	
	Our work is different from audio adversarial examples since \fw are natural audio samples but not synthetic ones with carefully-crafted noises. Compared with music~\cite{schonherr2018adversarial, yuan2018commandersong} or electronic-sounding noises~\cite{vaidya2015cocaine, carlini2016hidden}, \fw are more covert and raise less alarm. Our work is different from dolphin attacks~\cite{zhang2017dolphinattack} as the \fw are audible. We do not intend to activate voice assistants using out-of-band voice signals. The frequency range of our played \fw lies within the hearing range of human ears.

	\section{Conclusion}
   As smart devices are increasingly produced with embedded voice assistants, there is a growing need to understand the susceptibility of their wake-up word detectors to misclassifiying words. 
   In this paper, we present a systematic framework to generate, understand and mitigate \fw that can falsely activate voice assistants. Using our search framework customized for different languages, we have managed to pinpoint 965 \fw covering 8 most popular English and Chinese smart speakers. Our explanation of decisive factors and mitigation methods help strengthen wake-up word detectors against \fw. As such, our approach presents a promising way to find, understand and mitigate privacy and security issues in voice assistants.
	

\bibliographystyle{ACM-Reference-Format}
\bibliography{ccs-sample}
\clearpage
\onecolumn

\appendix
\section{Appendix}

\subsection{Impact of Speaker Gender}

\begin{figure*}[h]
    \centering
    	\begin{minipage}[b]{0.97\linewidth}
    		\centering
    		\includegraphics[trim=0mm 0mm 0mm 0mm, clip, width=0.3\textwidth]{Picture/Ablation_new/legend.pdf}
    		\vspace{-0.2cm}
    	\end{minipage}
    \subfigure[Amazon Echo: Gender]{
    	\begin{minipage}[b]{0.3\linewidth}
    		\centering
    		\includegraphics[trim=0mm 0mm 0mm 0mm, clip, width=0.95\textwidth]{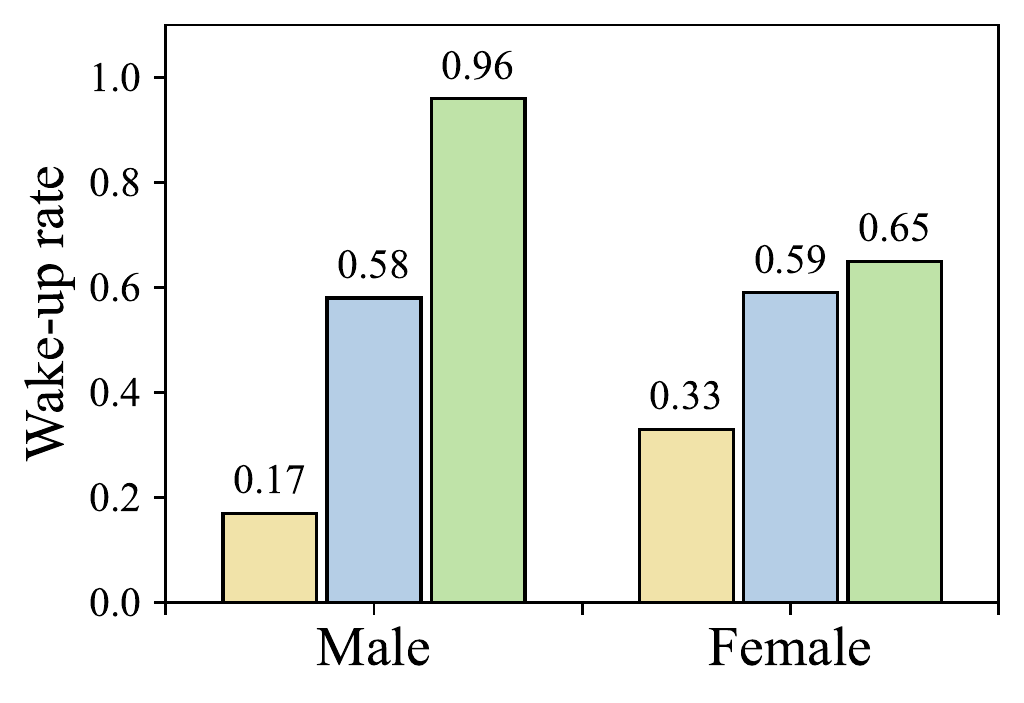}
    		\label{fig:ver_2figs_2cap_1}
    		\vspace{-0.2cm}
    	\end{minipage}
    }
    \vspace{-0.1cm}
    \subfigure[Echo Dot: Gender]{
    	\begin{minipage}[b]{0.3\linewidth}
    		\centering
    		\includegraphics[trim=0mm 0mm 0mm 0mm, clip,width=0.95\textwidth]{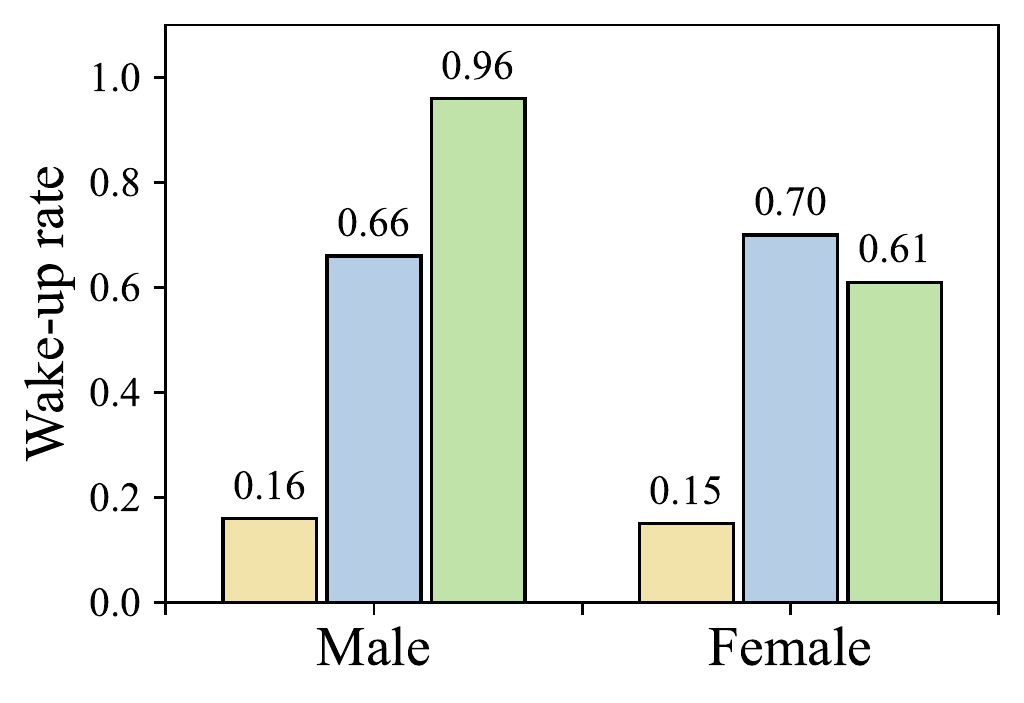}
    		\label{fig:ver_2figs_2cap_2}
    		\vspace{-0.2cm}
    	\end{minipage}
    }
    \vspace{-0.1cm}
    \subfigure[Google: Gender]{
    	\begin{minipage}[b]{0.3\linewidth}
    		\centering
    		\includegraphics[trim=0mm 0mm 0mm 0mm, clip,width=0.95\textwidth]{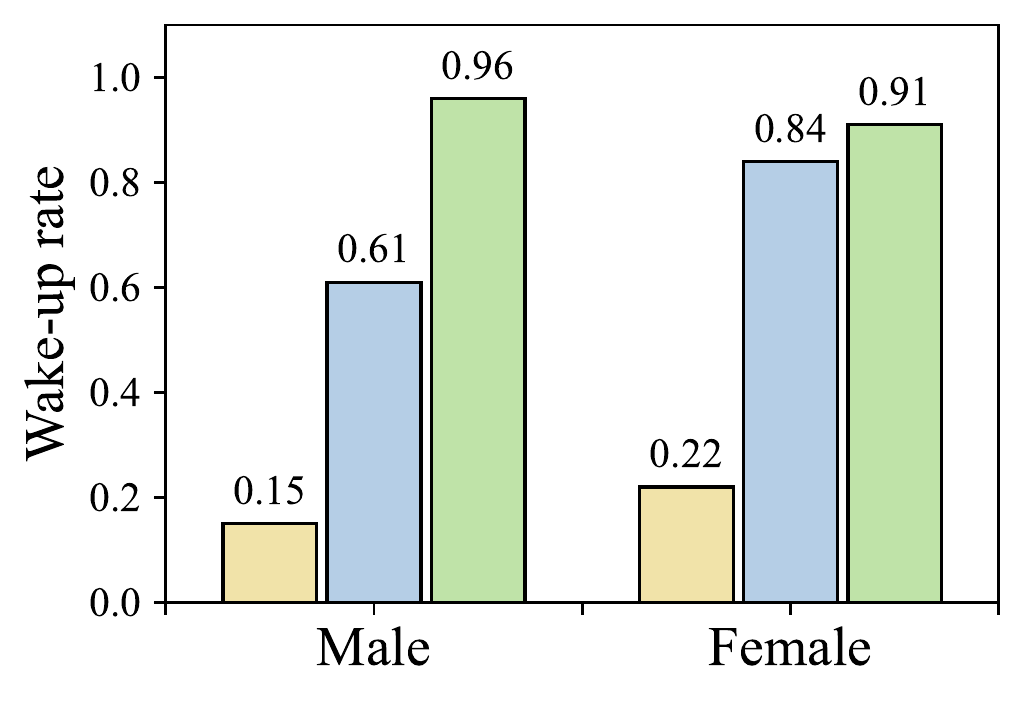}
    		\label{fig:ver_2figs_2cap_2}
    		\vspace{-0.2cm}
    	\end{minipage}
    }
    \vspace{-0.1cm}
    
	\caption{Wake-up rate of \fw under different speaker genders for English voice assistants.}
	\label{fig:Ablation_results_English}
\end{figure*}

\subsection{Subjective Tests}

	\begin{figure*}[h]
    \centering
    \begin{minipage}[b]{0.97\linewidth}
    		\centering
    		\includegraphics[trim=0mm 0mm 0mm 0mm, clip, width=0.3\textwidth]{Picture/User_study/legend-user.pdf}
    	\end{minipage}
    \subfigure[Amazon Echo]{
	\includegraphics[trim = 3mm 4mm 8mm 7mm, clip, width=0.23\linewidth]{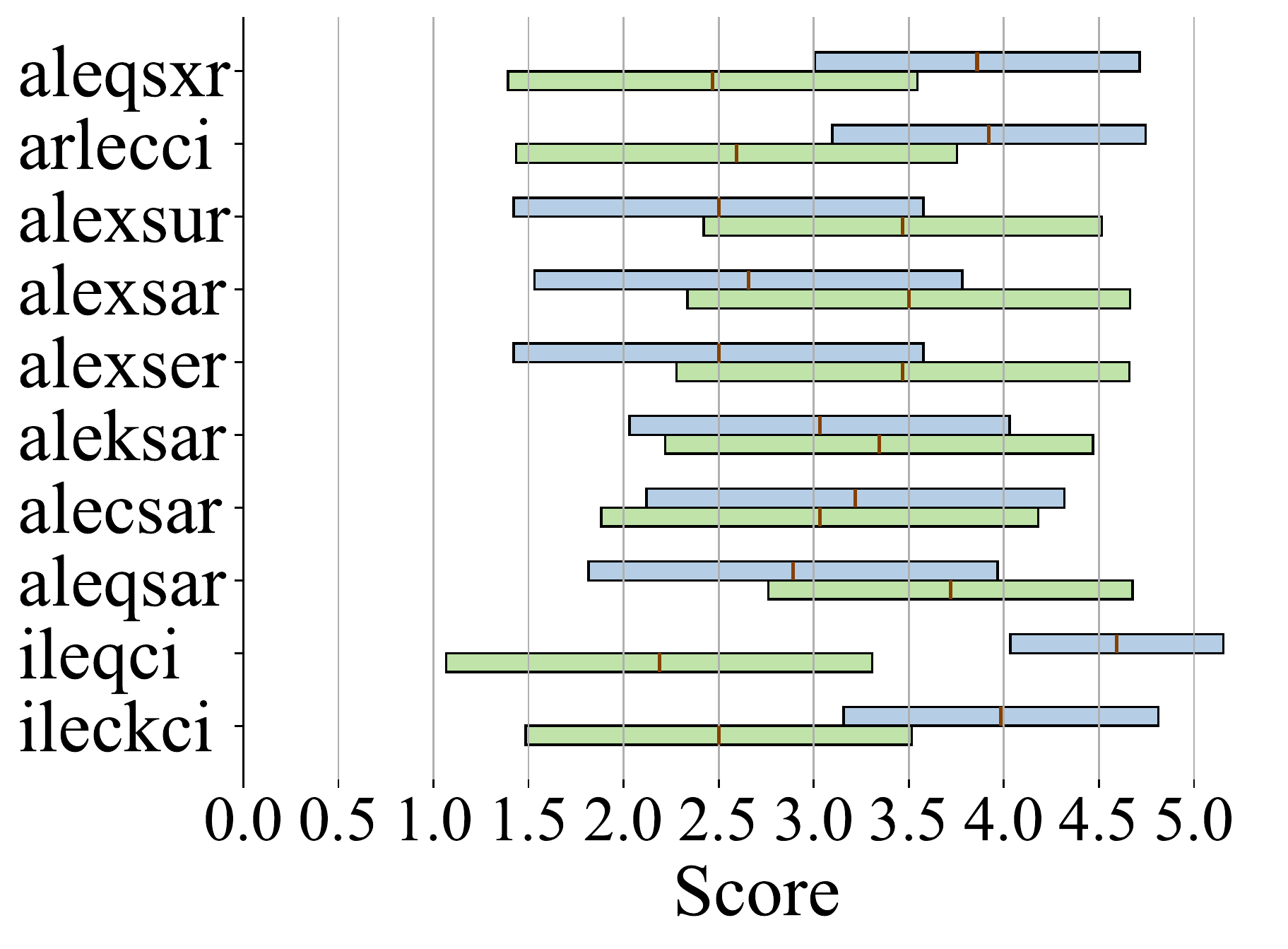}
	}
	\subfigure[Echo Dot]{
	\includegraphics[trim = 3mm 4mm 8mm 7mm, clip,width=0.23\linewidth]{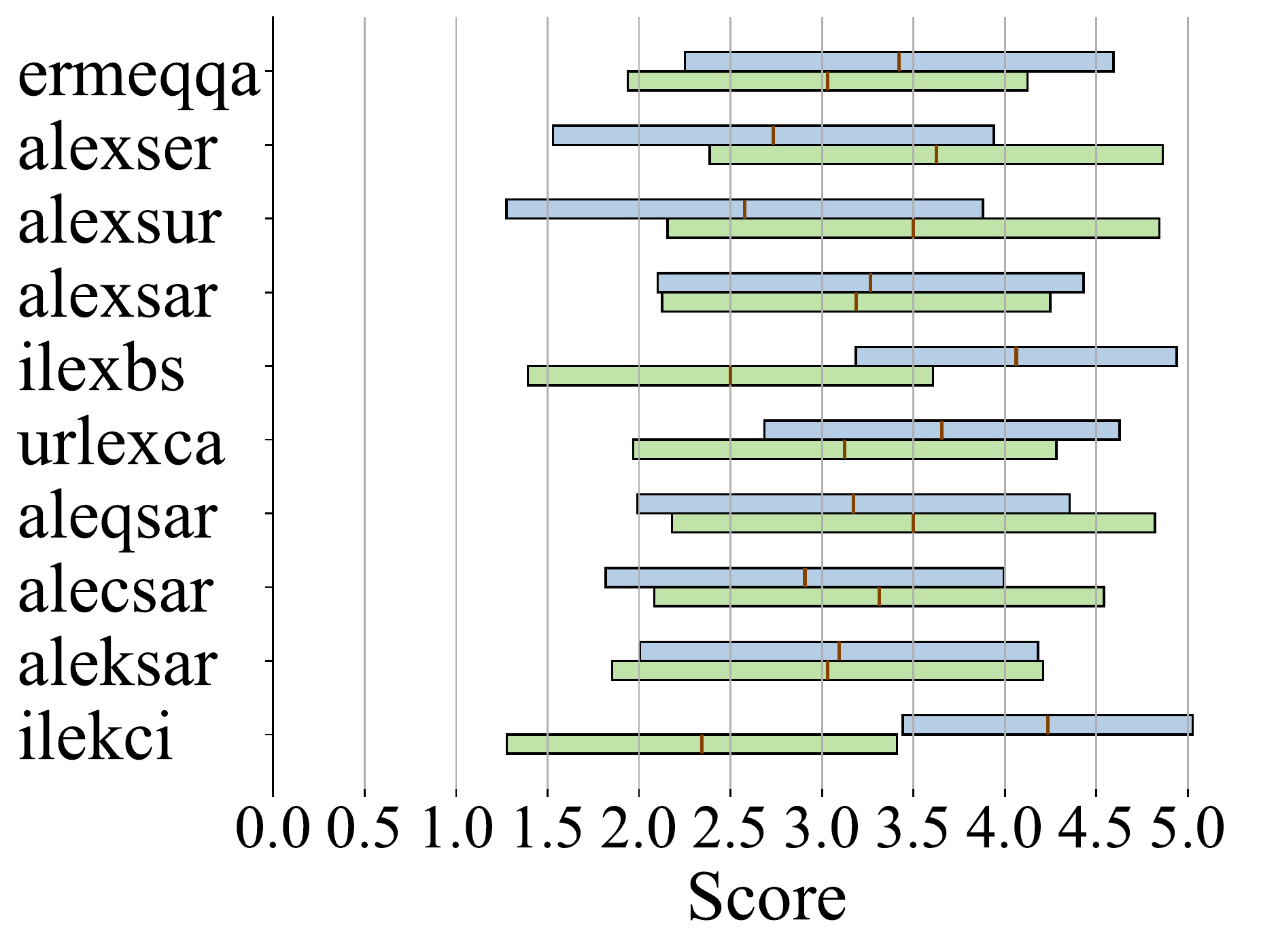}
	}
	\subfigure[Google]{
	\includegraphics[trim = 3mm 4mm 8mm 7mm, clip,width=0.23\linewidth]{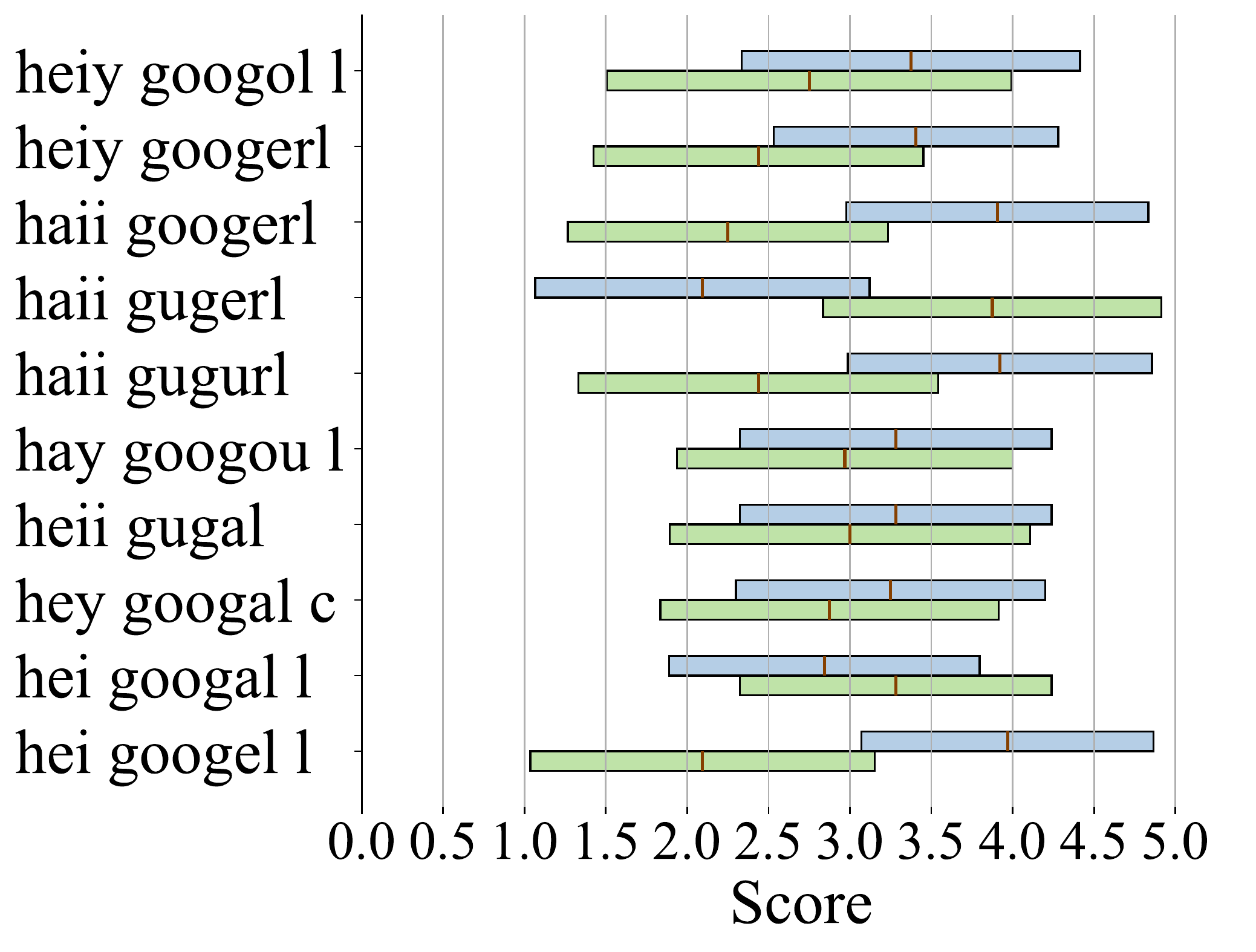}
	}
	\subfigure[Apple Siri]{
	\includegraphics[trim = 3mm 4mm 8mm 7mm, clip,width=0.235\linewidth]{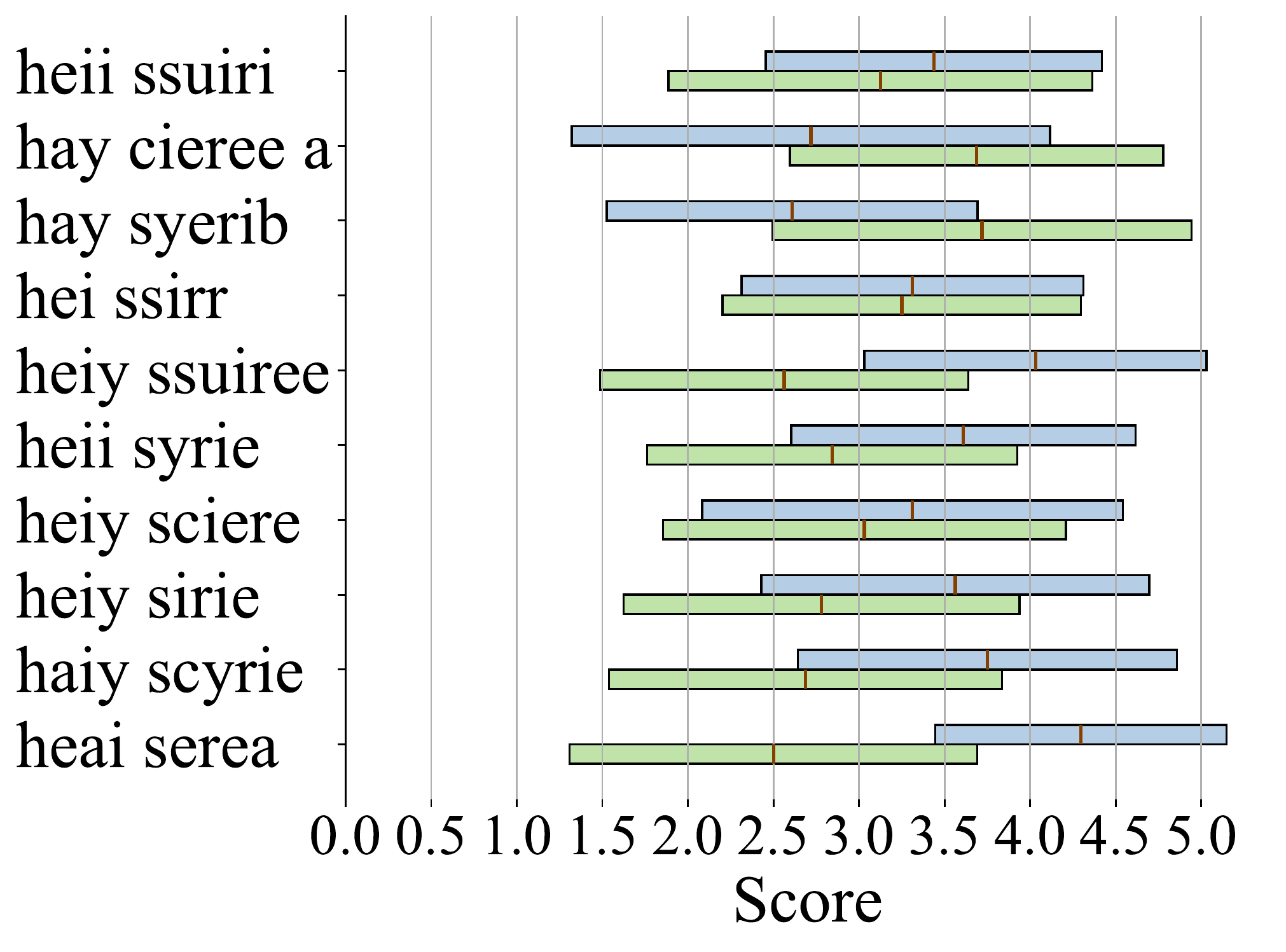}
	}
	\subfigure[Baidu]{
	\includegraphics[trim = 3mm 4mm 8mm 7mm, clip,width=0.23\linewidth]{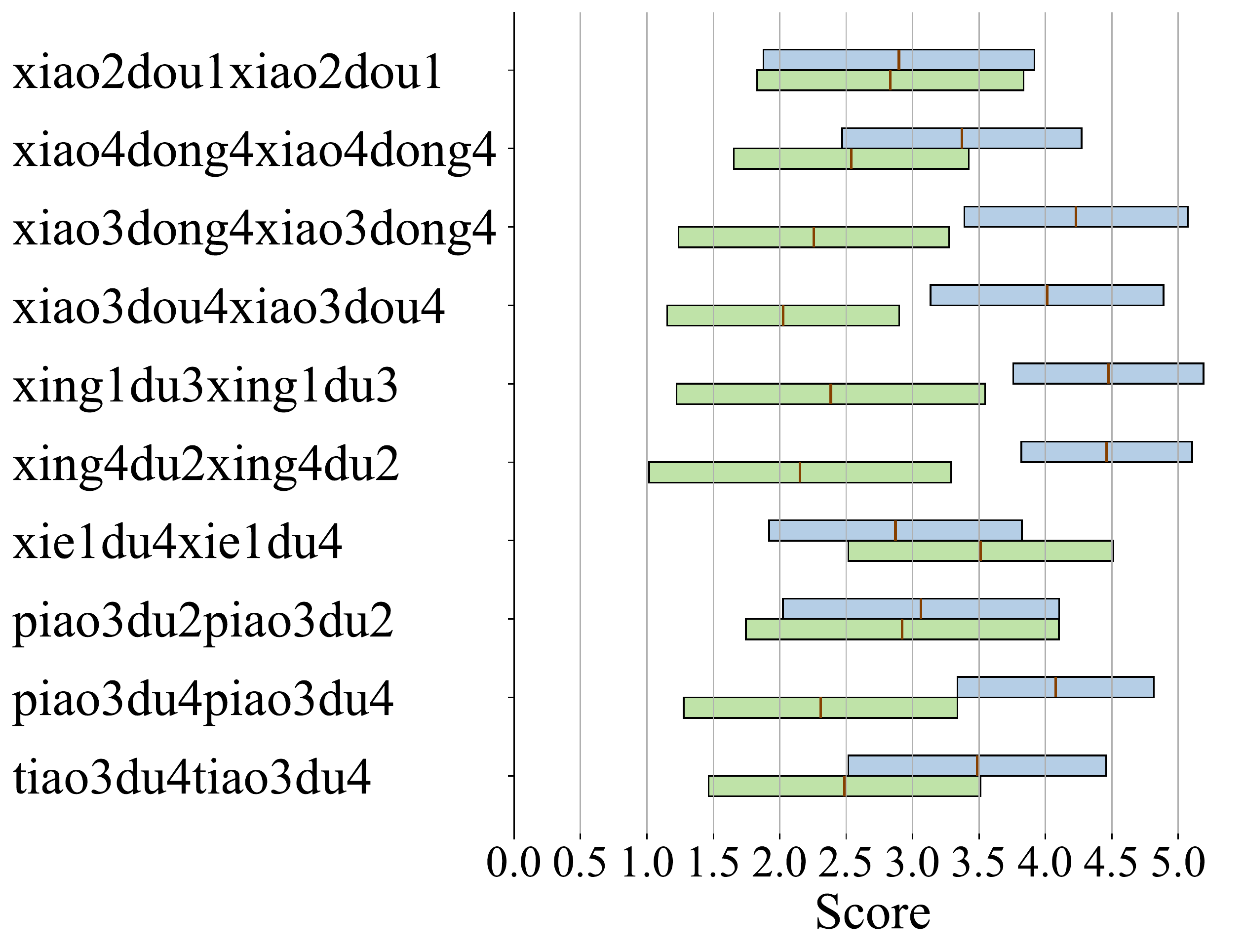}
	}
	\subfigure[Xiaomi]{
	\includegraphics[trim = 3mm 4mm 8mm 7mm, clip,width=0.23\linewidth]{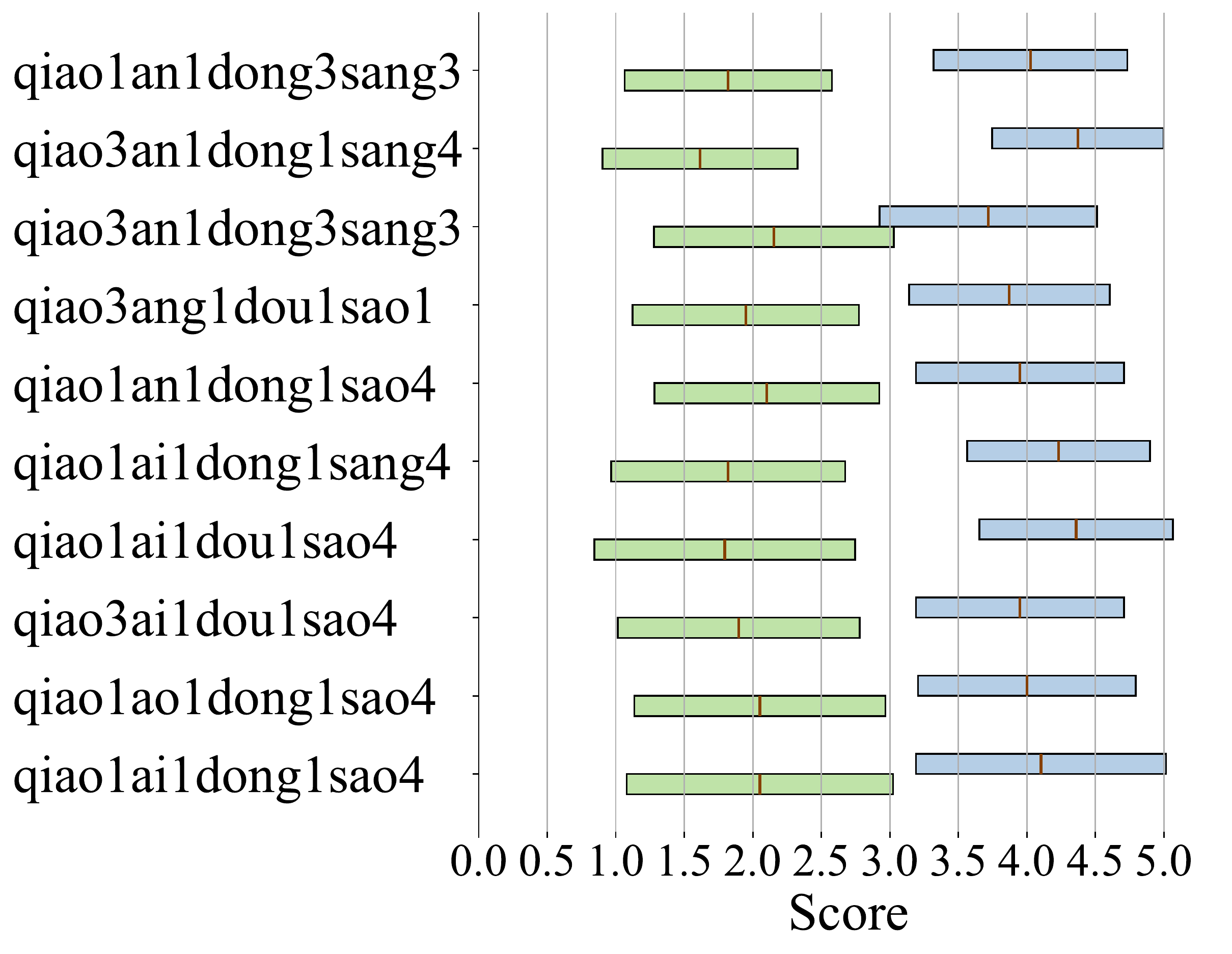}
	}
	\subfigure[AliGenie]{
	\includegraphics[trim = 3mm 4mm 8mm 7mm, clip,width=0.23\linewidth]{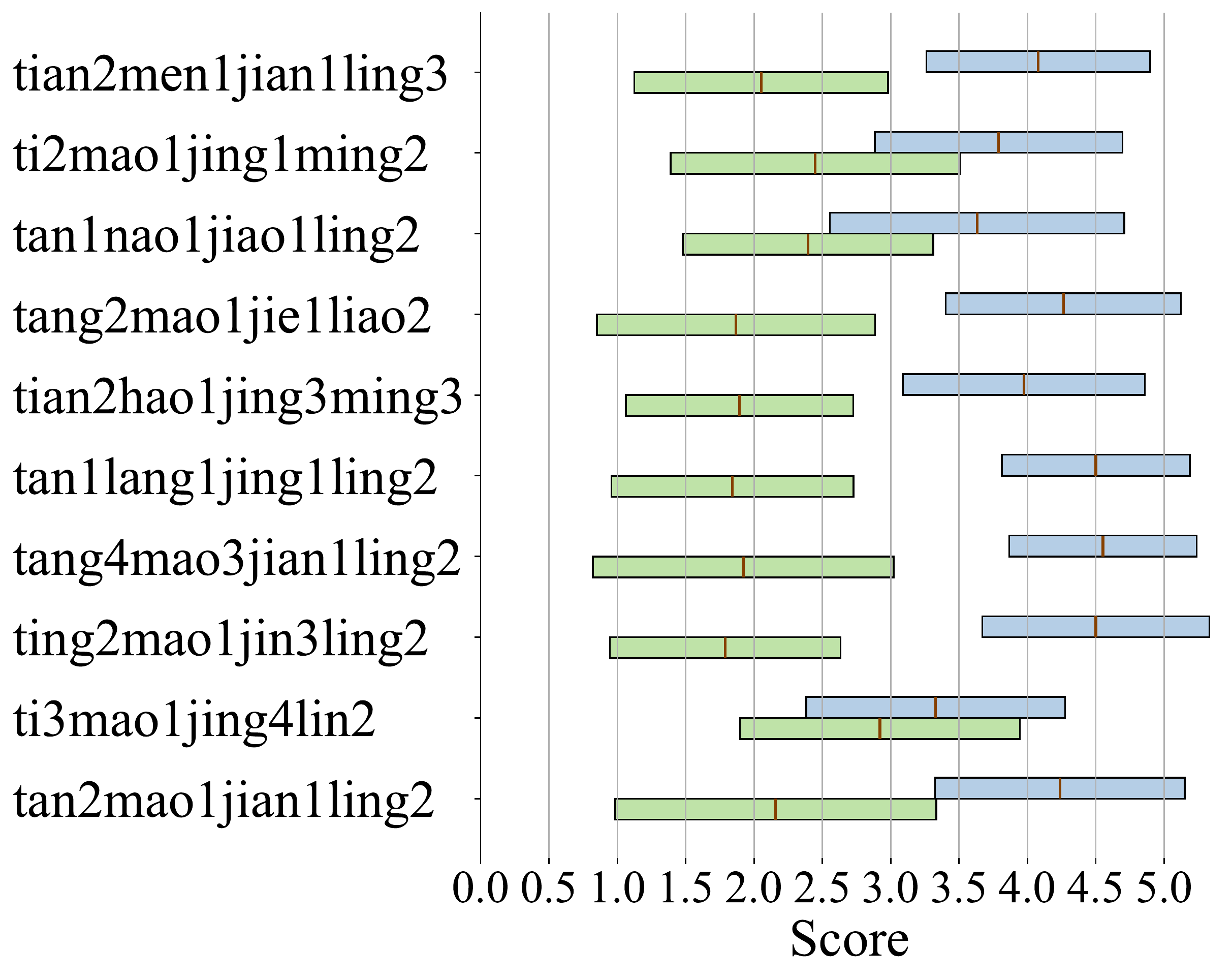}
	}
	\subfigure[Tencent]{
	\includegraphics[trim = 3mm 4mm 8mm 7mm, clip,width=0.23\linewidth]{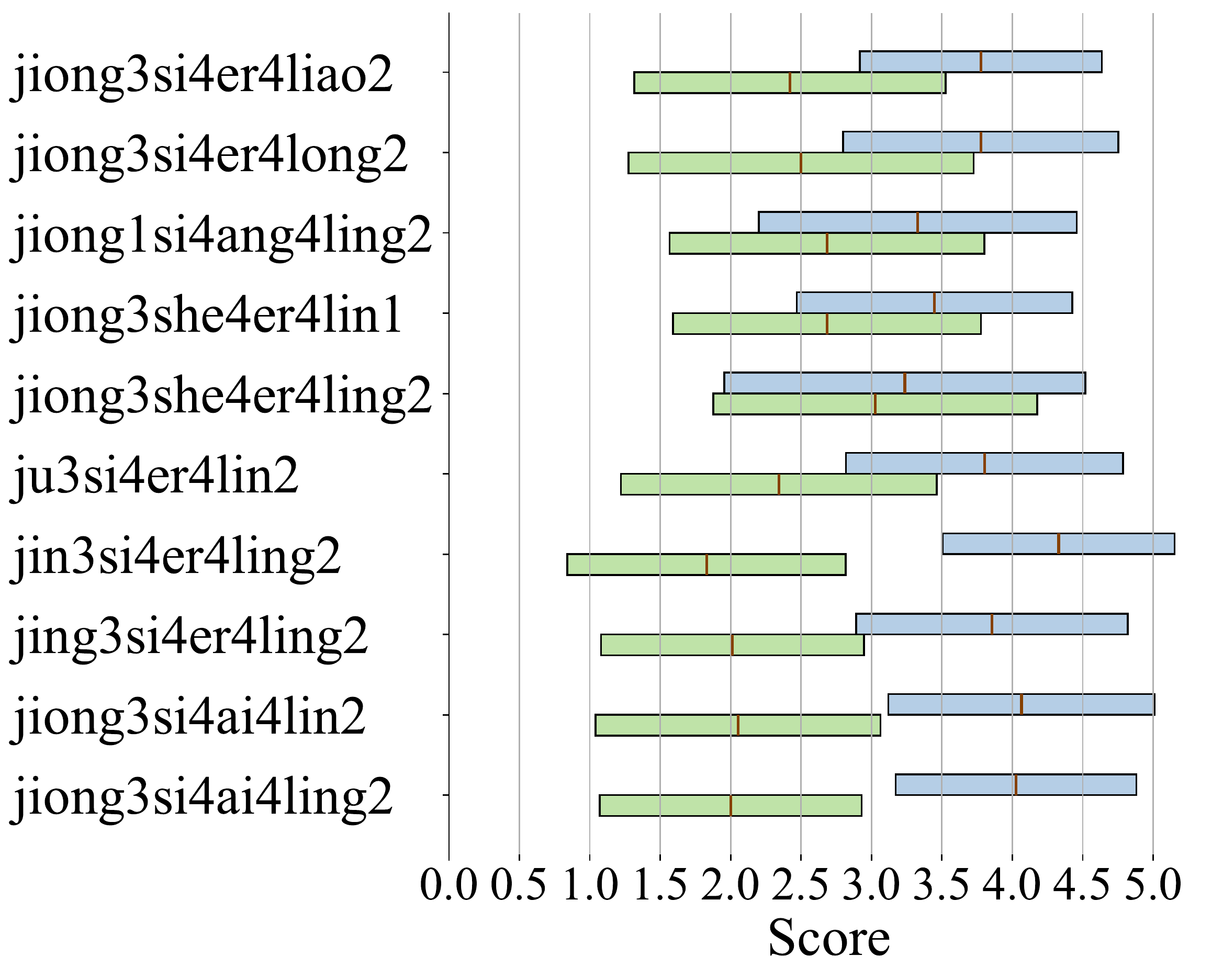}
	}

	\caption{The results of subjective tests for the top-11$\sim$20 \fw. The perceptual dissimilarity between \fw and the real wake-up words ranges from 1 (very similar) to 5 (very dissimilar). The commonality of a fuzzy word in daily life ranges from 1 (not common) to 5 (very common).}
	\label{expdist}
    \end{figure*}
    
    \clearpage

\subsection{Accuracy of Interpretable Model}

	\begin{table}[h]
	\centering
	\caption{Prediction accuracy of the interpretable model in differentiating \fw and non-\fw. }
	\begin{tabular}{cc||cc}
		\hline
		voice assistant &  accuracy & voice assistant &  accuracy \\
		\hline
		Amazon Echo & 85.68\% &Baidu & 94.52\% \\
		Echo Dot & 87.13\% & Xiaomi & 90.79\% \\
		Google & 90.20\%  & AliGenie & 88.73\% \\
		Apple Siri & 93.78\%  &  Tencent & 92.15\%\\
		\hline
		\end{tabular}
		\label{exp}
		
	\end{table}
	
	\subsection{Full List of \FW}
	\begin{table*}[h]
	\centering
	\small	
\caption{Fuzzy words of Baidu. The column \emph{dis.} presents the dissimilarity distance according to the generation algorithm, rounded up to the second decimal point. The column \emph{w. rate} presents the wake-up rate. Note that dissimilarity distances smaller than 0.005 are rounded up to 0.00. }
	\begin{tabular}{lcc||lcc||lcc||lcc}
		\hline
	 & dis. & w. rate & & dis. & w. rate & & dis. & w. rate & & dis. & w. rate\\
		\hline
	xi\v ao l\v ong xi\v ao l\v ong&	0.30&	0.2&
pi\v ao d\`ou pi\v ao d\`ou&	0.22&	0.1&
ti\v ao d\=ou ti\v ao d\=ou&	0.19&	0.6&
ti\v ao d\`ong ti\v ao d\`ong&	0.19&	0.2\\
sh\=ao d\=ou sh\=ao d\=ou&	0.16&	0.2&
 ji\v ao d\=ou ji\v ao d\=ou&	0.15&	0.4&
qi\'ao d\=ou qi\'ao d\=ou&	0.15&	0.3&
qi\v ao d\=ou qi\v ao d\=ou&	0.15&	0.5\\
qi\v ao d\`ong qi\v ao d\`ong&	0.15&	0.2&
 xi\=ao d\=ou xi\=ao d\=ou&	0.14&	0.5&
 xi\'ao d\=ou xi\'ao d\=ou&	0.14&	0.5&
 xi\`ao d\`ong xi\`ao d\`ong&	0.14&	0.2\\
 xi\v ao d\`ong xi\v ao d\`ong&	0.14&	0.7&
 xi\v ao d\`ou xi\v ao d\`ou&	0.14&	0.4&
 xīng d\v u xīng d\v u&	0.13&	0.3&
 xìng d\'u xìng d\'u&	0.13&	0.2\\
 xi\=e d\`u xi\=e d\`u&	0.08&	0.1&
 pi\v ao d\'u pi\v ao d\'u&	0.08&	0.3&
 pi\v ao d\`u pi\v ao d\`u&	0.08&	0.7&
 ti\v ao d\`u ti\v ao d\`u&	0.05&	0.7\\
 xi\=ong d\=u xi\=ong d\=u&	0.04&	0.7&
 xi\=ong d\`u xi\=ong d\`u&	0.04&	0.8&
 xi\`ong d\v u xi\`ong d\v u&	0.04&	0.7&
 qi\v ao b\`u qi\v ao b\`u&	0.03&	0.1\\
 xi\v ao b\`u xi\v ao b\`u&	0.03&	0.3&
 xi\v ao l\v u xi\v ao l\v u&	0.03&	0.3&
 xi\v ao l\`u xi\v ao l\`u&	0.03&	0.4&
 xi\v ao n\`u xi\v ao n\`u&	0.02&	0.2\\
 sh\=ao d\=u sh\=ao d\=u&	0.01&	0.4&
 sh\=ao d\'u sh\=ao d\'u&	0.01&	0.7&
 sh\=ao d\`u sh\=ao d\`u&	0.01&	0.7&
 xi\'a d\`u xi\'a d\`u&	0.01&	0.3\\
 xi\v a d\=u xi\v a d\=u	&0.01&	0.2&
 xi\v a d\`u xi\v a d\`u&	0.01&	0.5&
 xi\`an d\v u xi\`an d\v u&	0.01&	0.1&
  xi\`ang d\v u  xi\`ang d\v u &	0.01&	0.2\\
 xi\'ang d\v u xi\'ang d\v u&	0.01&	0.1&
 xi\'ang d\`u xi\'ang d\`u&	0.01&	0.2&
 xi\v an d\`u xi\v an d\`u&	0.01&	0.3&
 xi\v ang d\=u xi\v ang d\=u&	0.01&	0.3\\
 xi\v ang d\`u xi\v ang d\`u&	0.01&	0.5&
 ji\v ao d\`u ji\v ao d\`u&	0.01&	0.2&
 qi\=ao d\'u qi\=ao d\'u&	0.01&	0.1&
 qi\=ao d\`u qi\=ao d\`u&	0.01&	0.2\\
 qi\'ao d\v u qi\'ao d\v u&	0.01&	0.1&
 qi\'ao d\`u qi\'ao d\`u&	0.01&	0.5&
 qi\v ao d\=u qi\v ao d\=u&	0.01&	0.3&
 qi\v ao d\'u qi\v ao d\'u&	0.01&	0.3\\
 qi\v ao d\`u qi\v ao d\`u&	0.01&	0.4&
 xi\=ao d\=u xi\=ao d\=u&	0.00&	0.3&
 xi\'ao d\=u xi\'ao d\=u&	0.00&	0.8&
 xi\=ao d\'u xi\=ao d\'u&	0.00&	0.8\\
 xi\v ao d\=u xi\v ao d\=u&	0.00&	0.8&
 xi\=ao d\v u xi\=ao d\v u&	0.00&	0.7&
 xi\`ao d\'u xi\`ao d\'u&	0.00&	0.5&
 xi\'ao d\'u xi\'ao d\'u&	0.00&	0.4\\
 xi\=ao d\`u xi\=ao d\`u&	0.00&	0.9&
 xi\'ao d\v u xi\'ao d\v u&	0.00&	0.6&
 xi\v ao d\'u xi\v ao d\'u&	0.00&	0.6&
 xi\`ao d\v u xi\`ao d\v u&	0.00&	0.4\\
 xi\`ao d\`u xi\`ao d\`u&	0.00&	0.9&
 xi\'ao d\`u xi\'ao d\`u&	0.00&	0.7&
 xi\v ao d\v u xi\v ao d\v u&	0.00&	0.3&
 -&-&-\\

		\hline
		\end{tabular}
		\label{full-dot}
	\end{table*}
	\begin{table*}[h]
	\centering
	\small	
\caption{Fuzzy words of Xiaomi. The column \emph{dis.} presents the dissimilarity distance according to the generation algorithm, rounded up to the second decimal point. The column \emph{w. rate} presents the wake-up rate. Note that dissimilarity distances smaller than 0.005 are rounded up to 0.00. }
	\begin{tabular}{lcc||lcc||lcc}
		\hline
	 & dis. & w. rate & & dis. & w. rate & & dis. & w. rate\\
		\hline
		
		qi\v ao b\=ai d\=ong h\`e	&0.33&	0.9&
qi\=ang b\=ai d\=ong s\`e	&0.29&	1&
qi\`ao b\=ai t\=ou sh\`e	&0.27&	0.9\\
qi\=ao c\=ai d\=ong s\`e	&0.27&	1&
xi\=ao c\=ai d\=ong s\`e	&0.27&	1&
qi\=ao \=e d\=u s\`e	&0.19&	1\\
qi\=ao \=ai d\=u s\`e	&0.15&	1&
qi\`ang \=ai d\=ong s\`ang&	0.11&	1&
qi\`ang \=ai ch\=ou l\`e	&0.11&	1\\
qi\v ao \=a d\v ong s\`a	&0.11&	1&
qi\=ao \=an d\v ong s\v ang	&0.11&	0.1&
qi\v ao \=an d\=ong s\`ang	&0.11&	1\\
qi\v ao \=an d\v ong s\v ang	&0.11&	0.6&
qi\v ao \=ang d\=ou s\=ao	&0.10&	0.9&
qi\=ao \=an d\=ong s\`ao	&0.10&	0.9\\
qi\=ao \=ai d\=ong s\`ang	&0.10&	1&
qi\=ao \=ai d\=ou s\`ao	&0.10&	1&
qi\v ao \=ai d\=ou s\`ao	&0.10&	1\\
qi\=ao \=ao d\=ong s\`ao	&0.09&	0.1&
qi\=ao \=ai d\=ong s\`ao	&0.09&	1&
qi\=ang \=ai ch\=ou s\`e	&0.09&	1\\
pi\v ao \=an ch\=ong s\`e	&0.08&	1&
pi\v ao \=ai d\=ou l\`e	&0.08&	1&
qi\v ao \=an g\=ong t\`e	&0.08&	1\\
qi\=ao \=an ch\=ou s\`e	&0.08&	1&
qi\v ao \=an ch\=ou s\`e	&0.08&	1&
qi\=a \=ai du\`an ji\v e	&0.08&	0.1\\
qi\`ao \=ai d\=ong g\`e	&0.08&	1&
pi\v ao \=ai d\=ong r\`e	&0.08&	1&
qi\v ao \=ai d\=ou t\`e	&0.07&	0.8\\
qi\v ao \=ai g\`ong jìn	&0.07&	0.1&
qi\=ao \=an d\=ong sī	&0.06&	1&
pi\=ao \=an d\=ong s\`e	&0.06&	1\\
qi\v ang \=ai g\=ong l\`e	&0.06&	1&
pi\`ao \=ai d\=ou s\`e	&0.06&	0.2&
qi\=ao \=ai d\`ong jìn	&0.06&	0.1\\
qi\'ao \=ai d\`ong jìn	&0.06&	0.1&
qi\`ao \=ai d\=ong jìn	&0.06&	0.9&
qi\`ao \=ai t\=ou qìng	&0.06&	0.8\\
qi\=ao \=en d\=ong s\`e	&0.06&	1&
qi\=ao \=ai g\=ou r\`e	&0.06&	1&
pi\v ao \=ai g\=ong s\`e	&0.06&	1\\
qi\=ao \`ai d\=ong qíng	&0.06&	1&
qi\`ao \=an d\=ou l\`e	&0.06&	1&
qi\v ao \=an d\=ou l\`e	&0.06&	0.8\\
qi\v ao \=a d\=ou l\`e	&0.06&	1&
qi\v ao \=an g\=ong n\`e	&0.06&	0.1&
pi\`ao \=ai d\=ong sh\`e	&0.06&	0.8\\
pi\=ao \=ai d\=ong s\`e	&0.05&	1&
pi\`ao \=ai d\=ong s\`e	&0.05&	0.8&
pi\v ao \=ai d\=ong s\`e	&0.05&	1\\
qi\v ao \=an d\=ong n\`e	&0.05&	0.7&
qi\=ao \=an d\=ong r\`e	&0.05&	0.7&
qi\v ang \=an d\=ou s\`e	&0.05&	1\\
qi\=ao \=an d\=ong r\v e	&0.05&	0.6&
qi\`ao \=an ch\=ong s\`e	&0.05&	0.4&
qi\`ao \=ai g\=ong r\`e	&0.05&	0.9\\
qi\v ang \=an d\=ong s\`e	&0.04&	0.6&
qi\=ao \=ai d\=ong r\`e	&0.04&	1&
ti\`ao \`ai d\=ong sh\`e	&0.04&	0.1\\
qi\=ao \=ai ch\=ong s\`e	&0.04&	1&
qi\`ao \=ai ch\=ong s\`e	&0.04&	1&
xi\`ao \=ai d\`ong r\`e	&0.04&	0.3\\
xi\=ao \=ai c\=ong s\`e	&0.04&	1&
qi\`ao \=an d\=ou sh\`e	&0.04&	1&
qi\v ao \=an d\=ou sh\`e	&0.04&	0.9\\
qi\=ao \=an d\=ou s\`e	&0.04&	0.1&
qi\v ao \=an d\=ou s\`e	&0.04&	0.9&
qi\v ao \=a d\=ou s\`e	&0.04&	1\\
qi\=ang \=ai d\=ong s\`e	&0.04&	0.9&
xi\=ao \=ai c\=ong ji\`e	&0.03&	1&
qi\=ao \=ai g\`ou ji\`e	&0.03&	0.1\\
qi\=ao \=an d\=ong s\`e	&0.03&	1&
qi\=ao \=a d\=ong s\`e	&0.03&	0.9&
qi\`ao \=an d\=ong s\`e	&0.03&	0.5\\
qi\v ao \=an d\=ong s\`e	&0.03&	1&
qi\`ao \=ai d\=ou sh\`e	&0.03&	1&
qi\`ao \`ai d\=ou sh\`e	&0.03&	0.8\\
qi\=ao \=ai d\=ou s\`e	&0.03&	1&
qi\`ao \=ai d\=ou s\`e	&0.03&	0.9&
qi\v ao \=ai d\=ou s\`e	&0.03&	1\\
xi\`ao \=an d\=ong s\`e	&0.03&	1&
xi\=ao \=a d\=ong s\`e	&0.03&	1&
qi\=ao \=ai g\=ong s\`e	&0.02&	1\\
qi\v ao \=ao d\=ong sh\`e	&0.02&	1&
qi\=ao \=ai d\=ou ji\`e	&0.02&	0.2&
qi\=ao \=ai t\=ou s\`e	&0.02&	1\\
qi\=ao \=ai d\=ong sh\`e	&0.02&	1&
qi\`ao \`ai d\=ong sh\`e	&0.02&	0.3&
qi\=ao \=ai g\=ong ji\`e	&0.02&	0.9\\
qi\=ao \=ai d\`ong s\`e	&0.02&	0.5&
qi\=ao \=ai g\`ong ji\v e	&0.02&	0.4&
qi\`ao \=ai d\`ong s\`e	&0.02&	1\\
qi\=ao \=ai d\=ong s\`e	&0.02&	1&
qi\`ao \=ai d\=ong s\`e	&0.02&	1&
qi\v ao \=ai d\=ong s\`e	&0.02&	1\\
qi\=ao \=ai d\=ong sh\`eng	&0.02&	1&
xi\=ao \=ai g\`ong ji\`e	&0.02&	0.1&
xi\v ao \=ai g\`ong ji\`e	&0.02&	0.1\\
xi\=ao \=ai d\=ong s\`e	&0.02&	1&
xi\'ao \=ai d\=ong s\`e	&0.02&	1&
qi\=ao \=ai d\`ong ji\`e	&0.02&	0.1\\
qi\'ao \=ai d\`ong ji\`e	&0.02&	0.5&
qi\=ao \=ai d\=ong ji\`e	&0.02&	1&
qi\`ao \=ai d\=ong ji\`e	&0.02&	0.1\\
xi\v ao \=ai d\`ong ji\`e	&0.01&	0.2&
xi\=ao \=ai d\=ong ji\`e	&0.01&	1&
xi\v ao \=ai d\=ong ji\`e	&0.01&	1\\

				\hline
		\end{tabular}
		\label{full-dot}
	\end{table*}
	\begin{table*}[h]
	\centering
	\small	
\caption{Fuzzy words of AliGenie: Part I. The column \emph{dis.} presents the dissimilarity distance according to the generation algorithm, rounded up to the second decimal point. The column \emph{w. rate} presents the wake-up rate. Note that dissimilarity distances smaller than 0.005 are rounded up to 0.00. }
	\begin{tabular}{lcc||lcc||lcc}
		\hline
	 & dis. & w. rate & & dis. & w. rate & & dis. & w. rate\\
		\hline

 y\=an   m\=en    jīng  l\'ing	&0.29&	0.5&
 y\=an   m\=ang     jīng  l\'ing	&0.26&	1&
 w\'an  m\=ang     jīng  l\'ing	&0.26&	0.4\\
 w\'ang  m\` ao   jǐng  lǐn 	&0.25&	0.1&
 y\'an  m\=ao    jīng   lǐng 	&0.25&	0.9&
 y\=an  m\=ao    jīng  l\'ing	&0.25&	1\\
 w\=an  m\=ao    jīng  l\'ing	&0.25&	0.9&
 y\'an  m\=ao    jīng  l\'ing	&0.25&	0.9&
 w\'an  m\=ao    jīng  l\'ing	&0.25&	0.2\\
 w\'ang  m\=ao    jīng  l\'ing	&0.25&	0.1&
ti\'an  m\=en   ji\=an  lǐng	&0.11&	0.7&
 t\'i     m\=ao    jīng  m\'ing 	&0.10&	1\\
 t\= an   n\=ao     ji\=ao    l\'ing	&0.10&	0.4&
 t\'ang     m\=ao   ji\=e    li\'ao   	&0.09&	0.1&
ti\'an  h\=ao   jǐng  mǐng 	&0.08&	0.7\\
 t\= an   l\=ang    jīng  l\'ing	&0.07&	0.3&
 t\` ang     m\v ao   ji\=an  l\'ing	&0.07&	0.2&
t\'ing  m\=ao  jǐn  l\'ing	&0.07&	1\\
 t\v i  m\=ao   jìng  l\'in 	&0.07&	1&
 t\' an   m\=ao   ji\=an  l\'ing	&0.07&	0.4&
 t\= an   m\=ao    jīng   li\'an 	&0.07&	0.2\\
 t\= an   m\=ao    jīng   li\'ang	&0.07&	0.4&
ti\'an  m\=ao   ji\v ang   lìng&	0.07&	0.4&
ti\'an  m\=ao    jīng   li\v an  	&0.07&	1\\
 t\'i     m\'ao    jīng  l\'ing	&0.07&	0.8&
 t\'i     m\=ao    jīng  l\'ing	&0.07&	1&
tīng  m\=ao   jǐng   lǐng 	&0.07&	1\\
 ti\v an   m\` ao   jìng   li\'ang &	0.07&	1&
ti1 m\=ao    jīng  l\'ing	&0.07&	1&
 sh\=ai  m\=ao    jīng  l\'ing	&0.07&	0.6\\
 t\= an   m\=ao    jīng   li\'ao   	&0.07&	0.1&
 ti\=an   m\v ao   jìng   li\`ao 	&0.06&	0.9&
ti\'an  m\=ao   ji\=ao     lǐng 	&0.06&	0.2\\
ti\'an  m\=ao    jīng   li\'ao   	&0.06&	1&
 t\=ang     p\= ao      jīng  lìn  	&0.06&	0.1&
ti\'an  p\v ao    jìng  n\'ing   	&0.06&	1\\
 t\' an   l\v ao   ji\v e  l\'ing	&0.06&	0.1&
 t\= an   p\= ao      jīng  l\'ing	&0.06&	0.1&
 sh\`ang    m\=ao   jǐng   lǐng &	0.05&	0.1\\
 ti\=e  m\=a     jīng  l\'ing	&0.05&	1&
 sh\=an    m\=ao   jǐng   lǐng 	&0.05&	0.2&
 sh\=an    m\=ao    jīng  l\'ing	&0.05&	1\\
 s\= an   m\=ao   jǐng  l\'ing	&0.05&	1&
 s\= an   m\=ao    jīng   lǐng 	&0.05&	0.5&
 s\= an   m\=ao    jīng   l\'ing	&0.05&	1\\
 t\' an   h\=ao    jīng  l\'ing	&0.05&	0.1&
 t\= an   h\=ao    jīng  l\'ing	&0.05&	0.1&
 t\=ang     m\'ao    jīng   ni\`e   	&0.05& 0.1\\
 ti\=an   k\=ao    jīng  l\'ing	&0.05&	1&
teng1 m\=ao   jǐng   lǐng 	&0.05&	1&
 t\= an   m\=ao   ji\=ong  l\'ing	&0.05&	0.7\\
ti\'an  m\=ao   ji\v ong     līng &	0.05&	0.8&
ti\'an  m\=ao   ji\v ong     l\'ing&	0.05&	1&
ti\'an  m\` ao   ji\=ong   lǐng &	0.05&	1\\
ti\'an  m\=ao    jīng   l\' ou   	&0.05&	1&
 z\=an     m\=ao    jīng  l\'ing	&0.04&	1&
 ti\v an   m\=en  jǐng   lǐng 	&0.04&	0.5\\
ti\'an  m\=e      jìng  l\'ing	&0.04&	1&
 ti\=an   m\=e       jīng  l\'ing	&0.04&	1&
 di\= an   m\=ao q\'i mìng 	&0.04&	0.9\\
ti\'an  l\=ao līng  l\'ing	&0.04&	0.3&
 ti\v an  mie1 jǐng  līng 	&0.04&	0.8&
 t\= an   n\=ao     jīng  n\'ing   	&0.04&	0.5\\
 t\= an   l\=ao   jīng  n\'ing   	&0.04&	0.7&
 t\' an   m\=ao jīn  m\'ing 	&0.03&	0.2&
 t\' a      l\=ao   jīng  l\'ing	&0.03&	0.3\\
 ti\v an   n\`ao    jìng  līn 	&0.03&	0.9&
 t\' an   l\=ao   jīng   lǐng 	&0.03&	0.1&
 t\' an   m\=ao   jīng  m\'ing 	&0.03&	0.4\\
 t\= an   n\=ao    jǐng   lǐng 	&0.03&	0.2&
 t\= an   n\=ao    jìng  l\'ing	&0.03&	0.1&
 t\= an   n\=ao     jīng  l\'ing	&0.03&	0.2\\
 t\= an   l\=ao  jǐng   lǐng 	&0.03&	0.3&
 t\= an   l\=ao   jīng  l\'ing	&0.03&	0.8&
 t\= an   m\=ao   jīng  m\'ing 	&0.03&	0.3\\
 ti\=an   l\=ao  jǐng  līn 	&0.03&	1&
 ti\v an   m\v ao  jìng  mìng &	0.03&	1&
ti\'an  n\=ao    jǐng   lǐng &	0.03&	1\\
ti\'an  n\=ao     jīng   lǐng 	&0.03&	1&
ti\'an  n\=ao     jīng  l\'ing	&0.03&	1&
 ti\=an   n\=ao     jīng  l\'ing	&0.03&	1\\
 ti\v an   l\=ao  jǐng   lìng 	&0.03&	1&
 ti\v an   m\=ao  jǐng  m\'ing 	&0.03&	1&
ti\'an  m\=ao  jǐng  m\'ing 	&0.03&	1\\
ti\'an  l\=ao   jīng   lǐng 	&0.03&	1&
ti\'an  l\=ao   jīng  l\'ing	&0.03&	1&
 ti\=an   m\=ao   jīng  m\'ing 	&0.03&	1\\
 ti\=an   l\=ao   jīng  l\'ing	&0.03&	1&
  r\' an  m\=ao   jīng   lǐng 	&0.03&	0.6&
  r\' an  m\=ao   jīng  l\'ing	&0.03&	0.7\\
 xi\v an  mi\'ao  jǐng   lǐng &	0.03&	0.1&
 t\' an   m\=ao  ji\=e   l\'in 	&0.03&	0.4&
 t\= an   m\=ao  ji\v e   lǐng 	&0.03&	0.4\\
 t\'ang     m\` ao  ji\=e    lǐng 	&0.03&	0.2&
 t\' an   m\=ao  ji\=e   l\'ing	&0.03&	0.5&
 ti\v an   m\` ao jǐn   li\` e 	&0.03&	1\\
 t\= an   m\=ao  ji\=e   l\'ing	&0.02&	0.7&
ti\'an  m\` ao  ji\=e   lǐn 	&0.02&	1&
 ti\v an   m\=ao mǐng  l\'ing	&0.02&	0.2\\
ti\'an  m\` ao  ji\v e   lǐng 	&0.02&	1&
ti\'an  m\v ao  jǐng   li\v e 	&0.02&	1&
ti\'an  m\=ao  jǐng   li\v e 	&0.02&	1\\
 ti\v an   m\'ao  jìng   l\` eng   	&0.02&	1&
ti\'an  m\=ao  jìng   l\` eng   	&0.02&	0.5&
 ti\` an   m\=ao dǐng   lǐng 	&0.02&	1\\
ti\'an  m\' ou qìng  l\'ing	&0.02&	1&
 ti\v an   m\' ou  jìng  n\'ing   	&0.02&	1&
 ti\` an   m\' ou jìn   l\'ing	&0.02&	1\\
ti\'an  m\' ou  jìng  l\'ing	&0.02&	1&
pian2 m\=ao  jǐng  l\'ing	&0.01&	1&
 di\= an   mi\` ao n ìng  l\'ing&	0.01&	0.2\\
 t\`ai    m\` ao  jìng  l\'in 	&0.01&	0.3&
ti\'an  m\'ao līng  n\'ing 	&0.01&	0.3&
 t\=ai    m\=ao   jīng  l\'in 	&0.01&	0.4\\
 di\v an   m\v ao  q\'i  nǐng 	&0.01&	0.4&
 di\v an   m\v ao  q\'i  n\'ing 	&0.01&	1&
 t\=ai    m\=ao  jǐng   lǐng 	&0.01&	0.5\\
 t\=ai    m\=ao  jǐng  l\'ing	&0.01&	0.8&
 t\=ai    m\=ao   jīng  l\'ing	&0.01&	0.1&
ti\'an  m\=ao līng   lǐng 	&0.01&	0.2\\
ti\=an   m\=ao  q\'i  nìng 	&0.01&	0.9&
 di\= an   m\'ang  jǐn  l\'ing	&0.01&	1&
 ti\=an   mi\'ao q\'in  lǐn 	&0.01&	0.2\\
 ti\=an   mi\v ao q\'in  l\'in 	&0.01&	0.1&
ti\'an  m\=ang   jǐng  nǐng &	0.01&	1&
 t\= an   m\=ang   jǐng   lǐng 	&0.01&	0.2\\
 di\= an   m\=ang   jǐng  l\'ing&	0.01&	1&
 t\= an   m\=ang   jǐng  l\'ing	&0.01&	0.4&
 t\= an   m\=ang    jīng  l\'ing	&0.01&	0.6\\
tiao2 m\=ao   jīng  l\'ing	&0.01&	0.2&
 t\= an   m\= an    jīng  l\'ing	&0.01&	0.4&
 t\=ao   m\=ao  jǐng   lǐng 	&0.01&	0.1\\
 t\=ao   m\=ao   jīng  l\'ing	&0.01&	0.1&
 di\v an   m\` ao q\'in  līn 	&0.01&	0.7&
 t\= an   m\=a   jǐng   lǐng 	&0.01&	0.2\\
 di\v an   m\` ao q\'in  l\'in 	&0.01&	0.7&
ti\'an  m\=ang   jǐng   lǐng &	0.01&	1&
ti\'an  m\=ang    jīng  l\'ing&	0.01&	1\\
 ti\=an   m\=ang    jīng  līng &	0.01&	1&
 ti\=an   m\=ang    jīng  l\'ing&	0.01&	1&
 ti\v an   m\= an   jìng   lìng 	&0.00&	1\\
ti\'an  m\= an   jìng   lìng 	&0.00&	0.9&
 ti\v an   m\= an    jīng   lìng 	&0.00&	1&
ti\'an  m\= an    jīng  l\'ing	&0.00&	1\\
 ti\v an   m\=ao qǐn  lǐn 	&0.00&	1&
ti\'an  m\=a    jīng  l\'ing	&0.00&	1&
 ti\v an   m\=ao q\'in  līn 	&0.00&	1\\
 di\v an   mi\'ao jǐn  l\'ing	&0.00&	0.1&
 di\= an   mi\` ao jǐn  l\'ing	&0.00&	0.3&
 ti\v an   m\'ao qīn  l\'in 	&0.00&	1\\
ti\'an  m\` ao qǐn   lǐng 	&0.00&	1&
 di\= an   mi\` ao  jìng  l\'in 	&0.00&	0.3&
 di\= an   mi\` ao  jǐng  l\'in 	&0.00&	0.3\\
 ti\v an   mi\v ao jīn  n\'ing   	&0.00&	0.6&
 di\= an   mi\=ao jǐn   lǐng 	&0.00&	0.4&
ti\'an  m\=ao  jìng  n\'in 	&0.00&	1\\
 t\v ang     mi\'ao  jìng   lǐng &	0.00&	0.1&
 t\v ang     mi\=ao  jǐng  l\'ing&	0.00&	0.1&
 di\v an   m\=ao q\'ing  l\'in 	&0.00&	1\\
 di\= an   mi\` ao  jǐng  l\'ing&	0.00&	0.5&
 ti\=an   mi\=ao jīn  lìn  	&0.00&	0.9&
 t\v an   m\=ao jìn   lǐn 	&0.00&	0.1\\
 t\' an   m\=ao jīn  n\'ing   	&0.00&	0.2&
 di\= an   mi\=ao  jìng  l\'ing&	0.00&	0.8&
 ti\=an   m\=ao qǐng  nìng 	&0.00&	1\\
		
								\hline
		\end{tabular}
		\label{full-dot}
	\end{table*}

	\begin{table*}[h]
	\centering
	\small	
\caption{Fuzzy words of AliGenie: Part II. The column \emph{dis.} presents the dissimilarity distance according to the generation algorithm, rounded up to the second decimal point. The column \emph{w. rate} presents the wake-up rate. Note that dissimilarity distances smaller than 0.005 are rounded up to 0.00. }
	\begin{tabular}{lcc||lcc||lcc}
		\hline
	 & dis. & w. rate & & dis. & w. rate & & dis. & w. rate\\
		\hline

 di\v an   m\` ao q\'ing l\'ing	&0.00&	0.7&
 ti\=an   mi\v ao  jǐng lǐn 	&0.00&	0.1&
 di\v an   m\=ao q\'ing l\'ing	&0.00&	1\\
 ti\=an   mi\'ao  jìng līn 	&0.00&	0.5&
 t\=ang     m\=ao jīn lǐn 	&0.00&	0.6&
 di\v an   m\=ao  jìng n\'ing   	&0.00&	1\\
 di\v an   m\=ao  jǐng n\'ing   	&0.00&	1&
 di\v an   m\'ao jīn l\'in 	&0.00&	1&
 t\` ang     m\=ao jǐn l\'ing	&0.00&	0.6\\
 t\=ang     m\=ao   jīng n\'ing   	&0.00&	0.7&
 t\= an   m\=ao  jǐng nǐng 	&0.00&	0.5&
 t\' an   m\=ao   jīng n\'ing   	&0.00&	0.2\\
 di\= an   m\=ao jīn līn 	&0.00&	1&
 t\= an   m\=ao   jīng n\'ing   	&0.00&	0.6&
 ti\=an   mi\=ao   jīng l\'in 	&0.00&	1\\
 t\'ang     m\=ao jǐn l\'ing	&0.00&	0.3&
ti\'an  m\=ao qìng līn 	&0.00&	1&
 t\=ang     m\v ao  jǐng lìn  	&0.00&	0.2\\
 ti\v an   mi\` ao  jìng l\'ing	&0.00&	0.1&
ti\'an  m\=ao jǐn nǐng 	&0.00&	1&
 t\' an   m\=ao jǐn l\'ing	&0.00&	0.5\\
 t\'ang     m\=ao jīn l\'ing	&0.00&	0.3&
 t\=ang     m\v ao   jīng lǐn 	&0.00&	0.6&
 t\=ang     m\=ao  jǐng lǐn 	&0.00&	0.6\\
ti\'an  m\` ao q\'ing līn 	&0.00&	1&
 t\' an   m\=ao   jīng lǐn 	&0.00&	0.1&
 ti\=an   m\=ao qǐng l\'in 	&0.00&	1\\
 di\= an   m\'ao jǐn  lǐng 	&0.00&	1&
ti\'an  m\=ao jǐn n\'ing   	&0.00&	1&
 t\= an   m\=ao jǐn  lǐng 	&0.00&	0.6\\
 ti\v an   mi\'ao  jìng  lǐng &	0.00&	0.1&
 t\' an   m\=ao jīn līng 	&0.00&	0.3&
 t\' an   m\=ao   jīng l\'in 	&0.00&	0.2\\
 di\= an   m\'ao  jǐng lǐn 	&0.00&	1&
 t\= an   m\=ao  jǐng lǐn 	&0.00&	0.3&
 t\' a      m\=ao   jīng līng 	&0.00&	0.4\\
 ti\v an   mi\=ao  jǐng līng &	0.00&	0.9&
ti\'an  mi\=ao  jǐng  lìng &	0.00&	0.4&
 ti\v an   mi\=ao  jǐng  lǐng &	0.00&	0.2\\
 ti\=an   mi\=ao  jìng  lǐng &	0.00&	1&
ti\'an  mi\=ao  jìng l\'ing&	0.00&	0.2&
ti\'an  mi\=ao  jǐng l\'ing&	0.00&	0.4\\
 ti\v an   mi\=ao   jīng  lǐng &	0.00&	0.1&
ti\'an  mi\=ao   jīng l\'ing&	0.00&	0.6&
 t\= an   m\=ao   jīng l\'in 	&0.00&	0.7\\
 t\= an   m\=ao jīn l\'ing	&0.00&	0.4&
 ti\=an   mi\=ao   jīng l\'ing&	0.00&	1&
 ti\v an   m\` ao jǐn lìn  	&0.00&	1\\
 ti\v an   m\v ao jǐn lìn  	&0.00&	1&
 t\=a   m\=ao  jǐng l\'ing	&0.00&	0.2&
 t\=a   m\=ao   jīng l\'ing	&0.00&	0.3\\
 t\v ang     m\=ao  jìng  lǐng 	&0.00&	0.1&
 t\v ang     m\=ao  jìng l\'ing	&0.00&	0.1&
 t\v ang     m\=ao   jīng l\'ing	&0.00&	0.3\\
 ti\v an   m\` ao jǐn lǐn 	&0.00&	1&
ti\'an  m\` ao qìng līng 	&0.00&	1&
 ti\=an   m\v ao qìng l\'ing	&0.00&	1\\
 t\'ang     m\v ao  jìng l\'ing	&0.00&	0.2&
 t\'ang     m\=ao  jìng  lǐng 	&0.00&	0.1&
 t\'ang     m\'ao  jǐng  lǐng 	&0.00&	0.1\\
 t\'ang     m\v ao   jīng  lǐng 	&0.00&	0.1&
 t\'ang     m\=ao   jīng  lǐng 	&0.00&	0.2&
 ti\v an   m\v ao  jìng nìng 	&0.00&	1\\
ti\'an  m\v ao jìng nìng 	&0.00&	1&
 t\v an   m\` ao   jīng  lìng 	&0.00&	0.1&
 di\v an   m\'ao  jǐng  lǐng 	&0.00&	1\\
 t\v an   m\=ao  jìng l\'ing	&0.00&	0.2&
ti\'an  m\=ao qǐng l\'ing	&0.00&	1&
 t\v an   m\=ao   jīng l\'ing	&0.00&	0.3\\
 t\=ang     m\v ao  jìng līng 	&0.00&	0.6&
 t\=ang     m\=ao  jìng  lǐng 	&0.00&	0.6&
 t\=ang     m\=ao  jìng l\'ing	&0.00&	0.7\\
 t\=ang     m\=ao jǐng l\'ing	&0.00&	0.2&
 t\=ang     m\=ao   jīng l\'ing	&0.00&	0.5&
 t\' an   m\` ao  jǐng l\'ing	&0.00&	0.1\\
 t\' an   m\=ao  jìng  lǐng 	&0.00&	0.2&
 t\' an   m\v ao  jǐng l\'ing	&0.00&	0.1&
 t\' an   m\=ao  jǐng  lǐng 	&0.00&	0.3\\
 t\' an   m\=ao  jǐng l\'ing	&0.00&	0.3&
 t\' an   m\=ao   jīng  lǐng 	&0.00&	0.2&
 ti\=an   m\` ao jīn lǐn 	&0.00&	1\\
ti\'an  m\'ao jīn līn 	&0.00&	1&
 di\= an   m\'ao  jǐng  lǐng 	&0.00&	1&
ti\'an  m\'ao  jǐng n\'ing   	&0.00&	1\\
 ti\v an   m\'ao jīn l\'in 	&0.00&	1&
ti\'an  m\=ao  jǐng n\'ing   	&0.00&	1&
 t\= an   m\=ao  jìng l\'ing	&0.00&	0.7\\
 t\= an   m\=ao  jǐng  lǐng 	&0.00&	0.1&
 di\= an   m\=ao  jǐng l\'ing	&0.00&	1&
 t\= an   m\=ao  jǐng l\'ing	&0.00&	0.4\\
ti\'an  m\=ao   jīng n\'ing   	&0.00&	1&
 t\= an   m\=ao   jīng  lǐng 	&0.00&	0.6&
 t\= an   m\=ao   jīng līng 	&0.00&	0.4\\
 ti\=an   m\=ao   jīng n\'ing   &	0.00&	1&
 t\= an   m\=ao   jīng l\'ing	&0.00&	0.3&
 ti\v an   m\v ao  jìng lìn  	&0.00&	1\\
ti\'an  m\` ao  jǐng lìn  	&0.00&	1&
 ti\=an   m\'ao  jǐng lìn  	&0.00&	1&
 ti\v an   m\=ao jǐn  lǐng 	&0.00&	1\\
ti\'an  m\=ao jǐn  lǐng 	&0.00&	1&
 ti\v an   m\=ao   jīng lìn  	&0.00&	1&
ti\'an  m\=ao jǐn l\'ing	&0.00&	1\\
ti\'an  m\` ao  jìng līn 	&0.00&	1&
 ti\v an   m\v ao  jǐng lǐn 	&0.00&	1&
ti\'an  m\v ao  jìng lǐn 	&0.00&	1\\
 ti\v an   m\=ao  jìng lǐn 	&0.00&	1&
ti\'an  m\'ao  jǐng līn 	&0.00&	1&
 ti\=an   m\v ao  jǐng lǐn 	&0.00&	1\\
 ti\v an   m\` ao  jǐng l\'in 	&0.00&	1&
 ti\=an   m\` ao  jǐng l\'in 	&0.00&	1&
ti\'an  m\'ao  jǐng l\'in 	&0.00&	1\\
 ti\v an   m\'ao jīn līng 	&0.00&	1&
ti\'an  m\=ao jīn l\'ing	&0.00&	1&
 ti\=an   m\=ao jīn l\'ing	&0.00&	1\\
 ti\=an   m\=ao   jīng l\'in 	&0.00&	1&
 ti\v an   m\v ao  jìng  lìng 	&0.00&	1&
 ti\v an   m\v ao  jìng  lǐng 	&0.00&	1\\
 ti\v an   m\v ao  jǐng  lìng 	&0.00&	1&
ti\'an  m\` ao  jìng  lǐng 	&0.00&	1&
 ti\v an   m\v ao  jìng līng 	&0.00&	1\\
ti\'an  m\` ao  jǐng  lǐng 	&0.00&	1&
ti\'an  m\v ao  jìng līng 	&0.00&	1&
ti\'an  m\v ao  jìng  lǐng 	&0.00&	1\\
 ti\` an   m\=ao  jǐng  lìng 	&0.00&	1&
ti\'an  m\=ao  jìng  lìng 	&0.00&	1&
 ti\v an   m\=ao  jǐng  lìng 	&0.00&	1\\
 ti\v an   m\=ao  jìng  lǐng 	&0.00&	1&
ti\'an  m\'ao  jìng līng 	&0.00&	1&
 ti\v an   m\=ao  jǐng  lǐng 	&0.00&	1\\
ti\'an  m\'ao  jǐng  lǐng 	&0.00&	1&
ti\'an  m\` ao   jīng  lǐng 	&0.00&	1&
 ti\v an   m\=ao  jǐng līng 	&0.00&	1\\
 ti\=an   m\v ao  jǐng  lǐng 	&0.00&	1&
ti\'an  m\=ao  jǐng  lìng 	&0.00&	1&
 ti\=an   m\` ao  jǐng l\'ing	&0.00&	1\\
ti\'an  m\=ao  jìng  lǐng 	&0.00&	1&
ti\'an  m\v ao  jǐng l\'ing	&0.00&	1&
 ti\v an   m\=ao   jīng  lìng 	&0.00&	1\\
ti\'an  m\=ao  jǐng līng 	&0.00&	1&
 ti\v an   m\=ao  jǐng l\'ing	&0.00&	1&
ti\'an  m\=ao  jǐng  lǐng 	&0.00&	1\\
ti\'an  m\v ao   jīng  lǐng 	&0.00&	1&
ti\'an  m\=ao  jìng l\'ing	&0.00&	1&
ti\'an  m\=ao  jǐng l\'ing	&0.00&	1\\
 ti\=an   m\=ao  jǐng  lǐng 	&0.00&	1&
ti\'an  m\=ao   jīng  lìng 	&0.00&	1&
 ti\v an   m\=ao   jīng līng 	&0.00&	1\\
 ti\v an   m\=ao   jīng  lǐng 	&0.00&	1&
ti\'an  m\=ao   jīng  lǐng 	&0.00&	1&
ti\'an  m\=ao   jīng līng 	&0.00&	1\\
 ti\v an   m\=ao   jīng l\'ing	&0.00&	1&
 ti\=an   m\=ao  jǐng l\'ing	&0.00&	1&
ti\'an  m\'ao   jīng l\'ing	&0.00&	1\\
ti\'an  m\=ao   jīng l\'ing	&0.00&	1&
 ti\=an   m\=ao   jīng līng 	&0.00&	1&
 ti\=an   m\=ao   jīng  lǐng 	&0.00&	1\\
 ti\=an   m\=ao   jīng l\'ing	&0.00&	1&
 - & - &- & 
 -& - & -\\

								\hline
		\end{tabular}
		\label{full-dot}
	\end{table*}
	\begin{table*}[h]
	\centering
	\small	
\caption{Fuzzy words of Tencent. The column \emph{dis.} presents the dissimilarity distance according to the generation algorithm, rounded up to the second decimal point. The column \emph{w. rate} presents the wake-up rate. }
	\begin{tabular}{lcc||lcc||lcc}
		\hline
	 & dis. & w. rate & & dis. & w. rate & & dis. & w. rate\\
		\hline
		
ji\=ong ni\`ao \`er líng	&0.15&	0.3&
jǐn s\`i \`ao líng	&0.09&	0.1&
ji\=ong s\`i \`er li\'an	&0.08&	0.1\\
ji\v ong sh\`i \`er li\'an	&0.08&	0.2&
ji\v ong s\`i \`er li\'an	&0.08&	1&
ji\v ong s\`i \`er li\'ang	&0.08&	0.3\\
ji\v ong z\`i \`er li\'ao	&0.07&	0.9&
ji\=ong s\`i \`er mín	&0.07&	0.1&
ji\v ong s\`i \`er mín	&0.07&	0.3\\
ji\v ong s\`i \`er li\=ao	&0.07&	0.3&
ji\v ong s\`i \`er li\'ao	&0.07&	1&
ji\v ong s\`i \`er l\'ong	&0.05&	0.2\\
ji\=ong s\`i \`ang líng	&0.05&	0.1&
ji\v ong sh\`e \`er līn	&0.05&	0.8&
ji\v ong sh\`e \`er líng	&0.05&	0.1\\
j\v u s\`i \`er lín	&0.05&	0.1&
jǐn s\`i \`er líng	&0.05&	0.2&
jǐng s\`i \`er líng	&0.05&	0.3\\
ji\v ong s\`i \`ai lín	&0.05&	1&
ji\v ong s\`i \`ai líng	&0.05&	1&
ji\v ong s\`i \`er míng	&0.04&	0.7\\
ji\v ong t\`i \`er líng	&0.03&	0.1&
ji\v ong s\`e \`er lín	&0.03&	1&
ji\=ong s\`e \`er líng	&0.03&	0.5\\
ji\v ong s\`e \`er līng	&0.03&	1&
ji\v ong s\`e \`er líng	&0.03&	1&
ji\=ang s\`i \`er líng	&0.02&	0.5\\
ji\=e s\`i \`er líng	&0.02&	0.4&
ji\v ang s\`i \`er líng	&0.02&	0.1&
ji\v e s\`i \`er líng	&0.02&	0.4\\
ji\=ao shī \`er līn	&0.02&	0.4&
ji\=ao s\`i \`er níng	&0.02&	0.9&
ji\=ao sī \`er l\`in	&0.02&	0.4\\
ji\=ao s\`i \`er l\`in	&0.02&	1&
ji\=ao sī \`er lín	&0.02&	0.1&
ji\=ao sǐ \`er lín	&0.02&	0.2\\
ji\=ao s\`i \`er lín	&0.02&	1&
ji\=ao sī \'er líng	&0.02&	1&
ji\=ao sī \`er l\`ing	&0.02&	0.1\\
ji\=ao sī \`er lǐng	&0.02&	0.7&
ji\=ao sǐ \'er líng	&0.02&	1&
ji\=ao sǐ \`er l\`ing	&0.02&	0.2\\
ji\=ao s\`i \`er l\`ing	&0.02&	0.9&
ji\=ao sǐ \`er lǐng	&0.02&	0.8&
ji\=ao s\`i \`er līng	&0.02&	1\\
ji\=ao sǐ \`er líng	&0.02&	0.5&
ji\=ao s\`i \`er líng	&0.02&	0.9&
ji\v ao sǐ \`er líng	&0.02&	0.2\\
ji\v ong z\`i \`er líng	&0.01&	1&
ji\v ong q\`i \`er líng	&0.01&	0.1&
ji\v ong s\`i \`er lí	&0.01&	1\\
ji\=ong s\`i \`en líng	&0.01&	0.3&
ji\=ong s\`i \`er níng	&0.01&	0.1&
ji\v ong sǐ \`er níng	&0.01&	0.2\\
ji\v ong s\`i \`e líng	&0.01&	0.4&
ji\v ong s\`i \`er níng	&0.01&	0.3&
ji\v ong sh\`i \`er līn	&0.01&	0.5\\
ji\=ong sǐ \`er lín	&0.01&	0.3&
ji\v ong sh\`i \`er lín	&0.01&	0.2&
ji\v ong s\`i \`er līn	&0.01&	1\\
ji\v ong shī \`er líng	&0.01&	0.5&
ji\v ong s\`i \v er lín	&0.01&	1&
ji\v ong sǐ \`er lín	&0.01&	0.9\\
ji\v ong s\`i \`er lín	&0.01&	1&
ji\=ong sǐ \`er l\`ing	&0.01&	0.3&
ji\=ong s\`i \`er l\`ing	&0.01&	0.5\\
ji\=ong s\`i \'er líng	&0.01&	0.4&
ji\=ong s\`i \`er līng	&0.01&	0.4&
ji\=ong s\`i \`er lǐng	&0.01&	0.3\\
ji\=ong sǐ \`er líng	&0.01&	0.2&
ji\v ong sh\`i \v er līng	&0.01&	1&
ji\v ong sh\`i \v er líng	&0.01&	1\\
ji\v ong sh\`i \`er lǐng	&0.01&	0.4&
ji\v ong sh\`i \`er līng	&0.01&	0.1&
ji\v ong sh\`i \`er líng	&0.01&	0.4\\
ji\v ong sī \`er líng	&0.01&	1&
ji\v ong sǐ \`er l\`ing	&0.01&	1&
ji\v ong s\`i \'er líng	&0.01&	1\\
ji\v ong s\`i \v er līng	&0.01&	1&
ji\v ong sǐ \`er líng	&0.01&	1&
ji\v ong s\`i \`er lǐng	&0.01&	1\\
ji\v ong s\`i \`er līng	&0.01&	1&
ji\v ong s\`i \v er líng	&0.01&	1&
ji\v ong s\`i \`er líng	&0.01&	1\\

						\hline
		\end{tabular}
		\label{full-dot}
	\end{table*}
\begin{table*}[h]
	\centering
	\small
	\caption{Fuzzy words of Amazon Echo. The column \emph{dis.} presents the dissimilarity distance according to the generation algorithm, rounded up to the second decimal point. The column \emph{w. rate} presents the wake-up rate.}
	\begin{tabular}{lcc||lcc||lcc||lcc||lcc}
		\hline
	 & dis. & w. rate & & dis. & w. rate & & dis. & w. rate & & dis. & w. rate& & dis. & w. rate\\
		\hline
alexoer &	0.44&	0.6&
ilexcer	&0.37&	1&
ileqcer	&0.37&	0.9&
ilekcer	&0.37&	0.7&
ileqsar	&0.35&	1\\
ilexsar	&0.35&	1&
ilexsur	&0.35	&1&
ileksur&	0.34&	0.8&
ileqsur&	0.34&	0.6&
ilebser	&0.33&	0.7\\
aleqsxr	&0.32&	0.7&
arlecci&	0.30&	0.6&
alexsur	&0.30&	1&
alexsar	&0.30&	0.9&
alexser	&0.30&	0.8\\
aleksar	&0.28&	0.7&
alecsar	&0.28&	0.6&
aleqsar	&0.28&	0.5&
ileqci	&0.27&	1&
ileckci	&0.27&	0.1\\
ilekci	&0.27&	0.1&
alecsur	&0.26&	0.7&
aleqsur	&0.26&	0.6&
aleksur&	0.26&	0.6&
irlegsa	&0.25	&0.8\\
alexcer	&0.25&	0.8&
urlecsa	&0.24	&1&
urlexsa	&0.24&	1&
erlecsa	&0.24&	1&
urleksa	&0.24&	1\\
ilexca	&0.24&1&
ilexqa&	0.24&	1&
erleksa&	0.24&	0.9&
erlexsa	&0.24&	0.1&
alekcer&	0.23&	0.7\\
aleqcer	&0.23&	0.6&
alicsur&	0.23&	0.5&
ilexsca&	0.22&	1&
ileckca&	0.22&	0.2&
aleksr&	0.22&	0.5\\
lexac&	0.22&	0.4&
alexsib&	0.21&	1&
ilelksa&	0.19&	1&
ileccsa&	0.19&	1&
ilecksa&	0.19&	1\\
ilekhsa&	0.19&	1&
ilekcsa&	0.19&	1&
alexusi&	0.19&	0.3&
nlecses&	0.17&	1&
ileqsca&	0.17&	1\\
ileksca&	0.17&	1&
ilexssa&	0.17&	1&
alexcpa&	0.17&	0.7&
jleksib&	0.17&	0.9&
lleksra&	0.17&	0.6\\
lecsca&	0.17&	0.4&
lecsba&	0.17&	0.2&
alektci&	0.16&	0.5&
alexsca&	0.16&	0.7&
vqleksa&	0.16&	0.9\\
lfkqsa&	0.16&	0.6&
arlekca&	0.15&	0.9&
aldcxa&	0.15&	0.8&
ameqdsa&	0.14&	0.4&
dlecdsa	&0.13&	1\\
aleqsrk	&0.13&	0.5&
ileqsaa	&0.12&	1&
alexuaa	&0.12&	0.7&
alebqsa&	0.12&	1&
gleksta	&0.12&	0.5\\
allexsa&	0.12&	1&
alelksa&	0.12&	0.9&
aleqtca&	0.12&	0.7&
alebsab&	0.11&	0.8&
hlexqa&	0.11&	0.6\\
alfbsa&	0.11&	0.6&
dpecssa&	0.11&	1&
dpeccsa	&0.11&	0.9&
ileqsa	&0.10&	1&
ileqssa&	0.10&	1\\
ilekssa&	0.10&	1&
ilexsa&	0.10&	1&
alexca&	0.10&	1&
alexqa&	0.10&	1&
alecsoi&	0.10&	0.4\\
aleckci&	0.10&	0.7&
aleqci&	0.10&	0.5&
alekci&	0.10&	0.5&
arlexsa&	0.10&	1&
arlecsa	&0.10&	1\\
arleksa	&0.10&	1&
alexssa	&0.10&	1&
alexsra	&0.10&	1&
alexsia	&0.10&	0.9&
alexsta	&0.10&	0.9\\
lecsa	&0.09&	0.7&
aleksai	&0.09&	0.5&
alecdsa	&0.08&	1&
alecpsa	&0.08&	0.9&
alecrsa	&0.08&	0.7\\
aleqsca	&0.08&	0.8&
aleksua	&0.08&	0.8&
alecsda	&0.08&	0.8&
aleksca	&0.08&	0.6&
aleksba	&0.08&	0.6\\
aleqsda	&0.08&	0.6&
alecsca	&0.08&	0.5&
aleqsra	&0.08&	0.5&
cleksaa	&0.07&	0.9&
aldqssa	&0.07&	1\\
alexci	&0.05&	0.1&
aleckca	&0.05&	0.5&
alekca	&0.05&	0.5&
alecssj	&0.05&	0.1&
alexsaa	&0.04&	1\\
alebsa	&0.04&	1&
alepsa	&0.03&	1&
alepsah	&0.03&	0.9&
alexsa	&0.02&	1&
anexa	&0.02&	0.5\\
alekssi	&0.02&	0.3&
aleqsaa	&0.02&	0.9&
aleccsa	&0.01&	1&
aleqhsa	&0.01&	1&
alekcsa	&0.01&	1\\
alecqsa	&0.01&	1&
alechsa	&0.01&	0.9&
- &- &- & 
- &- &- & 
- &- &- \\

		\hline
		\end{tabular}
		\label{full-echo}
	\end{table*}
\begin{table*}[h]
	\centering
	\small
	\caption{Fuzzy words of Amazon Echo Dot. The column \emph{dis.} presents the dissimilarity distance according to the generation algorithm, rounded up to the second decimal point. The column \emph{w. rate} presents the wake-up rate. }
	\begin{tabular}{lcc||lcc||lcc||lcc||lcc}
		\hline
	 & dis. & w. rate & & dis. & w. rate & & dis. & w. rate & & dis. & w. rate& & dis. & w. rate\\
		\hline
ureqssr&	0.44&	0.1&
arleqsr	&0.40&	1&
ilekcer	&0.37&	1&
ilexcer	&0.37&	0.1&
ilexsur	&0.35&	1\\
ileqsar	&0.35&	1&
ileksar&	0.35&	1&
ileksur&	0.34&	1&
alekcir&	0.32&	0.7&
ileqser&	0.30&	1\\
ermeqqa&	0.30&	0.9&
alexser	&0.30&	1&
alexsur	&0.30&	1&
alexsar	&0.30&	1&
ilexbs	&0.28&	0.1\\
urlexca	&0.28&	1&
aleqsar	&0.28&	0.9&
alecsar	&0.28&	0.7&
aleksar	&0.28&	0.7&
ilekci	&0.27&	1\\
ileqci	&0.27&	1&
ilefksa	&0.26&	1&
alecsur	&0.26&	0.8&
aleksur	&0.26&	0.8&
aleqsur	&0.26&	0.7\\
alexcer	&0.25&	0.8&
urleqsa	&0.24&	1&
urleksa	&0.24&	1&
erleqsa	&0.24&	1&
urlecsa	&0.24&	1\\
erleksa	&0.24&	1&
ilexca	&0.24&	1&
erlexsa	&0.24&	0.7&
allesab	&0.24&	0.1&
aleqcer	&0.23&	0.7\\
alekcer	&0.23&	0.7&
ilexsca	&0.22&	0.8&
ileckca	&0.22&	1&
ilekca	&0.22&	0.7&
alekser	&0.22&	1\\
aleqser	&0.22&	0.7&
alekqci	&0.22&	0.5&
blexasi	&0.21&	0.4&
uklefca	&0.20&	1&
alexa c	&0.20&	0.7\\
aleqxda	&0.19&	0.9&
arlesca	&0.19&	1&
ilelksa	&0.19&	1&
allexba	&0.19&	0.9&
ilecksa	&0.19&	1\\
ileccsa	&0.19&	1&
arlexca	&0.18&	1&
arlcxsa	&0.17&	0.5&
ileksda	&0.17&	1&
ilexssa	&0.17&	0.9\\
ileksca	&0.17&	0.2&
leqsuq	&0.17&	0.7&
ikecssa	&0.16&	0.9&
clebkca	&0.16&	0.2&
alexsca	&0.16&	1\\
alexsua	&0.16&	1&
alelkca	&0.15&	0.8&
cleqsai	&0.15&	0.1&
aleqcda	&0.15&	0.7&
ajecksa	&0.14&	0.3\\
alexa a	&0.14&	0.7&
akhdxsa	&0.14&	0.7&
aleqciq	&0.13&	0.3&
brleksa	&0.13&	1&
alecsid	&0.13&	0.6\\
aqlecsa	&0.13&	1&
clemcsa	&0.13&	1&
cleqsra	&0.13&	0.7&
ilexsaa	&0.12&	1&
allexsa	&0.12&	1\\
alelksa	&0.12&	1&
aleqtba	&0.11&	0.7&
ileksa	&0.10&	1&
ileqsa	&0.10&	1&
ilexsa	&0.10&	1\\
ilekssa	&0.10&	1&
alexca	&0.10&	1&
ileqssa	&0.10&	1&
aleckci	&0.10&	0.7&
alekci	&0.10&	0.7\\
aleqci	&0.10&	0.6&
lecta	&0.10&	0.3&
arlecsa	&0.10&	1&
arleksa	&0.10&	1&
arleqsa	&0.10&	0.9\\
arlexsa	&0.10&	0.9&
alexssa	&0.10&	1&
alexsia	&0.10&	1&
alexsda	&0.10&	0.9&
anexsba	&0.10&	0.1\\
blebssa	&0.09&	1&
alectda	&0.09&	0.8&
ulzxsw	&0.09&	0.1&
akekfa	&0.09&	0.1&
alecpsa	&0.08&	1\\
aleqpsa	&0.08&	1&
alaxssa	&0.08&	0.9&
aleqtsa	&0.08&	1&
aleqsca	&0.08&	0.8&
aleksca	&0.08&	0.7\\
alecsra	&0.08&	0.7&
alecsca	&0.08&	0.6&
alfcssa	&0.07&	0.6&
alfcsa	&0.07&	0.2&
aldcsa	&0.07&	0.8\\
hleqsaa	&0.06&	1&
dlechsa	&0.06&	0.9&
fleqsia	&0.06&	0.9&
dleccsa	&0.06&	0.9&
alersa	&0.06&	0.7\\
alexci	&0.05&	1&
blecssa	&0.05&	1&
blecksa	&0.05&	1&
kleqsia	&0.05&	0.9&
alekca	&0.05&	0.9\\
aleckca	&0.05&	0.7&
gleccsa	&0.05&	0.9&
hleqsa	&0.04&	1&
alebssa	&0.04&	1&
alekfa	&0.03&	1\\
alepsia	&0.03&	0.7&
aleqsha	&0.02&	0.8&
alechsa	&0.01&	1&
alecksa	&0.01&	1&
aleqtta	&0.01&	0.6\\

		\hline
		\end{tabular}
		\label{full-dot}
	\end{table*}
	\begin{table*}[h]
	\centering
	\small	
\caption{Fuzzy words of Google. The column \emph{dis.} presents the dissimilarity distance according to the generation algorithm, rounded up to the second decimal point. The column \emph{w. rate} presents the wake-up rate. }
	\begin{tabular}{lcc||lcc||lcc||lcc}
		\hline
	 & dis. & w. rate & & dis. & w. rate & & dis. & w. rate & & dis. & w. rate\\
		\hline
hea gougll&	0.33&	0.5&
heiigoogaa&	0.30&	0.3&
heay gugal&	0.30&	1&
hey gooogov a&	0.30&	1\\
heii googurl&	0.27&	0.7&
heii gugurl	&0.27	&0.6&
heii googerl&	0.27&	0.6&
hei googll a&	0.27&	1\\
hei gooo r	&0.23	&0.2&
heiy googow l&	0.23&	0.5&
heiy googol l&	0.23&	0.1&
heiy googerl&	0.20&	0.8\\
haii googerl&	0.19&	0.1&
haii gugerl	&0.19	&0.1&
haii gugurl	&0.19	&0.1&
hay googou l&	0.19&	0.9\\
heii gugal	&0.18	&1&
hey googal c&	0.18&	1&
hei googal l&	0.18&	1&
hei googel l&	0.18&	1\\
heii googourl&	0.17&	0.9&
hay googgrl	&0.16	&0.6&
heii gugull	&0.16	&1&
hei googerl	&0.15	&0.7\\
hey googurl	&0.15&	0.7&
hay googerl	&0.15&	0.6&
hey googerl	&0.15&	0.6&
hay gugerl	&0.15&	0.6\\
hei gugurl	&0.15&	0.6&
hey gugerl	&0.15&	0.5&
hay googurl	&0.15&	0.5&
hei gugerl	&0.15&	0.5\\
he googarl	&0.15&	0.6&
heii googal	&0.14&	1&
hei googal a&	0.14&	1&
hey goooarr	&0.14&	1\\
hei googar	&0.13&	0.7&
hay googour	&0.13&	0.7&
heiy gugourl&	0.13&	1&
heiy googrln&	0.13&	1\\
heiy googarl&	0.13&	0.9&
haii googarl&	0.13&	0.8&
haii gugarl	&0.13	&0.1&
hey googou	&0.11&	1\\
hei goooar	&0.11&	1&
haii gugal	&0.10&	1&
hey googout	&0.10&	0.5&
haii googak	&0.09&	1\\
heiy googourl&	0.08&	1&
hey gugarl	&0.08&	0.8&
hei gugourl	&0.08&	0.8&
hay googarl	&0.08&	0.7\\
hay gugarl	&0.08&	0.7&
hey googarl	&0.08&	0.6&
hei gugarl	&0.08&	0.6&
hay gugorl	&0.08&	0.6\\
hey goegil	&0.07&	0.9&
heyy gugil	&0.07&	0.1&
he googal	&0.07&	1&
hey googouraa&	0.07&	1\\
hei googala	&0.06&	1&
hey googau	&0.06&	1&
heiy googall&	0.06&	1&
heih googal	&0.06&	1\\
hey googourm&	0.06&	0.7&
haii googal	&0.05&	1&
hay gugal	&0.05&	1&
hei gugal	&0.05&	1\\
hey guugal	&0.05&	0.7&
hey googav	&0.04&	1&
hey ggugil	&0.04&	0.3&
hay ggufal	&0.03&	1\\
hay googourl&	0.03&	0.8&
hei googourl&	0.03&	0.8&
hey googourl&	0.03&	0.7&
hay gugil	&0.02&	0.2\\
hey gugil	&0.02&	0.1&
hei gugil	&0.02&	0.1&
jay googal	&0.02&	0.8&
- &- &- \\

		\hline
		\end{tabular}
		\label{full-google}
	\end{table*}	
\begin{table*}[h]
	\centering
	\small
	\caption{Fuzzy words of Apple Siri. The column \emph{dis.} presents the dissimilarity distance according to the generation algorithm, rounded up to the second decimal point. The column \emph{w. rate} presents the wake-up rate.}
	\begin{tabular}{lcc||lcc||lcc||lcc}
		\hline
	 & dis. & w. rate & & dis. & w. rate & & dis. & w. rate & & dis. & w. rate\\
		\hline
hey sirr e&	0.37&	1&
hei suru r a&	0.35&	0.7&
hay scir e	&0.33&	0.6&
haiiasciree	&0.28&	0.6\\
heyisyree	&0.27&	0.2&
heii sirea	&0.25&	0.7&
haiy cire	&0.22&	0.8&
heai ssuree	&0.22&	0.5\\
hey sserea	&0.20&	0.8&
hay syrrie e&	0.20&	1&
heii ssuiri	&0.20&	0.8&
hay cieree a&	0.20&	0.7\\
hay syerib	&0.19&	0.1&
hei ssirr	&0.19&	0.8&
heiy ssuiree&	0.18&	0.5&
heii syrie	&0.16&	0.2\\
heiy sciere	&0.13&	0.6&
heiy sirie	&0.13&	0.1&
haiy scyrie	&0.13&	0.4&
heai serea	&0.12&	0.9\\
haii sciree	&0.12&	0.7&
heai ceri	&0.12&	0.7&
heai ssuiri	&0.12&	0.5&
hey sierie	&0.11&	1\\
hay sere e	&0.11&	1&
hay siri e	&0.11&	0.8&
haiy cori	&0.10&	0.4&
hay ssirib	&0.10&	0.9\\
heiy suree	&0.09&	0.9&
hei syeii	&0.08&	0.1&
hai sirie	&0.07&	1&
hey ssire	&0.07&	0.9\\
hai scere	&0.07&	0.8&
heiy cieri	&0.07&	0.8&
hey ceree	&0.07&	0.1&
heiy scirie	&0.06&	0.1\\
heiy syrie	&0.06&	0.1&
hay syrieqe	&0.06&	1&
hay sciria	&0.06&	1&
hay psyrria	&0.06&	1\\
haiy syrie	&0.06&	0.7&
bay scirie	&0.03&	0.8&
hey sori	&0.02&	1&
hay scirih	&0.02&	1\\
hey scirih	&0.02&	1&
hei ssoree	&0.02&	0.9&
hay surie	&0.01&	1&
hay suri	&0.01&	0.6\\
hey suri	&0.01&	0.5&
hay cieri	&0.01&	1&
hai sciri	&0.01&	0.9&
hay sceri	&0.01&	0.7\\

		\hline
		\end{tabular}
		\label{full-siri}
	\end{table*}	

\end{CJK*}
\end{document}